%% file: paper.tex
\documentclass[10pt,a4paper,logo]{paper}

\usepackage[all]{hypcap}

\usepackage[authoryear, round]{natbib}

\usepackage{nicefrac}       
\usepackage{multirow}
\usepackage[capitalize,nameinlink]{cleveref}
\usepackage{dsfont}
\usepackage{makecell}
\usepackage{subcaption}
\usepackage{bbm}
\usepackage{wrapfig}
\usepackage{mathtools,xparse}
\usepackage{array}
\usepackage{scalerel}
\usepackage{tablefootnote}
\usepackage{tocloft}
\usepackage{adjustbox}
\usepackage{setspace}
\usepackage[most,skins,theorems,listings]{tcolorbox}

\usepackage{xspace}
\usepackage{threeparttable}
\usepackage{algorithm}
\usepackage{algpseudocode}
\usepackage[normalem]{ulem}
\usepackage{titletoc}

\newtcolorbox{promptbox}{
  colback=gray!5,
  colframe=gray!70,
  boxrule=0.8pt,
  arc=2mm,
  left=10pt,
  right=10pt,
  top=10pt,
  bottom=10pt,
  width=\linewidth,
  fontupper=\ttfamily\footnotesize,
  enhanced,
}

\newboolean{showsection}
\setboolean{showsection}{true}
\makeatletter
\@namedef{ver@everyshi.sty}{}
\makeatother

\definecolor{custom_green}{rgb}{0.0, 0.5, 0.0}
\definecolor{custom_red}{rgb}{1.0, 0.01, 0.24}
\definecolor{custom_blue}{HTML}{C9DAF7}
\definecolor{custom_purple}{HTML}{D9D1E9}
\definecolor{title_blue}{HTML}{204899}
\definecolor{cite_blue}{HTML}{044dc1}
\definecolor{cite_purple}{HTML}{7406a7}

\hypersetup{
    colorlinks = true,
    citecolor = {cite_blue},
    linkcolor = {cite_purple},
    urlcolor = {cite_purple},
}

\input{macro}

\input{math_commands.tex}

\makeatletter
\def\mathcolor#1#{\@mathcolor{#1}}
\def\@mathcolor#1#2#3{%
  \protect\leavevmode
  \begingroup
    \color#1{#2}#3%
  \endgroup
}
\makeatother

\Crefformat{equation}{#2Eq.\;(#1)#3}
\Crefformat{figure}{#2Figure #1#3}
\Crefformat{assumption}{#2Assumption #1#3}
\Crefname{assumption}{Assumption}{Assumptions}

\usepackage{crossreftools}
\makeatletter
\renewcommand\footnoterule{%
  \kern 15\p@
  \hrule \@width 2in \kern 2.6\p@
  \vspace{4pt}
}
\makeatother

\providecolor{my-red}{RGB}{200,55,55}
\providecolor{ForestGreen}{RGB}{34,139,34}

\renewcommand{\eg}[1]{{\textit{e.g.}, }{#1}}

\newcommand{\ours}{\textsc{TRIAGE}\xspace}

\title{TRIAGE: Dialectical LLM Reasoning for Explainable Risk Prediction on Irregularly Sampled Medical Time Series}

\reportnumber{}

\author[1,\footnotesize{\textbf{*}}]{Hyeongwon Jang}
\author[1,\footnotesize{\textbf{*}}]{Gyouk Chu}
\author[2,3]{Changhun Kim}
\author[1]{Hangyul Yoon}
\author[3]{Jeonguk Lee}
\author[1,3]{Eunho Yang}
\author[4,\footnotesize{$\dagger$}]{Joonhyung Park}

\affil[1]{KAIST AI}
\affil[2]{University of Wisconsin--Madison}
\affil[3]{AITRICS}
\affil[4]{Kyung Hee University}

\correspondingauthor{Correspondence to: \email{\{janghw0911,kyouwook\}@kaist.ac.kr}, \email{joonhyp@khu.ac.kr}.}

\paperlinks{%
\href{https://hyeongwon-jang.github.io/triage-project-page/}{ 
  \raisebox{-0.2ex}{\includegraphics[height=0.9em]{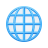}}\,
  \textcolor{RoyalBlue}{Project Page}
}
\hspace{0.8em}
\href{https://github.com/HyeongWon-Jang/TRIAGE}{ 
  \raisebox{-0.2ex}{\includegraphics[height=0.9em]{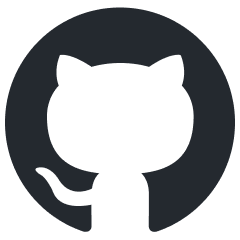}}\,
  \textcolor{RoyalBlue}{GitHub}
}
\hspace{0.8em}
\href{https://huggingface.co/collections/Hyeongwon/triage}{ 
  \raisebox{-0.2ex}{\includegraphics[height=0.9em]{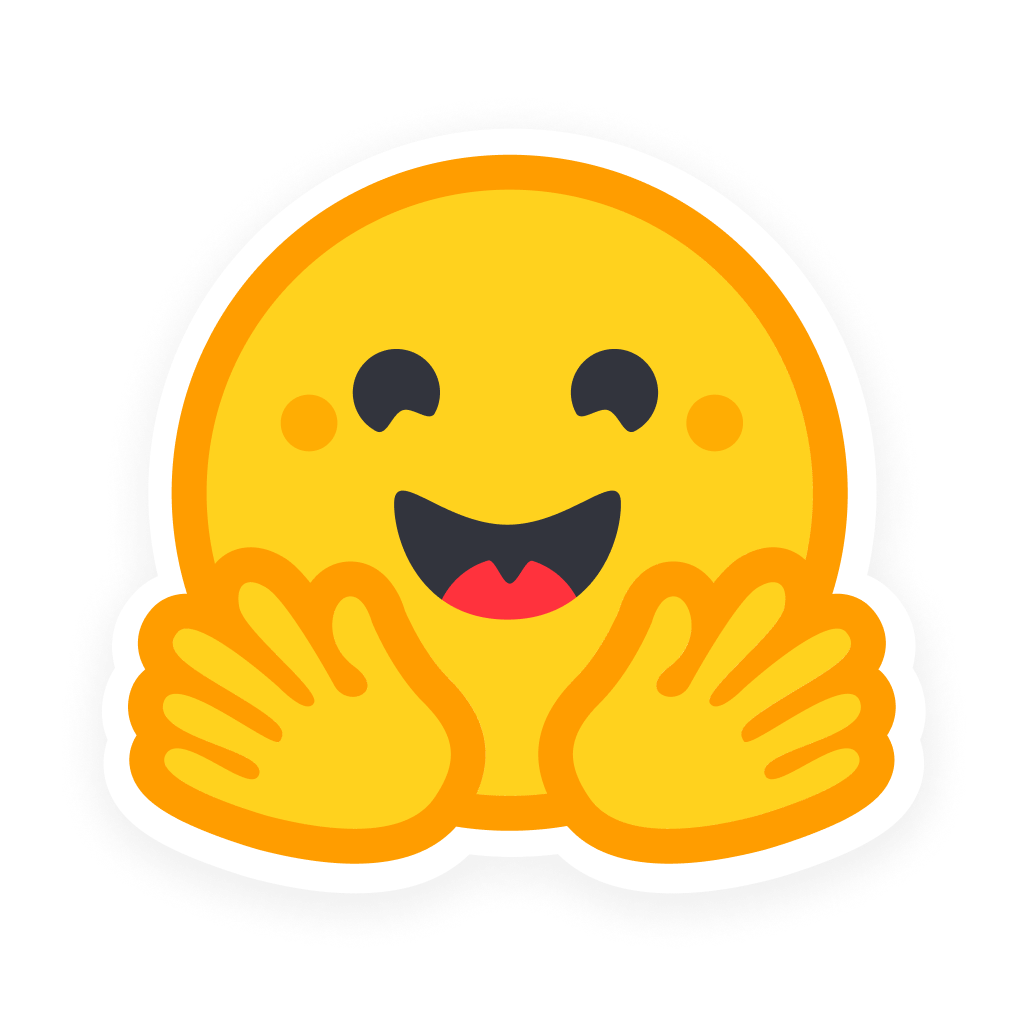}}\,
  \textcolor{RoyalBlue}{Hugging Face}
}
}

\input{sections/0_abstract}

\begin{document}

\maketitle

\input{sections/1_introduction}
\input{sections/2_related_work}
\input{sections/3_setup}
\input{sections/4_motivation}

\input{sections/5_method}
\input{sections/6_exp}
\input{sections/7_analysis}
\input{sections/8_conclusion}
\input{sections/9_acknowledgements}

\bibliographystyle{plainnat}
\bibliography{paper}

\clearpage
\appendix
\startcontents[appendix]
\noindent{\headingfont\LARGE Appendix}
\vspace{0.5em}
\printcontents[appendix]{}{1}{%
    \section*{Table of Contents}
    \setcounter{tocdepth}{2}
}
\clearpage
{\input{sections/appendix}}

\end{document}

%% file: macro.tex
\newcommand{\eg}{\emph{e.g.},\ }

\definecolor{blanchedalmond}{rgb}{1.0, 0.92, 0.8}
\definecolor{carmine}{rgb}{0.59, 0.0, 0.09}
\definecolor{lightblue}{rgb}{0.22,0.45,0.70}%

\renewcommand{\mathbf}{\boldsymbol}

\makeatletter
\def\Ddots{\mathinner{\mkern1mu\raise\p@
\vbox{\kern7\p@\hbox{.}}\mkern2mu
\raise4\p@\hbox{.}\mkern2mu\raise7\p@\hbox{.}\mkern1mu}}
\makeatother

\numberwithin{equation}{section}

\definecolor{amaranth}{rgb}{0.9, 0.17, 0.31}
\definecolor{antiquebrass}{rgb}{0.8, 0.58, 0.46}
\definecolor{antiquefuchsia}{rgb}{0.57, 0.36, 0.51}
\definecolor{chromeyellow}{rgb}{0.31, 0.47, 0.26}

\newcommand{\1}{\mathds 1}


%% file: math_commands.tex
\usepackage{amsmath,amsfonts,bm}

\def\eqref#1{equation~\ref{#1}}

\def\1{\bm{1}}

\DeclareMathAlphabet{\mathsfit}{\encodingdefault}{\sfdefault}{m}{sl}
\SetMathAlphabet{\mathsfit}{bold}{\encodingdefault}{\sfdefault}{bx}{n}



%% file: sections/0_abstract.tex
\begin{abstract}

Clinical early warning systems built on irregularly sampled medical time series (ISMTS) from electronic health records must deliver continuous risk scores for patient triage as well as interpretable rationales that clinicians can verify.
Large language models (LLMs) are uniquely positioned for both, deriving risk from their output probabilities and rationales from their medical knowledge.
However, we find that conventional LLM reasoning collapses graded risk into overconfident predictions and thereby undermines the cross-patient comparability on which triage depends.
We refer to this failure mode as \textit{risk polarization} and identify two underlying behaviors: early commitment to a single outcome, and one-sided reasoning that focuses only on the evidence for that outcome.
To address this, we propose \textbf{\ours}, a framework that trains an LLM to reason dialectically over competing clinical outcomes by eliciting outcome-specific rationales.
This dialectical formulation mitigates risk polarization, enabling a single LLM to jointly provide explicit clinical rationales and risk scores comparable across patients.
Across five ISMTS benchmarks, \ours improves mean AUPRC by 17.0\% and reduces mean calibration error by 82.8\% relative to the competitive LLM-based baseline, while surpassing the strongest ISMTS baseline by 3.5\% in mean AUPRC. 
\end{abstract}

%% file: sections/1_introduction.tex
\section{Introduction}
\label{sec:intro}

Predicting adverse clinical events from electronic health records (EHRs) is central to early warning systems in critical care. In EHRs, clinical observations are typically recorded sparsely at irregular intervals with substantial missingness and are thus commonly treated as irregularly sampled medical time series (ISMTS)~\citep{ISMTS}.
An effective clinical decision support system for ISMTS must satisfy two requirements in tandem that existing systems rarely provide together: (1) a \textit{well-calibrated, cross-patient comparable risk score} for triage and resource allocation~\citep{van2019calibration, foroutan2026discriminating}, and (2) a \textit{natural language explanation} grounded in reasoning over clinical knowledge that clinicians can evaluate~\citep{hou2025explainable}.

Existing approaches to this task focus solely on risk prediction.
Specialized deep learning models for ISMTS~\citep{che2018recurrent, horn2020seft, luo2024kedgn} achieve strong predictive performance through dedicated architectures tailored to irregular sampling and inter-variable dependencies, but produce opaque predictions with no accompanying rationale.
Post-hoc explainability methods~\citep{shap,dynamask} can be applied to identify contributing variables or time points, but their explanations are limited to feature-level or temporal importance, rather than the higher-level abstractions that clinicians reason over.

Large language models (LLMs) are uniquely positioned to bridge this gap as their output distributions provide a continuous signal for risk estimation, while their learned medical knowledge enables them to articulate complex reasoning in natural language.
Nevertheless, existing LLM-based approaches still do not satisfy both requirements within a single prediction process (Table~\ref{tab:related_work_comparison}).
One line of work~\citep{belyaeva2023helm, ehr-r1} extracts a continuous risk score from the model's implicit answer-token probability, thereby treating the LLM as a classifier without any natural language reasoning.
A second line of work~\citep{jiang2025kare, opentslm} elicits reasoning alongside the prediction but optimizes solely for the hard label, yielding discrete risk estimates that are not comparable across patients.
Hybrid pipelines pair LLM-generated interpretations with a neural predictor~\citep{lee2025timecap, record2vec}, but the risk score and the rationale for the prediction originate from separate components and are therefore not inherently coupled.

We first scrutinize why current LLMs struggle to meet both requirements: comparable risk scores and natural language explanations.
Our preliminary study uncovers a \emph{risk polarization problem}: when reasoning is elicited before risk scoring, the scores shift toward the extremes.
We further identify two recurring reasoning behaviors that contribute to this polarization.
First, rationales frequently commit to a single outcome \textit{in advance} of the final prediction (\eg{\textit{``Therefore, this patient is likely to die. Answer:''}}), making the resulting prediction nearly deterministic given the preceding biased text.
Second, rationales are typically \emph{one-sided}, presenting evidence for the committed outcome while omitting evidence for alternatives.
This issue is particularly problematic for ISMTS, where trajectories can contain signals of both deterioration and stabilization~\citep{arnold2009multicenter,hernandez2014stop, timing}.
Consequently, conventional LLM reasoning is fundamentally insufficient to provide both clinically grounded explanations and cross-patient comparable risk scores.

Motivated by these observations, we propose \textbf{T}ime Series \textbf{R}easoning by \textbf{I}nspecting \textbf{A}lternative Outcomes for \textbf{G}rounded Risk \textbf{E}stimation (\textbf{\ours}), a framework that mitigates risk polarization by grounding risk estimation in dialectical reasoning over alternative outcomes.
\ours generates a dedicated rationale for each candidate outcome, requiring the LLM to articulate patient-specific evidence for each possible alternative before making the final prediction.
It then derives the final risk from the LLM's implicit probability, without any intervening verdicts.
This design preserves a graded risk estimate while making the evidence considered for each outcome explicit.
\ours learns this behavior through a two-stage training pipeline: \emph{Dialectical Reasoning Supervision}, which teaches the LLM to generate outcome-specific rationales, followed by \emph{On-policy Self-refinement}, which uses reinforcement learning to further improve risk discrimination and calibration.

Experiments on five ISMTS benchmarks spanning mortality and sepsis prediction show that \ours improves both risk discrimination and calibration, achieving an improvement of \textbf{17.0\%} in mean AUPRC and a reduction of \textbf{82.8\%} in mean expected calibration error (ECE) compared with the strongest LLM-based baseline, while also surpassing the strongest specialized ISMTS baseline by 3.5\% in mean AUPRC.
We further verify the medical correctness and clinical helpfulness of \ours's dialectical rationales through case studies and human evaluation by clinical experts.

%% file: sections/2_related_work.tex
\section{Related Work}
\label{sec:related}

\input{tables/related_work_comparison}

\paragraph{Specialized neural architectures for ISMTS.}
Prior work has developed specialized architectures for irregularly sampled time series, including RNN-based models~\citep{che2018recurrent}, continuous-time models~\citep{chen2018neural,rubanova2019latent,bilovs2021neural,gao2026one}, set-based
encoders~\citep{horn2020seft}, interpolation-based models~\citep{shukla2018ipnet}, attention-based models~\citep{shukla2021mtand,zhang2023warpformer}, transformer-based models~\citep{tipirneni2022strats,li2023vitst}, and graph-based models~\citep{zhang2022raindrop,luo2024kedgn,luo2025hipatch,li2026garlic}.
While these methods advance representation learning and predictive accuracy on ISMTS, they do not provide any natural language explanations, which clinicians have found more understandable and reasonable than common explainable AI (XAI) methods~\citep{hou2025explainable}. 

\paragraph{LLMs as predictors without reasoning.}
One line of LLM-based approaches uses LLMs directly as risk predictors.
Some use the LLM as a \emph{pretrained backbone}, attaching a separate task head that maps its hidden states to a continuous prediction~\citep{timellm,s2ipllm,fsca,istsplm}.
Others adopt \emph{answer-token risk scoring}, treating the LLM as a classifier and reading a risk score from its answer-token probabilities, without an additional head~\citep{belyaeva2023helm,ehr-r1}.
Nevertheless, neither yields a rationale as part of the prediction.

\paragraph{LLMs as reasoners.}

Another line generates natural language reasoning, but in neither of its forms does the reasoning explain a continuous risk score. Hard-label methods produce a rationale followed by a discrete class label~\citep{jiang2025kare,liu2025vltime,opentslm}, leaving no continuous score with which to compare patients. Composite pipelines pair the generated text with a separate predictor, embedding LLM-generated rationales or summaries as features for a downstream predictor~\citep{nguyen2024carer,record2vec}, combining a discrete LLM prediction with a time-series encoder~\citep{lee2025timecap,jiang2026timexl}, or fusing the LLM's hidden states with a separate encoder while generating explanations in parallel~\citep{pathwayllm}. Since a downstream predictor or fusion module produces the final score, the text is an input to or a by-product of that score rather than its explanation.

Existing work thus uses LLMs as risk scorers without generated reasoning, reasoners for hard-label prediction, or auxiliary components in composite predictors.
\ours instead unifies reasoning and risk prediction in a single LLM, generating outcome-specific rationales and deriving a continuous risk score from the implicit probability conditioned on those rationales.
This autoregressive formulation directly couples explanation and prediction.

%% file: tables/related_work_comparison.tex
\begin{table}[t]
\vspace{-2mm}
\centering
\small

\providecommand{\xno}{\textcolor{my-red}{\texttimes}}
\providecommand{\cyes}{\textcolor{ForestGreen}{\ensuremath{\checkmark}}}

\setlength{\tabcolsep}{6pt}

\caption{
Comparison of ISMTS risk prediction approaches.
\textit{Natural language reasoning} covers any rationales or summaries in text format.
\textit{Score explained by reasoning}: the reasoning explains the continuous risk score rather than being its input or by-product.
}
\vspace{-1mm}
\begin{tabular}{@{}lccc@{}}
\toprule
Method
& \shortstack{Continuous\\[-1pt]risk score}
& \shortstack{Natural language\\[-2.5pt]reasoning}
& \shortstack{Score explained\\[-2.5pt]by reasoning} \\
\midrule

Specialized ISMTS architectures
& \multirow{2}{*}{\cyes}
& \multirow{2}{*}{\xno}
& \multirow{2}{*}{\xno} \\
{\scriptsize \citep{che2018recurrent,tipirneni2022strats,li2026garlic}}
& & & \\
\addlinespace[0.5pt]

LLMs as predictors
& \multirow{2}{*}{\cyes}
& \multirow{2}{*}{\xno}
& \multirow{2}{*}{\xno} \\
{\scriptsize \citep{timellm,belyaeva2023helm,ehr-r1}}
& & & \\
\addlinespace[0.5pt]

LLMs as reasoners, hard label
& \multirow{2}{*}{\xno}
& \multirow{2}{*}{\cyes}
& \multirow{2}{*}{\xno} \\
{\scriptsize \citep{jiang2025kare,liu2025vltime,opentslm}}
& & & \\
\addlinespace[0.5pt]

LLMs as reasoners, composite pipeline
& \multirow{2}{*}{\cyes}
& \multirow{2}{*}{\cyes}
& \multirow{2}{*}{\xno} \\
{\scriptsize \citep{lee2025timecap,record2vec,pathwayllm}}
& & & \\

\midrule
\textbf{\ours{} (Ours)}
& \cyes & \cyes & \cyes \\
\bottomrule
\end{tabular}

\vspace{-1mm}

\label{tab:related_work_comparison}
\vspace{-3mm}
\end{table}

%% file: sections/3_setup.tex
\section{Problem Setup: Clinical Risk Prediction}
\label{sec:problem_setup}

We formulate clinical risk prediction from ISMTS as a supervised classification problem. Let $\mathcal{D} = \{(\mathbf{s}_i, \mathbf{z}_i, y_i)\}_{i=1}^{N}$ denote a cohort of $N$ patients.
The temporal record $\mathbf{s}_i$ is a multivariate ISMTS represented as $\mathbf{s}_i = \{(t_{i,k}, v_{i,k}, m_{i,k})\}_{k=1}^{L_i}$, where $L_i$ is the total number of observations and each tuple specifies the time $t_{i,k}$, value $v_{i,k}$, and variable index $m_{i,k}\!\in\!\{1,\dots,D\}$ of the $k$-th measurement among $D$ clinical variables.
The static descriptor $\mathbf{z}_i\!\in\!\mathbb{R}^{d_s}$ encodes time-invariant patient attributes such as age and gender, and $y_i\!\in\!\{0,\dots,C-1\}$ is the clinical outcome label among $C$ classes.

Our objective is to learn a predictor that jointly outputs a patient-specific class distribution $\mathbf{p}_i = (p_{i,0}, \dots, p_{i,C-1})$ and a natural language rationale $r_i$ grounded in the patient's record, where $p_{i,c} \geq 0$ is the probability that patient $i$ belongs to class $c$.
While a discrete prediction can be obtained as $\hat{y}_i = \arg\max_{c} p_{i,c}$, our primary quantity of interest is the probability vector, which yields a continuous risk estimate over the candidate outcomes that is comparable across patients.

%% file: sections/4_motivation.tex
\section{Preliminary Study: Diagnosing Risk Polarization}
\label{sec:preliminary}

In this section, we identify why standard LLM reasoning fails at clinical risk prediction.
A natural approach to explainable risk prediction is to have the LLM reason over the patient record before making a prediction, ensuring that the risk score is grounded in explicit clinical evidence.
Through a preliminary study, we show that this approach instead collapses the score distribution into a degenerate extreme. We term this failure \textbf{risk polarization} and identify its two underlying causes. Full details of the preliminary study are provided in Appendix~\ref{app:det_preliminary}.
\paragraph{Setup.}
We study in-hospital mortality prediction on MIMIC-III~\citep{mimic3} over 3{,}228 patients with gpt-oss-120b~\citep{gptoss} and Qwen3-14B~\citep{qwen3}, reporting metrics for both discrimination (AUROC, AUPRC) and calibration (ECE, BS). 
The \textit{risk score} refers to the model's \emph{implicit} probability: the softmax probability assigned to the positive-class answer token derived from the output logits, rather than any verbalized number. This implicit probability provides a finer-grained and directly trainable risk signal in our setting (Appendix~\ref{app:prob_choice}).

\paragraph{Reasoning causes risk polarization.}

\begin{wrapfigure}{r}{0.49\columnwidth}
\centering
\includegraphics[width=\linewidth]{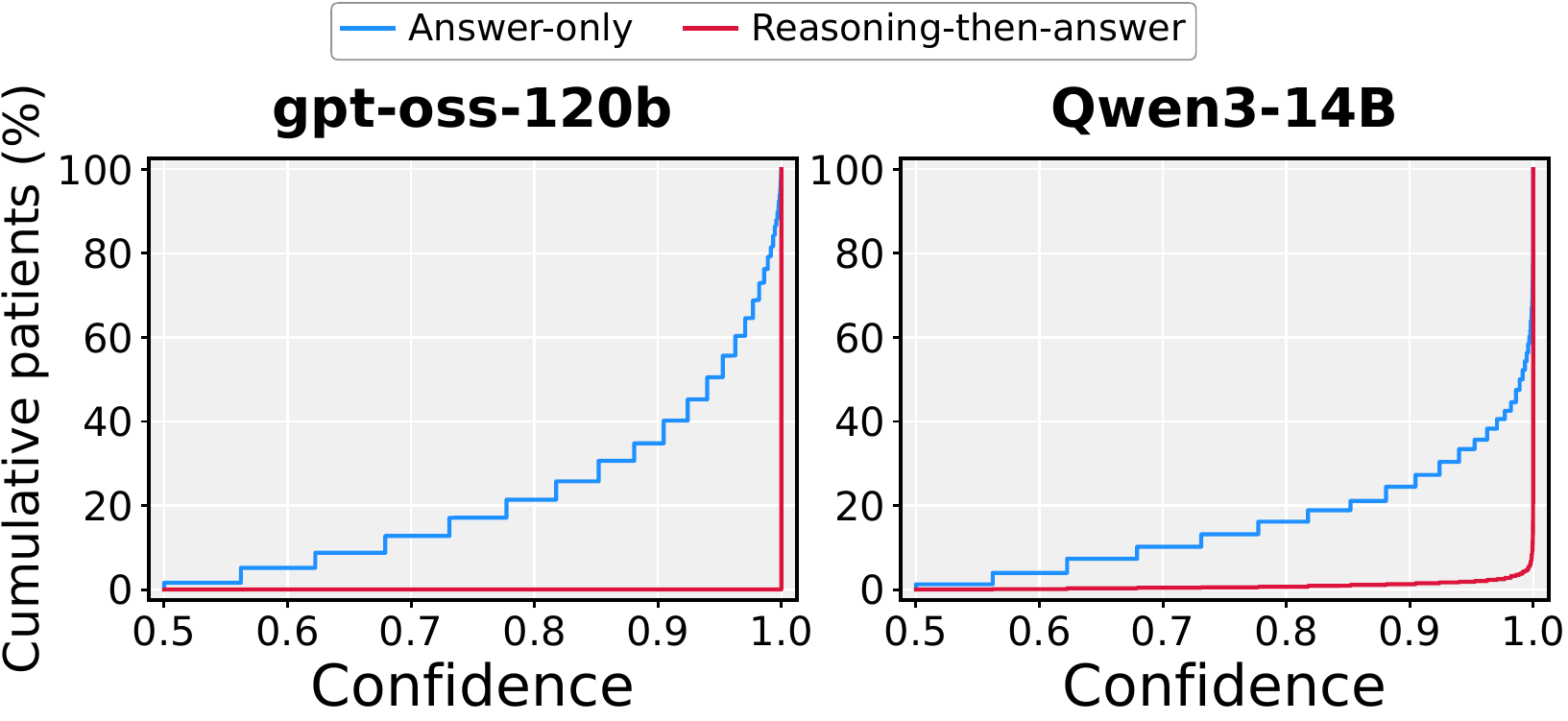}
\vspace{-1.5em}
\caption{Cumulative distribution of confidence on MIMIC-III in-hospital mortality prediction.}
\label{fig:polarization}
\vspace{-1.0em}
\end{wrapfigure}

We compare \emph{reasoning-then-answer}, where the model generates a rationale before predicting (\eg{\texttt{[Rationale]$\to$\allowbreak Yes/No}}), against an \emph{answer-only} baseline (\eg{\texttt{Yes/No}} without reasoning).
Under \emph{answer-only} prediction, confidence, the larger of the two outcome probabilities $\max_c p_{i,c}$, spreads over $[0.5, 1]$ for both models (Figure~\ref{fig:polarization}).
Under \emph{reasoning-then-answer}, however, it concentrates near 1; for gpt-oss-120b, even the lowest confidence across all patients is $0.9998$.
The risk score then reduces to a near-binary label and gives no meaningful basis for ranking patients.

\paragraph{Reasoning often pre-commits to an outcome.}

Under the \emph{reasoning-then-answer} paradigm, regardless of the preceding reasoning quality, rationales tend to terminate in an explicit verdict immediately before the answer position (\eg{\textit{``Therefore, this patient is likely to die. Answer:''}}). 
To quantify this tendency, we use gpt-oss-120b as the LLM judge to classify whether each rationale closes with such a verdict (judging instruction in Appendix~\ref{app:prompts}), and find that this pattern appears in 71.7\% of rationales from gpt-oss-120b and 44.1\% from Qwen3-14B.
Once such a verdict appears in the prefix, it dominates the conditioning context. Although the clinical evidence remains in the context, the answer token is overwhelmingly steered by the nearby verdict, concentrating the predicted-class probability on the outcome already stated in the rationale.
The rationale thus acts as a self-imposed hard prior that absorbs residual uncertainty before the label is produced.

\begin{wraptable}{r}{0.49\columnwidth}
\centering
\scriptsize
\vspace{-1.75em}
\caption{Results on MIMIC-III in-hospital mortality prediction, with risk estimated as the fraction of 10 sampled responses predicting death. AUROC and AUPRC are in \%.}
\label{tab:preliminary}
\vspace{-0.55em}
\setlength{\tabcolsep}{4pt}
\resizebox{\linewidth}{!}{%
\begin{tabular}{lcccc}
\toprule
Method & AUROC $\uparrow$ & AUPRC $\uparrow$ & ECE $\downarrow$ & BS $\downarrow$ \\
\midrule
\rowcolor{cyan!12}
\multicolumn{5}{l}{\textit{gpt-oss-120b}} \\
Standard  & \textbf{76.0} & 28.8 & 0.236 & 0.211 \\
Two-sided & 75.5 & \textbf{30.3} & \textbf{0.205} & \textbf{0.185} \\
\midrule
\rowcolor{cyan!12}
\multicolumn{5}{l}{\textit{Qwen3-14B}} \\
Standard  & 69.3 & 19.0 & 0.583 & 0.519 \\
Two-sided & \textbf{70.8} & \textbf{20.1} & \textbf{0.575} & \textbf{0.507} \\
\bottomrule
\end{tabular}%
}
\vspace{-0.90em}
\end{wraptable}

\paragraph{Reasoning is typically one-sided.}
Beyond the final verdict, the reasoning itself lacks balanced deliberation.
Inspecting traces reveals that the model exhibits \emph{confirmation bias}, committing to one outcome and citing only supportive evidence while disregarding countervailing signals (see Appendix~\ref{app:one_sided}).
This conflicts with clinical trajectories, where indicators of deterioration and stabilization may coexist~\citep{arnold2009multicenter,hernandez2014stop,timing}.
To test whether this tendency is systematic, we append a minimal two-sided instruction: \textit{``Weigh the evidence for survival and for in-hospital death, then decide.''}
This improves AUPRC and reduces ECE and Brier score while maintaining comparable AUROC (Table~\ref{tab:preliminary}); one-sided reasoning is therefore a systematic limitation, motivating dialectical deliberation as a better inductive bias.

\paragraph{Takeaways.}
Our preliminary study thus motivates two design principles: 
\textbf{(P1)} \textit{Reasoning should be structured, avoiding any explicit outcome verdict before risk estimation.}
\textbf{(P2)} \textit{Reasoning should inspect each candidate outcome separately, identifying the clinical evidence that supports it.}

%% file: sections/5_method.tex
\section{\ours}
\label{sec:method}

\begin{figure*}[t]
\centering
\begin{subfigure}{\textwidth}
    \centering
    \includegraphics[width=\textwidth]{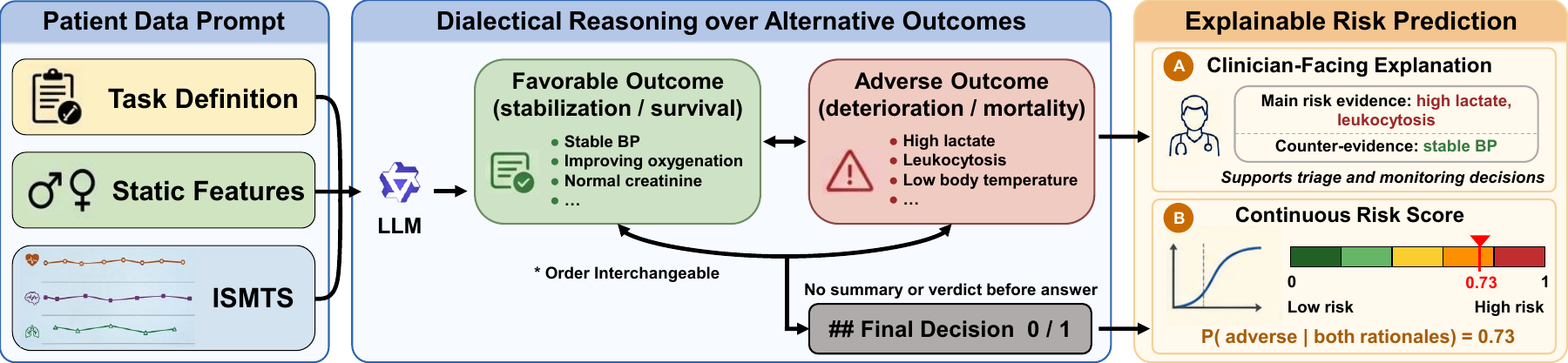}
    \label{fig:overview_a}
\end{subfigure}
\\[-1.0em]
\begin{subfigure}{\textwidth}
    \centering
    \includegraphics[width=\textwidth]{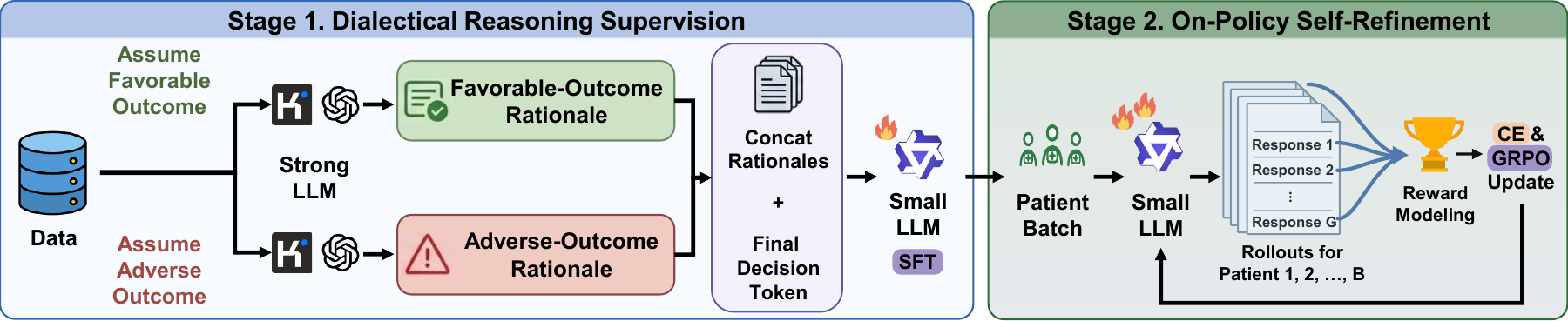}
    \label{fig:overview_b}
\end{subfigure}
\vspace{-2.5em}
\caption{Overview of \ours. \textbf{(Top)} Given irregular patient records represented as language inputs, \ours inspects alternative outcomes dialectically to jointly produce a patient risk score and clinician-facing explanations.
\textbf{(Bottom)} The training pipeline consists of dialectical reasoning supervision followed by on-policy self-refinement.}
\vspace{-1.25em}
\label{fig:overview}
\end{figure*}

Guided by the design principles from Section~\ref{sec:preliminary}, we propose \textbf{T}ime Series \textbf{R}easoning by \textbf{I}nspecting \textbf{A}lternative Outcomes for \textbf{G}rounded Risk \textbf{E}stimation (\textbf{\ours}; see Figure~\ref{fig:overview}). 
\ours introduces two components: (1) \emph{alternative outcome inspection}, which addresses \textbf{P2} by generating a dedicated rationale for each candidate outcome, and (2) \emph{verdict-free risk estimation}, which addresses \textbf{P1} by terminating the reasoning chain directly with the outcome token, so that no verdict or summary follows the rationales, and deriving a graded risk from the model's implicit distribution at that position (Section~\ref{subsec:reasoning}).
To learn this reasoning behavior, we introduce a two-stage training pipeline of \ours: (1) \emph{Dialectical Reasoning Supervision} instills the structured reasoning behavior using supervised fine-tuning (SFT) on synthesized outcome-conditioned rationales, and (2) \emph{On-policy Self-refinement} refines the model through reinforcement learning (RL) on its own samples to improve both discrimination and calibration (Section~\ref{subsec:training}).

\subsection{Dialectical Reasoning over Alternative Outcomes}
\label{subsec:reasoning}

\textbf{Input representation.}
Following set-based encoding for irregular time series~\citep{horn2020seft}, we serialize only the observed measurements in a variable-centric format rather than representing the full observation matrix. The input prompt $x_i$ for patient $i$ concatenates a structured task definition $\mathcal{P}$, textualized time-invariant attributes $\mathbf{t}_{z_i}$ (\eg{age, gender}), and temporally ordered observations $\mathbf{t}_{s_i}$ (see Figure~\ref{fig:prompt_format} for an example):
\begin{equation}
    x_i = \texttt{concat}(\mathcal{P},\, \mathbf{t}_{z_i},\, \mathbf{t}_{s_i}).
\end{equation}

\textbf{Alternative outcome inspection (P2).} As shown in Section~\ref{sec:preliminary}, LLM reasoning tends to align with a single outcome direction, citing only supportive evidence while disregarding evidence for alternative outcomes. This one-sided tendency is especially problematic for ISMTS, where patient trajectories frequently exhibit concurrent indicators of both deterioration and stabilization~\citep{arnold2009multicenter,hernandez2014stop, timing}.

To address this, \ours generates a dialectical reasoning chain that examines each candidate outcome separately before the final prediction, as illustrated in the upper part of Figure~\ref{fig:overview}. For a binary prediction task with candidate outcomes $\mathcal{Y} = \{y^+, y^-\}$, the model produces a dedicated rationale $r_{y_k}$ for each $y_k \in \mathcal{Y}$, articulating the evidence in the patient's observations that supports that outcome:
\begin{equation}
\label{eq:chain}
    \texttt{chain} = [r_{y_1},\, r_{y_2},\, \hat{y}], \quad \{y_1, y_2\} = \{y^-, y^+\},
\end{equation}
where $r_{y_k}$ is the rationale for outcome $y_k$, and $\hat{y}$ is the final predicted outcome token.
The order of the two rationales is interchangeable.
This structure realizes \textbf{P2} by requiring the model to consider both sides of the clinical evidence before reaching a decision, surfacing signals that one-sided reasoning would discard.

\textbf{Verdict-free risk estimation (P1).}
Even with dialectical rationales, a reasoning chain that concludes with a verdict before $\hat{y}$ reintroduces the risk polarization that motivates \textbf{P1}.
To preserve a graded risk signal, \ours structures the reasoning chain to end with a ``\#\# Final Decision'' header followed directly by a single outcome token, without any intermediate summary or verdict.
As detailed in Section~\ref{subsec:training}, we instill this structure through SFT on chains whose headers delimit each rationale and the final decision.
We map each outcome $y_k \in \mathcal{Y}$ to a designated token (\eg{``0'' for negative, ``1'' for positive}) and extract the corresponding logit $\ell_k$ at this fixed position.
The risk score is then computed as
\(
    P(y_k \mid x) = \exp(\ell_k) / \sum_{k'} \exp(\ell_{k'}).
\)
Because no verdict precedes this position, the distribution at the outcome token stays graded, and the resulting risk estimates are continuous and comparable across patients rather than near-deterministic.

\subsection{Two-Stage Training Pipeline}
\label{subsec:training}

\textbf{Dialectical Reasoning Synthesis.}
As illustrated in the lower-left part of Figure~\ref{fig:overview}, we elicit rationales from a strong LLM \emph{separately} for each candidate outcome: the LLM is prompted to assume that a given outcome holds and to identify only the patient features that support it (prompts are provided in Appendix~\ref{app:prompts}).
This process must adhere to two constraints.
First, the LLM must not reference any alternative outcome, ensuring that each rationale remains strictly outcome-specific rather than contrastive.
Second, the LLM must not fabricate unobserved evidence. In cases where no supporting evidence exists, the rationale is left blank.
These constraints yield concise, feature-grounded rationales while discouraging hallucinated or contrastive content.
The resulting rationales are concatenated with the ground-truth answer token to form reasoning traces as in Equation~\ref{eq:chain}.

\textbf{Stage 1: Dialectical Reasoning Supervision.}
Using the synthesized reasoning traces above, we fine-tune the target LLM to learn the dialectical reasoning procedure: generating a dedicated rationale for each candidate outcome followed by the ground-truth decision token.
Because the ordering of the outcome-specific rationales is interchangeable, we augment each training instance with both orderings to avoid associating a particular outcome with a fixed position in the reasoning chain.
The resulting model initializes the on-policy self-refinement in Stage~2.

\textbf{Stage 2: On-policy Self-refinement.}
After SFT on off-policy dialectical reasoning traces built from a strong model, the model can generate high-quality rationales.
However, its self-generated rationales may differ in distribution from the SFT targets produced by the strong model, leaving it susceptible to exposure bias when conditioning on its own prefixes at inference~\citep{agarwal2024policy}.

To address this limitation, we further post-train the model with Group Relative Policy Optimization (GRPO;~\citealp{shao2024deepseekmath}). The cross-entropy objective is applied only to the final decision token to supervise the implicit predictive probability, whereas the preceding reasoning tokens are optimized by a GRPO objective modified following Dr. GRPO~\citep{drgrpo} and DAPO~\citep{dapo}:\footnote{We apply a KL penalty term $\beta D_{\mathrm{KL}}(\pi_\theta\Vert\pi_{\mathrm{ref}})$ during training, which is omitted here for brevity.}
\begin{equation}
\label{eq:entire_loss}
\mathcal{L}(\theta)
=
\mathcal{L}_{\mathrm{GRPO}}(\theta)
+
\lambda \mathcal{L}_{\mathrm{CE}}(\theta),
\end{equation}
\begin{equation}
\label{eq:grpo}
\resizebox{\dimexpr\linewidth-2.5em\relax}{!}{%
$\displaystyle
\mathcal{L}_{\mathrm{GRPO}}(\theta)
=
-\mathbb{E}_{x\sim \mathcal{D}, \{r_j\}_{j=1}^G\sim \pi_{\theta_\mathrm{old}}(\cdot\mid x)}\Bigg[
\frac{1}{L_{\mathrm{r}}}
\sum_{j=1}^{G}\sum_{\tau\in\mathcal{T}^{\mathrm{r}}_j}
\min\!\left(
\rho_{j,\tau}\hat{A}_{j},
\operatorname{clip}\!\left(
\rho_{j,\tau},
1-\epsilon_{\text{l}},
1+\epsilon_{\text{h}}
\right)\hat{A}_{j}
\right)
\Bigg],
$%
}
\end{equation}
\begin{equation*}
\text{where}\quad
L_{\mathrm{r}}
=
\sum_{j=1}^{G}|\mathcal{T}^{\mathrm{r}}_j|,
\qquad
\rho_{j,\tau}
=
\frac{
\pi_{\theta}(r_{j,\tau}\mid x,r_{j,<\tau})
}{
\pi_{\theta_{\mathrm{old}}}(r_{j,\tau}\mid x,r_{j,<\tau})
},
\qquad
\hat{A}_{j}
=
R_j-\operatorname{mean}(\{R_k\}_{k=1}^{G})
,
\end{equation*}
and $\mathcal{T}^{\mathrm{r}}_j$ denotes the reasoning-token positions excluding the final decision token.

While intra-sample rewards provide a learning signal, they fail to explicitly encourage cross-patient discrimination. 
To ensure that risk scores are comparable across patients, we define rewards at the batch level.
Inspired by \citet{yuan2021large}, we define the reward for the $j$-th response of patient $i$ as
\begin{equation}
\label{eq:rl_reward}
\small
R_{i,j}
=
-\begin{cases}
\displaystyle
\frac{1}{|\mathcal{B}^{-}|}
\sum_{i'\in\mathcal{B}^{-}}
\left[m-(\sigma_{i,j}-\bar{\sigma}_{i'})\right]_+^2
& \text{if } y_i=1,\\[2mm]
\displaystyle
\frac{1}{|\mathcal{B}^{+}|}
\sum_{i'\in\mathcal{B}^{+}}
\left[m-(\bar{\sigma}_{i'}-\sigma_{i,j})\right]_+^2
& \text{if } y_i=0,
\end{cases}
\end{equation}
where
$\sigma_{i,j}$ is the log-odds of the $j$-th response $r_{i,j}$ for patient $i$,
$\bar{\sigma}_i=\frac{1}{G}\sum_{j=1}^{G}\sigma_{i,j}$,
and $[z]_+=\max\{z,0\}$.
Here, $m$ is the margin, and $\mathcal{B}^{-}$ and $\mathcal{B}^{+}$
denote the indices of negative and positive patients in the batch, respectively.
The reward pushes the log-odds of each response at least $m$ above (for positive patients) or below (for negative patients) the mean log-odds of each opposite-class patient in the batch.
This further strengthens the cross-patient comparability and calibration that on-policy training with a sample-level reward already improves (ablation in Appendix~\ref{app:reward_design}).

%% file: sections/6_exp.tex
\section{Experiments}
\label{sec:exp}

\subsection{Experimental Setup}

\paragraph{Datasets and metrics.}
We conduct experiments on five ISMTS benchmarks, P12~\citep{p12}, P19~\citep{p19}, MIMIC-III~\citep{mimic3}, MIMIC-IV~\citep{mimic4}, and eICU~\citep{eicu}.\footnote{\label{fn:physionet}MIMIC-III, MIMIC-IV, and eICU are distributed under PhysioNet's credentialed Data Use Agreement, which prohibits sending the data to third-party LLM services (\url{https://physionet.org/news/post/llm-responsible-use/}).}
P19 involves predicting sepsis onset within 6 hours, while the other four target in-hospital mortality.
For discrimination, we report the area under the ROC curve (AUROC) and the area under the precision-recall curve (AUPRC), with AUPRC as the primary metric given the severe class imbalance~\citep{davis2006relationship, saito2015precision}.
For calibration, we report the expected calibration error (ECE) and the Brier score (BS).
We additionally evaluate robustness under limited information, a practically relevant setting in real-world clinical environments (Appendix~\ref{app:robustness}), and report the inference cost of every method (Appendix~\ref{app:cost}).

\input{tables/main_result_emphasize}
\input{tables/main_calibration}

\paragraph{Baselines.}
We compare against representative ISMTS baselines, namely GRU-D~\citep{che2018recurrent}, mTAND~\citep{shukla2021mtand}, SeFT~\citep{horn2020seft}, Raindrop~\citep{zhang2022raindrop}, STraTS~\citep{tipirneni2022strats}, ViTST~\citep{li2023vitst}, KEDGN~\citep{luo2024kedgn}, Hi-Patch~\citep{luo2025hipatch}, GSNF~\citep{gao2026one}, and GARLIC~\citep{li2026garlic}.
We further compare with LLM-based methods, TimeCAP~\citep{lee2025timecap} and Record2Vec~\citep{record2vec}, as well as GPT-5.1\footnote{GPT-5.1 is not evaluated on MIMIC-III, MIMIC-IV, or eICU due to the data-use restriction (footnote~\ref{fn:physionet}).}~\citep{openai2025gpt51systemcard}, gpt-oss-120b~\citep{gptoss}, and Qwen3-4B~\citep{qwen3} under zero-shot prompting (scored by verbalized probabilities; Appendix~\ref{app:verbalized_zeroshot}).

\paragraph{Implementation details.}

To collect training data, we use GPT-5.1~\citep{openai2025gpt51systemcard} for P12 and P19, and, in accordance with data-use restrictions, locally served Kimi K2 Thinking~\citep{kimi} for MIMIC-III, MIMIC-IV, and eICU. 
Since the datasets are highly imbalanced, we apply oversampling following prior work~\citep{zhang2022raindrop, luo2024kedgn}, collecting multiple distinct rationales per minority sample. 
We apply the two-stage training pipeline to Qwen3-4B-Base~\citep{qwen3} (see Appendix~\ref{app:backbone} for backbone ablations).
Full details are in Appendix~\ref{app:exp_setup_details}.

\subsection{Main Results}

\paragraph{Discrimination (Table~\ref{tab:main_results_compact}).}
Zero-shot LLMs rank at the bottom.
Larger LLMs do better, but even GPT-5.1 remains far from the specialized ISMTS models: its AUPRC trails GRU-D by 7.0 points on P12 and by 47.3 points on P19.
This indicates that general-purpose reasoning alone does not transfer well to clinical risk prediction on ISMTS.
With SFT alone, \ours achieves a mean AUPRC of 55.3, which is 12.6\% higher than TimeCAP (49.1), the strongest LLM-based prediction method, and within 0.3 points of the best ISMTS baselines, GRU-D (55.5), KEDGN (55.4), and STraTS (55.0).
\ours thus matches specialized architectures while still producing natural language rationales.
RL adds a further gain. \ours reaches a mean AUPRC of 57.4, a 3.9\% relative improvement over its SFT counterpart and 3.5\% over GRU-D, with the best AUPRC on four of the five benchmarks and the second best on the remaining one.
Mean AUROC follows the same pattern with smaller margins (87.3 for \ours vs.\ 87.0 for STraTS, the best baseline).

\paragraph{Calibration (Table~\ref{tab:calibration}).}
TimeCAP and Record2Vec remain poorly calibrated, on par with or worse than zero-shot gpt-oss-120b (mean ECE of 0.219 and 0.285 vs.\ 0.239), and \ours with SFT alone improves on them only modestly (mean ECE 0.180, BS 0.138).
RL closes this gap, reducing the mean ECE from 0.180 to 0.038 and the mean BS from 0.138 to 0.077, relative reductions of 79\% and 45\%.
As a result, \ours achieves the lowest ECE and Brier score on every benchmark.

%% file: tables/main_result_emphasize.tex
\definecolor{refgray}{gray}{0.4}

\newcommand{\val}[2]{#1{\scriptsize$\pm$#2}}
\newcommand{\gval}[2]{\textcolor{refgray}{#1{\scriptsize$\pm$#2}}}
\newcommand{\gtxt}[1]{\textcolor{refgray}{#1}}
\begin{table*}[t]
\centering
\small
\setlength{\tabcolsep}{1pt}
\caption{
Discrimination performance on clinical risk prediction benchmarks.
Specialized ISMTS models (\gtxt{gray}) cannot explain their predictions and are shown for reference only;
\textbf{bold} and \underline{underline} mark the best and second-best results among natural-language-explainable methods.
For clarity, we report only the top-performing and recent ISMTS baselines; full results are provided in
Appendix~\ref{app:full_results}.
}
\vspace{-0.5em}
\label{tab:main_results_compact}
\begin{tabular*}{\textwidth}{@{\extracolsep{\fill}}lcccccccccc}
\toprule
\multirow{2}{*}{\textbf{Method}}
& \multicolumn{2}{c}{\textbf{P12}}
& \multicolumn{2}{c}{\textbf{P19}}
& \multicolumn{2}{c}{\textbf{MIMIC-III}}
& \multicolumn{2}{c}{\textbf{MIMIC-IV}}
& \multicolumn{2}{c}{\textbf{eICU}} \\
\cmidrule(lr){2-3} \cmidrule(lr){4-5} \cmidrule(lr){6-7} \cmidrule(lr){8-9} \cmidrule(lr){10-11}
& AUROC & AUPRC & AUROC & AUPRC & AUROC & AUPRC & AUROC & AUPRC & AUROC & AUPRC \\
\midrule

\rowcolor{gray!12}
\multicolumn{11}{c}{\textit{Specialized ISMTS Models} (no natural-language explanation)} \\
\midrule

\gtxt{GRU-D}
& \gval{87.1}{0.9} & \gval{56.3}{2.3}
& \gval{89.7}{0.9} & \gval{56.6}{2.9}
& \gval{84.7}{0.2} & \gval{49.0}{0.6}
& \gval{87.5}{0.5} & \gval{59.2}{1.4}
& \gval{84.4}{0.7} & \gval{56.4}{2.2} \\

\gtxt{STraTS}
& \gval{87.3}{1.5} & \gval{57.4}{3.9}
& \gval{88.7}{1.4} & \gval{50.4}{2.3}
& \gval{85.3}{0.5} & \gval{48.3}{0.6}
& \gval{88.6}{0.4} & \gval{61.3}{1.1}
& \gval{85.3}{0.7} & \gval{57.4}{2.6} \\

\gtxt{KEDGN}
& \gval{87.8}{1.1} & \gval{58.3}{3.2}
& \gval{88.2}{0.5} & \gval{53.5}{3.0}
& \gval{84.7}{0.4} & \gval{48.5}{0.9}
& \gval{87.9}{0.5} & \gval{60.1}{1.2}
& \gval{85.3}{0.6} & \gval{56.4}{1.4} \\

\gtxt{Hi-Patch}
& \gval{87.3}{0.8} & \gval{57.0}{3.7}
& \gval{88.7}{1.6} & \gval{52.1}{4.3}
& \gval{84.6}{0.6} & \gval{46.2}{0.3}
& \gval{87.7}{0.3} & \gval{57.7}{0.6}
& \gval{85.0}{0.8} & \gval{56.5}{2.5} \\

\gtxt{GSNF}
& \gval{84.7}{1.3} & \gval{50.7}{2.4}
& \gval{86.2}{2.0} & \gval{46.2}{3.2}
& \gval{83.6}{0.4} & \gval{40.5}{0.7}
& \gval{86.0}{0.5} & \gval{54.6}{1.6}
& \gval{83.4}{0.7} & \gval{54.4}{1.1} \\

\gtxt{GARLIC}
& \gval{86.6}{0.8} & \gval{55.3}{2.2}
& \gval{88.8}{0.6} & \gval{52.4}{1.8}
& \gval{85.5}{0.3} & \gval{50.5}{1.5}
& \gval{87.2}{0.5} & \gval{57.9}{1.8}
& \gval{85.2}{1.0} & \gval{58.2}{1.5} \\

\midrule
\rowcolor{cyan!12}
\multicolumn{11}{c}{\textit{Zero-shot LLMs}} \\
\midrule

GPT-5.1
& \val{83.9}{1.0} & \val{49.3}{0.5}
& \val{72.2}{2.9} & \val{9.3}{0.9}
& -- & --
& -- & --
& -- & -- \\

gpt-oss-120b
& \val{81.2}{0.7} & \val{43.4}{3.0}
& \val{64.2}{2.1} & \val{6.7}{0.6}
& \val{76.6}{0.6} & \val{31.0}{1.0}
& \val{70.8}{0.5} & \val{27.9}{1.0}
& \val{79.5}{1.2} & \val{47.4}{1.0} \\

Qwen3-4B
& \val{69.7}{1.9} & \val{26.7}{0.8}
& \val{56.6}{1.7} & \val{5.0}{0.6}
& \val{69.3}{0.5} & \val{20.2}{0.2}
& \val{65.6}{0.7} & \val{20.9}{0.8}
& \val{71.2}{1.4} & \val{31.4}{0.9} \\

\midrule
\rowcolor{cyan!12}
\multicolumn{11}{c}{\textit{LLM-based Prediction Methods}} \\
\midrule

Record2Vec
& \val{76.9}{1.3} & \val{38.2}{2.9}
& \val{79.5}{1.6} & \val{33.4}{3.8}
& \val{73.6}{0.1} & \val{31.9}{0.1}
& \val{79.5}{1.2} & \val{43.6}{1.9}
& \val{75.9}{0.5} & \val{42.5}{1.1} \\

TimeCAP
& \val{85.2}{1.0} & \val{52.8}{3.9}
& \val{85.4}{1.3} & \val{33.7}{2.3}
& \val{84.2}{0.1} & \val{46.9}{0.2}
& \val{87.3}{0.3} & \val{57.5}{0.5}
& \val{83.8}{1.0} & \val{54.6}{3.2} \\

\midrule

\textbf{\ours}$_{\text{SFT}}$
& \underline{\val{86.9}{1.0}} & \underline{\val{56.4}{1.9}}
& \underline{\val{88.9}{1.0}} & \underline{\val{52.2}{3.2}}
& \underline{\val{86.4}{0.3}} & \underline{\val{51.4}{0.9}}
& \underline{\val{87.5}{0.2}} & \underline{\val{59.7}{0.8}}
& \underline{\val{84.2}{1.3}} & \underline{\val{56.7}{0.9}} \\

\textbf{\ours}
& \textbf{\val{87.3}{1.2}} & \textbf{\val{59.0}{3.4}}
& \textbf{\val{89.3}{1.0}} & \textbf{\val{53.8}{1.4}}
& \textbf{\val{86.7}{0.2}} & \textbf{\val{54.1}{0.5}}
& \textbf{\val{87.8}{0.5}} & \textbf{\val{61.6}{2.1}}
& \textbf{\val{85.2}{0.5}} & \textbf{\val{58.7}{1.4}} \\

\bottomrule
\end{tabular*}
\vspace{-0.5em}
\end{table*}

%% file: tables/main_calibration.tex
\begin{table*}[t]
\centering
\small
\setlength{\tabcolsep}{1pt}
\caption{
Calibration performance on clinical risk prediction benchmarks. ECE and Brier score (BS); lower is better. The best results are indicated in \textbf{bold}.
}
\vspace{-0.5em}
\label{tab:calibration}
\begin{tabular*}{\textwidth}{@{\extracolsep{\fill}}lcccccccccc}
\toprule
\multirow{2}{*}{\textbf{Method}}
& \multicolumn{2}{c}{\textbf{P12}}
& \multicolumn{2}{c}{\textbf{P19}}
& \multicolumn{2}{c}{\textbf{MIMIC-III}}
& \multicolumn{2}{c}{\textbf{MIMIC-IV}}
& \multicolumn{2}{c}{\textbf{eICU}} \\
\cmidrule(lr){2-3}
\cmidrule(lr){4-5}
\cmidrule(lr){6-7}
\cmidrule(lr){8-9}
\cmidrule(lr){10-11}
& ECE & BS
& ECE & BS
& ECE & BS
& ECE & BS
& ECE & BS \\
\midrule

GPT-5.1
& 0.085 & 0.103
& 0.161 & 0.084
& -- & --
& -- & --
& -- & -- \\

gpt-oss-120b
& 0.263 & 0.185
& 0.321 & 0.187
& 0.244 & 0.174
& 0.172 & 0.161
& 0.196 & 0.165 \\

Qwen3-4B
& 0.427 & 0.309
& 0.679 & 0.527
& 0.609 & 0.484
& 0.355 & 0.262
& 0.490 & 0.382 \\

\midrule

Record2Vec
& 0.298 & 0.212
& 0.292 & 0.137
& 0.311 & 0.207
& 0.269 & 0.187
& 0.256 & 0.195 \\

TimeCAP
& 0.245 & 0.190
& 0.215 & 0.113
& 0.230 & 0.155
& 0.180 & 0.138
& 0.225 & 0.173 \\

\midrule

\textbf{\ours}$_{\text{SFT}}$
& 0.190 & 0.149
& 0.149 & 0.091
& 0.214 & 0.153
& 0.174 & 0.142
& 0.172 & 0.156 \\

\textbf{\ours}
& \textbf{0.044} & \textbf{0.089}
& \textbf{0.039} & \textbf{0.035}
& \textbf{0.032} & \textbf{0.077}
& \textbf{0.035} & \textbf{0.080}
& \textbf{0.038} & \textbf{0.102} \\

\bottomrule
\end{tabular*}
\end{table*}

%% file: sections/7_analysis.tex
\section{Analysis of \ours Reasoning}
\label{sec:analysis}

We analyze the reasoning of \ours from three perspectives:\footnote{Unless otherwise noted, we report all analyses on P12 with Qwen3-4B-Base.} (i) the effect of dialectical reasoning on risk estimation (Section~\ref{sec:reasoning_structure}), (ii) the clinical correctness of the resulting rationales (Section~\ref{sec:case_study}), and (iii) their faithfulness to the predictions (Section~\ref{sec:faithfulness_hallucination}).
Appendix~\ref{app:human_study} complements (i) and (ii) with an expert evaluation by clinicians on rationale consistency, correctness, and helpfulness.

\subsection{Effectiveness of Dialectical Reasoning}
\label{sec:reasoning_structure}

\begin{wraptable}{r}{0.49\columnwidth}
\centering
\vspace{-2.75em}
\captionsetup{justification=centering}
\caption{
Ablation on the reasoning structure.
}
\vspace{-0.5em}
\label{tab:ablation_reasoning}
\small
\setlength{\tabcolsep}{4pt}
\begin{tabular}{lcc}
\toprule
Method & AUROC & AUPRC \\
\midrule
Zero-shot inference
& 69.7{\scriptsize$\pm$1.9} & 26.7{\scriptsize$\pm$0.8} \\
\midrule
\rowcolor{cyan!12}
\multicolumn{3}{l}{\textit{SFT baselines}} \\
Answer-only
& 86.4{\scriptsize$\pm$1.5} & 53.4{\scriptsize$\pm$2.2} \\
One-sided rationale$^{\dagger}$
& 83.8{\scriptsize$\pm$1.0} & 43.1{\scriptsize$\pm$3.6} \\
\midrule
\textbf{\ours}$_{\text{SFT}}$
& \textbf{86.9}{\scriptsize$\pm$1.0}
& \textbf{56.4}{\scriptsize$\pm$1.9} \\
\bottomrule
\addlinespace[2pt]
\multicolumn{3}{l}{\scriptsize
$^{\dagger}$ Risk estimated by averaging over 10 sampled responses.}
\end{tabular}
\vspace{-1.5em}
\end{wraptable}

To isolate the effect of reasoning structure, Table~\ref{tab:ablation_reasoning} compares the SFT checkpoint of \ours before RL refinement with zero-shot inference and two alternative SFT variants.
The \textit{answer-only} variant removes reasoning entirely and directly predicts the label, treating the LLM as a classifier.
The \textit{one-sided rationale} variant uses only the rationales corresponding to the ground-truth outcomes from our collected data.
Detailed prompts are in Appendix~\ref{app:prompts}.

\textit{Answer-only} SFT substantially outperforms zero-shot inference (Table~\ref{tab:ablation_reasoning}) but provides no clinical justification.
Adding \textit{one-sided rationales} introduces clinical reasoning but inherits the risk polarization problem identified in Section~\ref{sec:preliminary}, where the answer-token probability collapses toward 1.0.
Even after averaging predictions over 10 sampled outputs at $10\times$ inference cost, it underperforms \textit{answer-only} SFT on both metrics, indicating that one-sided supervision harms risk estimation.
In contrast, \ours's dialectical reasoning achieves the best AUROC (86.9\%) and AUPRC (56.4\%) while retaining clinically grounded explanations.
Beyond discrimination, clinicians rate the dialectical rationales as more helpful than one-sided ones ($+0.62$, $p<0.001$), with correctness unchanged (Appendix~\ref{app:human_study}).

\subsection{Case Studies of \ours Reasoning}
\label{sec:case_study}
We qualitatively examine whether \ours provides clinically coherent rationales that preserve temporal context.
Figures~\ref{fig:sample1} and~\ref{fig:sample2} present two illustrative cases.
For comparison, we apply Integrated Gradients (IG;~\citealp{IG}) post-hoc to STraTS, a strong ISMTS baseline on P12, and use GPT-5.1 to interpret the resulting textualized attributions.
The following interpretations are grounded in the medical literature and reviewed by two clinicians.

For the survivor (Figure~\ref{fig:sample1}), \ours treats the consistently maximal GCS of 15 as favorable and HCO$_3$ of 19\,mEq/L as adverse evidence, consistent with \emph{GCS 15 indicating intact consciousness}~\citep{munakomi2026glasgow} and \emph{low HCO$_3$ potentially indicating metabolic acidosis}~\citep{castro2024arterial}.
In contrast, IG attributions for STraTS carry the opposite sign for both.
\ours also tracks the urine output trajectory from 5--30\,mL/h to 300--400\,mL/h, consistent with \emph{rising urine output indicating renal recovery}~\citep{khwaja2012kdigo}, whereas STraTS surfaces only isolated late measurements.

For the non-survivor (Figure~\ref{fig:sample2}), \ours identifies severely elevated early potassium and elevated troponin T as adverse evidence~\citep{simon2025hyperkalemia,babuin2008elevated}, while recognizing the later decline in potassium to 4.3\,mEq/L as countervailing evidence.
Neither appears among the high-attribution IG features of STraTS, which instead place WBC 23.3 and early BUN 98--99\,mg/dL on the favorable side, despite \emph{the former representing marked leukocytosis}~\citep{mank2026leukocytosis} and \emph{the latter being markedly elevated and associated with ICU mortality}~\citep{arihan2018blood}.

We further evaluate the rationales on consistency, correctness, and helpfulness through human expert evaluation (Appendix~\ref{app:human_study}), with an LLM-as-a-judge evaluation as a scalable \textit{proxy} (Appendix~\ref{app:llm_as_a_judge}).

\subsection{Faithfulness Analysis of \ours Reasoning}
\label{sec:faithfulness_hallucination}

\begin{wrapfigure}{r}{0.49\columnwidth}
\centering
\vspace{-1.5em}
\includegraphics[width=\linewidth]{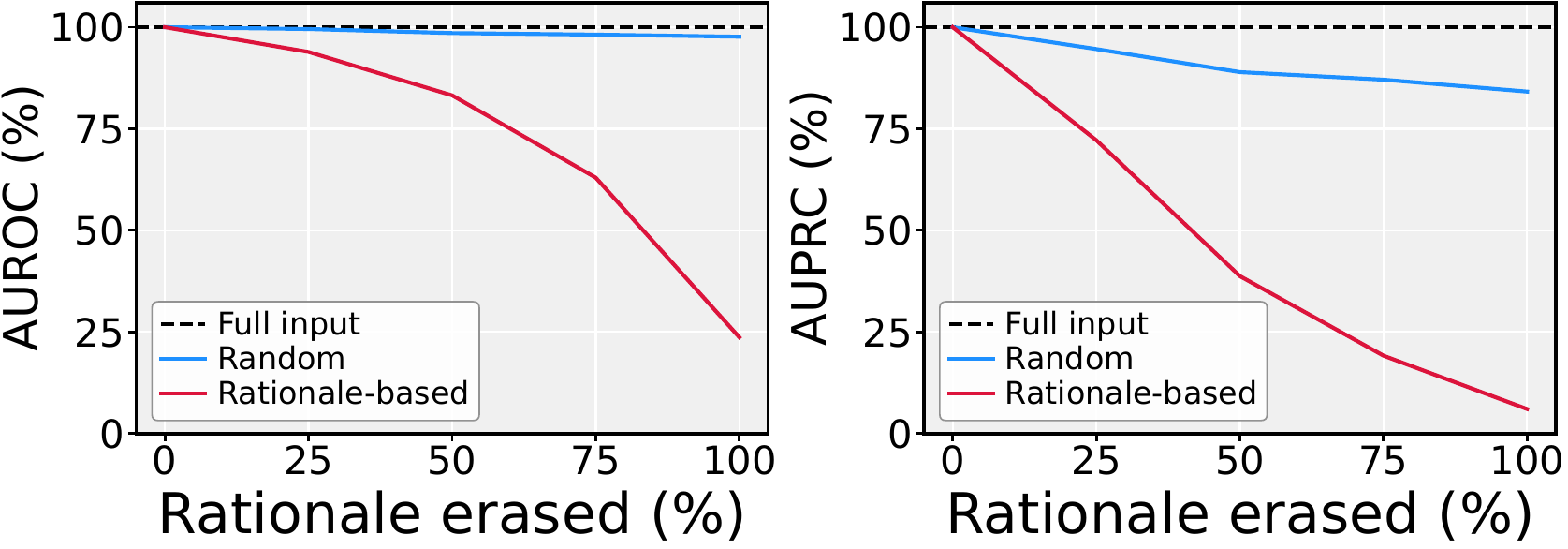}
\vspace{-1.5em}
\caption{AUROC (left) and AUPRC (right) against the full-input prediction on P12 as a function of the deletion ratio.}
\label{fig:faithfulness}
\vspace{-1.0em}
\end{wrapfigure}

We test whether the rationales of \ours are faithful, that is, whether the predicted risk depends on the evidence they cite, following the input-removal protocol of \citet{dynamask}.
For 1,000 patients, we randomly select a fraction of the sentences in the rationale for the predicted outcome, locate the measurements they cite with GPT-5.1, and erase them before re-running the model (\textit{Rationale-based}).
The control erases the same number of observed values at random (\textit{Random}), which separates reliance on the cited evidence from mere input sparsity.
Agreement with the full-input prediction is nearly unchanged under \textit{Random}, but falls monotonically under \textit{Rationale-based} and drops below chance at a deletion ratio of 100\% (Figure~\ref{fig:faithfulness}).

We also test whether the rationales stay grounded in the modified input by regenerating them after erasure.
At a deletion ratio of 50\%, a GPT-5.1 judge, blind to which sentences were targeted, rates how fully each sentence of the original rationale reappears in the regenerated rationale, from retained (1) to absent (5); the prompt is given in Appendix~\ref{app:prompts}.
Sentences whose evidence was erased average 4.63 (near absent), against 1.86 (near retained) for those left intact.
Together with the removal experiment, this indicates that the rationales are both faithful and grounded in the input.
A hallucination analysis further shows that only 1.3\% of rationales cite a feature or value absent from the record (Appendix~\ref{app:hallucination}).

%% file: sections/8_conclusion.tex
\section{Conclusion}
\label{sec:conclusion}

We have introduced \ours, a framework that grounds clinical risk estimation in reasoning over alternative outcomes rather than a single committed trajectory.
By inspecting each candidate outcome before deriving the final risk from the LLM's implicit probability, \ours addresses the risk polarization problem that arises when reasoning is elicited before risk scoring.
Through dialectical reasoning supervision followed by on-policy self-refinement, \ours outperforms competitive baselines while producing rationales that align with how clinicians weigh patient evidence.
More broadly, this work signals a path forward for LLM-based clinical decision support, where predictive performance and grounded explanation are pursued together rather than as competing objectives.

%% file: sections/9_acknowledgements.tex
\subsection*{Ethics statement}

\textbf{Data governance.}
MIMIC-III, MIMIC-IV, and eICU are accessed under the PhysioNet Credentialed Data Use Agreement.\footref{fn:physionet} Accordingly, no patient-level data from these datasets is sent to any third-party LLM service: preprocessing, rationale generation, training, and evaluation (including all zero-shot LLM baselines) are performed locally with self-hosted, open-weight models.\\\\
\textbf{Potential clinical risks.}
\ours is a research prototype and has not been validated for clinical
deployment.
Its natural language rationales may contain inaccurate clinical interpretations or reflect biases in the underlying data and models.
Over-reliance on these outputs without clinician oversight could lead to harm.
The outputs of \ours should not be treated as medical advice or used for autonomous clinical decision-making.
Any deployment should follow standard LLM safety practices, with clinical decisions remaining with qualified practitioners.

%% file: sections/appendix.tex
\section{Limitations: Computational Cost}
\label{app:cost}

\input{tables/app_cost_analysis}

The principal limitation of \ours is its inference cost.
We measure this cost as the mean wall-clock time per patient on a single B200 GPU.
For each method and dataset, we tune the batch size over $\{32, 128, 512, 2048\}$ (and the request concurrency for LLM-based methods) on the validation set and keep the setting with the lowest total time, so each row reflects the throughput-optimal configuration for that method.
We then time inference on the test set with this setting and divide the total time by the number of test patients to obtain the per-patient cost in Table~\ref{tab:cost_analysis}. GPT-5.1 is queried through its API: its entries are the mean latency of 10 synchronous calls, depend on server-side conditions we do not control, and are listed for reference only.

\begin{figure}[t]
\centering
\includegraphics[width=0.5\columnwidth]{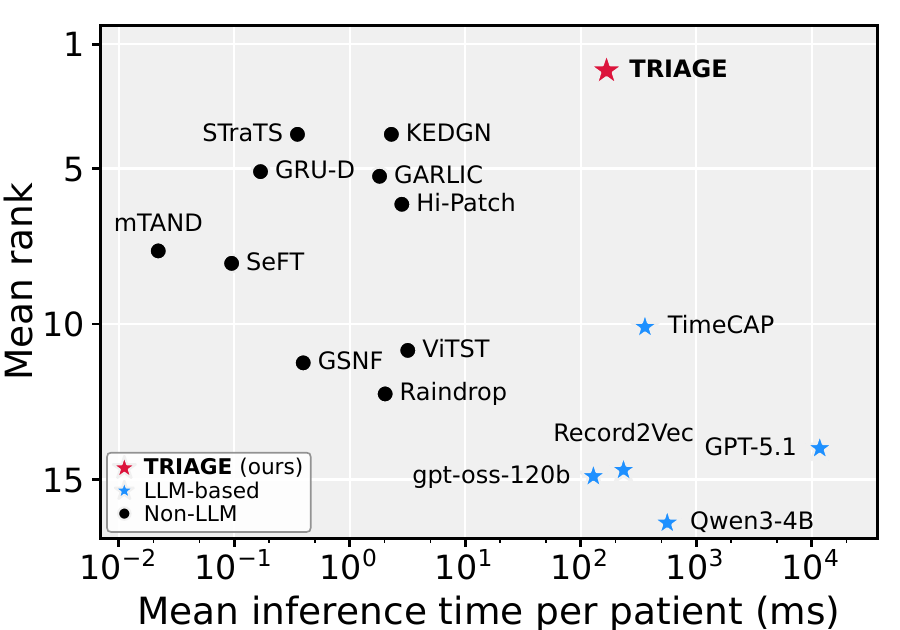}
\caption{Mean inference time per patient (ms, log scale) versus mean rank over the ten discrimination metrics. Lower is better on both axes.}
\label{fig:cost_rank}
\end{figure}

Figure~\ref{fig:cost_rank} plots this cost against each method's mean rank over the ten discrimination metrics (AUROC and AUPRC on five datasets) in Table~\ref{tab:main_results_full}.
\ours takes 167.4\,ms per patient, faster than both LLM-based prediction methods, Record2Vec (234.9\,ms) and TimeCAP (360.1\,ms), and faster than Qwen3-4B (561.6\,ms); among the locally run LLMs, only gpt-oss-120b (128.8\,ms) is faster. It also has a lower mean rank than every LLM-based baseline and every specialized ISMTS model.
The specialized ISMTS models are far cheaper, averaging under 4\,ms per patient; the fastest, mTAND, takes 0.022\,ms, nearly four orders of magnitude less than \ours.
We attribute most of this gap to the natural language explanation that \ours generates with each prediction, which the specialized models do not produce.
This overhead remains practical for ICU risk scoring, but it becomes a limitation in high-throughput screening, where many patients must be scored quickly, and in hospitals without dedicated GPU servers, since \ours requires serving a 4B-parameter LLM.

\section{Preliminary Study Details}
\label{app:det_preliminary}

\subsection{Implementation Details}

We conduct the preliminary experiment on in-hospital mortality prediction using the MIMIC-III~\citep{mimic3} test split ($n = 3{,}228$) with gpt-oss-120b~\citep{gptoss} and Qwen3-14B~\citep{qwen3}.
All runs use a temperature of 0.7.
For reasoning-then-answer, reasoning is enabled with \texttt{reasoning\textunderscore effort=medium} for gpt-oss-120b and \texttt{enable\textunderscore thinking=True} for Qwen3-14B.
We provide the full prompts in Appendix~\ref{app:prompts}.

\subsection{Implicit Probabilities as Risk Scores}
\label{app:prob_choice}

We consider two ways of eliciting probabilities from the LLM. The \emph{verbalized probability} $p^{\text{verb}} \in [0,1]$ is the numeric estimate in the model's response text, and the \emph{implicit probability} $P(y{=}1 \mid \text{prompt})$ is read off the next-token distribution over the answer tokens. We use the implicit probability as the risk score in the preliminary study (Section~\ref{sec:preliminary}). This choice is supported by \citet{gu2024probabilistic}, who report that, across six open-source LLMs and five medical datasets, verbalized probabilities give lower discrimination, precision, and recall than implicit probabilities, with larger gaps for smaller models and imbalanced datasets. Two further properties favor the implicit probability in our setting: it gives finer-grained scores, and it receives a direct training signal.

\begin{wraptable}{r}{0.49\textwidth}
\vspace{-1em}
\centering
\caption{
Concentration of risk scores on MIMIC-III in \emph{non-reasoning} mode.
Top-$k$ is the share (in \%) of outputs falling on the $k$ most frequent
values; \#Distinct counts unique values produced across all test samples.
}
\label{tab:anchor_concentration}
\vspace{-0.5em}
\scriptsize
\setlength{\tabcolsep}{4pt}
\resizebox{\linewidth}{!}{%
\begin{tabular}{lcccc}
\toprule
Method
& Top-1
& Top-3
& Top-7
& \#Distinct \\
\midrule
\rowcolor{cyan!12}
\multicolumn{5}{l}{\textit{gpt-oss-120b}} \\
Verbalized
& 37.0
& 55.0
& 75.3
& 55 \\
Implicit
& 2.3
& 6.0
& 10.5
& 1{,}375 \\
\addlinespace[2pt]
\rowcolor{cyan!12}
\multicolumn{5}{l}{\textit{Qwen3-14B}} \\
Verbalized
& 32.9
& 59.3
& 86.5
& 37 \\
Implicit
& 2.3
& 6.5
& 14.0
& 137 \\
\bottomrule
\end{tabular}%
}
\vspace{-1em}
\end{wraptable}

\paragraph{Granularity.} With reasoning disabled, verbalized outputs on MIMIC-III concentrate on a few anchor values (Table~\ref{tab:anchor_concentration}): seven values cover 75.3\% of gpt-oss-120b estimates and 86.5\% of Qwen3-14B estimates.
Patients who share one of these values receive the same score and cannot be ranked against each other. The implicit probability is continuous in $[0,1]$; on the same test set it takes 1{,}375 distinct values for gpt-oss-120b and 137 for Qwen3-14B, and its seven most frequent values cover only 10.5\% and 14.0\% of outputs.

\paragraph{Training signal.} During supervised fine-tuning, the cross-entropy loss on the answer token is the negative log of the implicit probability assigned to the true label, so the binary label supervises it directly for every patient.
The verbalized estimate has no direct target for supervised fine-tuning, since the data contain binary outcomes and no ground-truth probabilities, and using the label itself as the target number would teach the model to state only extreme values.
Reinforcement learning with a proper-scoring-rule reward can still train it, but the reward scores only the numbers the model samples, and training starts from the coarse, less discriminative estimates described above.

\begin{figure}[t]
    \centering
    \includegraphics[width=0.5\linewidth]{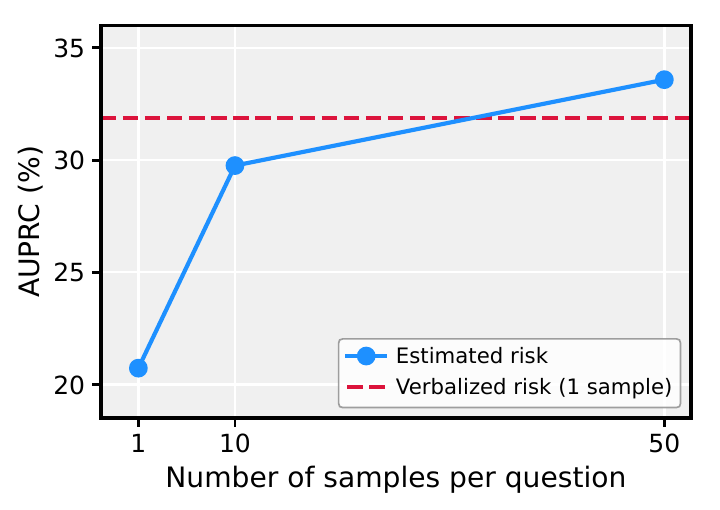}
    \vspace{-0.5em}
    \caption{AUPRC of the outcome-based and verbalized risk scores under reasoning, on MIMIC-III with gpt-oss-120b.}
    \label{fig:verbalized_efficiency}
\end{figure}

\subsection{Verbalized Probabilities for Zero-Shot Evaluation}
\label{app:verbalized_zeroshot}

For zero-shot LLM benchmarking under reasoning, two candidates are available for a graded risk score: (i) an \emph{outcome-based} estimator that approximates the model's underlying probability by averaging binary answers across many sampled responses, and (ii) the \emph{verbalized} probability stated in a single response. 
The outcome-based estimator surpasses the verbalized score only at around $30$ sampled responses per question, at a proportional inference cost (Figure~\ref{fig:verbalized_efficiency}). The verbalized probability is a clean single-number readout from a single response, providing a far more efficient zero-shot signal.
We therefore adopt the verbalized probability whenever reporting zero-shot LLM baselines in subsequent sections.

\subsection{One-Sided Reasoning Traces}
\label{app:one_sided}

We observe that LLM reasoning traces typically confirm a single outcome rather than weigh evidence on both sides. MIMIC-III is a credentialed dataset, so we do not reproduce raw patient records here. We instead draw an illustrative case from P12~\citep{p12}, which targets the same task as MIMIC-III (in-hospital mortality prediction from early ICU trajectories) and exhibits the same one-sidedness under our prompting protocol.

To make this behavior easy to inspect, we set \texttt{reasoning\textunderscore effort=None} for gpt-oss-120b and prompt the model with \textit{``Provide your clinical reasoning, then state your final prediction.''} This places the entire rationale in the visible response, where the one-sided pattern is straightforward to identify.

\begin{figure*}[t]
\centering
\vspace{-1.5em}
\includegraphics[width=\textwidth]{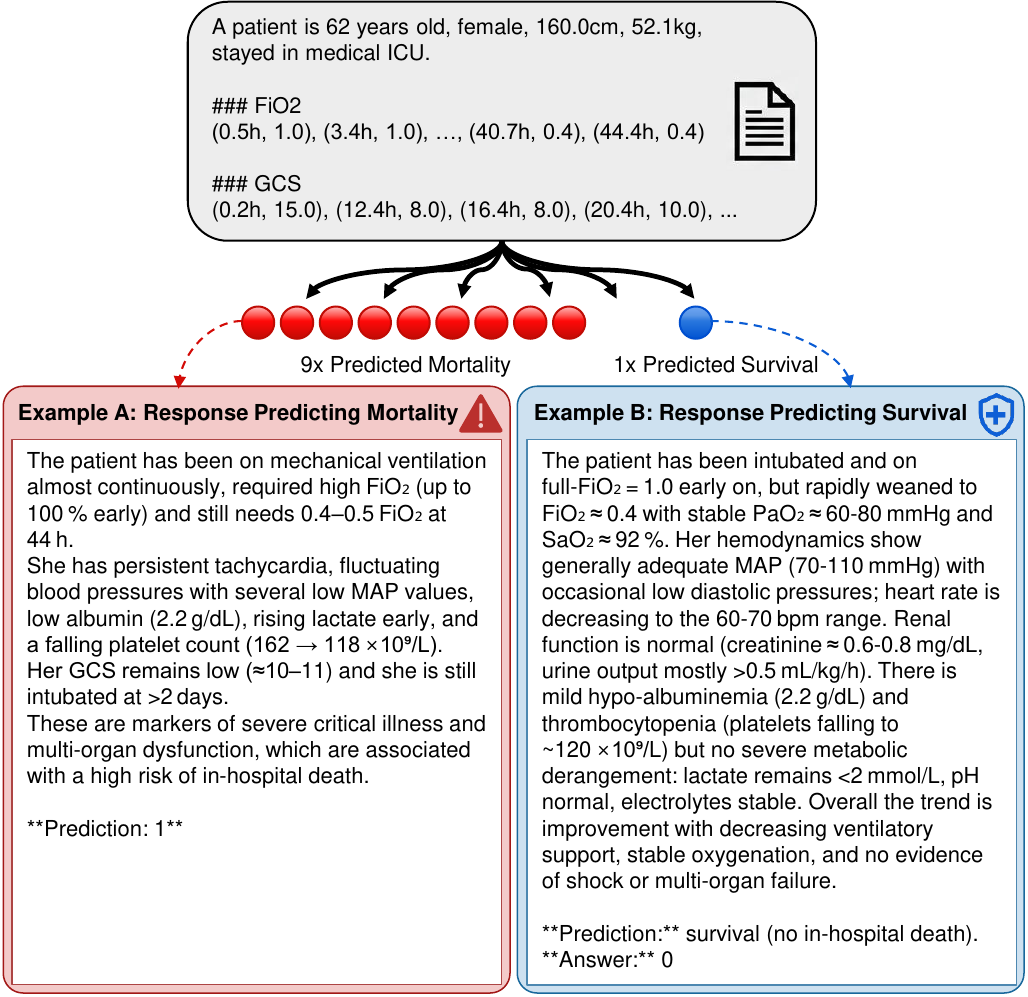}
\vspace{-1.5em}
\caption{
Two reasoning traces sampled from gpt-oss-120b on the same P12 patient. The mean implicit mortality probability over 10 samples for this patient is $0.9$. Each trace cites only one-sided evidence and commits to the corresponding outcome.
}
\vspace{-1.5em}
\label{fig:one_sided_cases}
\end{figure*}

Figure~\ref{fig:one_sided_cases} shows two reasoning traces out of ten sampled from gpt-oss-120b on the same patient. One trace cites only deterioration signals and predicts death; the other cites only stabilization signals and predicts survival. Each trace is internally coherent, yet neither weighs evidence from the opposite direction.

\section{Experimental Setup Details}
\label{app:exp_setup_details}

\subsection{Datasets}
\label{app:datasets}

We evaluate our method on five medical time series classification benchmarks: P12~\citep{p12}, P19~\citep{p19}, MIMIC-III~\citep{mimic3}, MIMIC-IV~\citep{mimic4}, and eICU~\citep{eicu}. Table~\ref{tab:dataset_stats} summarizes their statistics. All five are binary classification tasks with severe class imbalance.
Following~\citet{zhang2022raindrop, li2023vitst}, we use five random 8:1:1 splits for P12, P19, MIMIC-IV, and eICU. For MIMIC-III, we adopt the fixed 7:1.5:1.5 split defined by prior work~\citep{mimic_prep} and run five random seeds. In all cases, we report the mean and standard deviation across the five runs. We describe each dataset below, including our preprocessing refinements over the original releases.

\paragraph{P12.}
The P12 dataset~\citep{p12} is taken from the PhysioNet 2012 Challenge.\footnote{\url{https://physionet.org/content/challenge-2012/1.0.0/}}
Following~\citet{zhang2022raindrop, li2023vitst}, we exclude 12 inappropriate samples, leaving 11{,}988 ICU patient records.
Each record consists of irregularly sampled measurements from 36 sensors collected during the first 48 hours of an ICU stay, together with 5 static demographic features (Age, Gender, Height, ICUType, Weight).
The task is binary in-hospital mortality prediction. We adopt the training, validation, and test splits released by Raindrop~\citep{zhang2022raindrop}.

We also correct a parsing misalignment in the original preprocessing, where the five demographic features are extracted by row position rather than by parameter name. When a sensor reading at $t{=}00{:}00$ is interleaved with the demographic rows, the two streams contaminate each other: real sensor values are parsed as demographics, while later static fields such as weight are pushed into the time series. This affects 207 of 11{,}988 patients (1.7\%) and produces implausible values such as height~$=1$~cm.

\paragraph{P19.}
The P19 dataset~\citep{p19} is taken from the PhysioNet 2019 Challenge on sepsis prediction.\footnote{\url{https://physionet.org/content/challenge-2019/1.0.0/}}
Each sample contains 34 irregularly sampled time series variables observed for up to 60 hours, together with 6 static features (Age, Gender, Unit1, Unit2, HospAdmTime, ICULOS).
The task is binary classification of whether sepsis will occur within the next 6 hours. The public release used by prior work~\citep{zhang2022raindrop, li2023vitst, luo2024kedgn} comprises 38{,}803 ICU stays.

We further exclude 65 patients whose 34 time series variables contain no non-zero finite observation within their reported sequence length, yielding 38{,}738 samples. All 65 carry a positive sepsis label, so removing them lowers the positive rate from 4.19\% to 4.03\%. Such records provide no measurement signal for any time series classifier and only inflate the apparent positive rate. Because we deviate from the sample set used by prior work, we generate five new splits for P19.

\paragraph{MIMIC-III.}
MIMIC-III~\citep{mimic3} is a publicly released critical care database distributed via PhysioNet.\footnote{\url{https://physionet.org/content/mimiciii/1.4/}}
We adopt a benchmark preprocessing pipeline~\citep{mimic_prep} and target binary in-hospital mortality prediction over the first 48 hours after ICU admission. The resulting dataset comprises 21{,}107 patient stays, each described by 16 time series variables and one static feature (Height).

\paragraph{MIMIC-IV.}
MIMIC-IV~\citep{mimic4} is a critical care database of sparse, irregularly sampled physiological records collected at the Beth Israel Deaconess Medical Center between 2008 and 2019.\footnote{\url{https://physionet.org/content/mimiciv/1.0/}} We follow the preprocessing of GSNF~\citep{gao2026one}, extracting 96 features (patient intake and output, laboratory results, and medication prescriptions) from the first 48 hours of recorded measurements, with binary in-hospital mortality as the target. A total of 26{,}070 hospital admissions are retained. Unlike P12, P19, and MIMIC-III, this cohort carries no static features, as the preprocessing produces none. 

\paragraph{eICU.}
eICU~\citep{eicu} contains records from patients admitted to intensive care units across 208 hospitals in the United States between 2014 and 2015.\footnote{\url{https://physionet.org/content/eicu-crd/2.0/}} We follow GSNF~\citep{gao2026one}, extracting 14 features from the first 48 hours after ICU admission for the same in-hospital mortality task. A total of 12{,}312 ICU stays are retained, each belonging to a distinct patient, and, as for MIMIC-IV, no static features are used.

\begin{table*}[t]
\centering
\setlength{\tabcolsep}{4.5pt}
\caption{Statistics of the five medical time series classification datasets. \#Avg.~obs.\ denotes the average number of observed measurements per sample. The missing ratio is computed with respect to each patient's actual observation length. Pos. rate denotes the proportion of positive samples.}                               
\label{tab:dataset_stats}
\begin{tabular}{lccccc}\toprule                                       
Dataset & \#Samples & \#Variables & \#Avg.~obs. & Missing ratio & Pos.\ rate\\
\midrule        
P12       & 11{,}988 & 36 & 401 & 84.9\% & 14.2\% \\
P19       & 38{,}738 & 34 & 238 & 80.5\% & 4.0\%  \\
MIMIC-III & 21{,}107 & 16 & 431 & 65.5\% & 13.2\% \\
MIMIC-IV & 26{,}070 & 96 & 341 & 98.0\% & 13.4\%  \\
eICU & 12{,}312 & 14 & 57 & 65.2\% & 17.6\%  \\
\bottomrule
\end{tabular}
\end{table*}

\subsection{Metrics}
\label{app:metrics}
AUROC and AUPRC follow their standard definitions. The Brier score is the mean squared error between the predicted probability and the binary outcome. ECE \citep{ece} measures the average gap between predicted confidence and empirical accuracy across $M=10$ equal-width bins on $[0,1]$. For both the Brier score and ECE, lower values indicate better calibration.

\subsection{Baselines}
\label{app:baselines}

We employ three categories of baselines: zero-shot LLMs, ISMTS baselines, and LLM-based prediction methods.

For the zero-shot LLMs, we use GPT-5.1~\citep{openai2025gpt51systemcard} and gpt-oss-120b~\citep{gptoss}, both with \texttt{reasoning\textunderscore effort=medium}, and Qwen3-4B~\citep{qwen3} with \texttt{enable\textunderscore thinking=True}.
These include one proprietary model and two open-weight models of different sizes.
Based on the preliminary study in Appendix~\ref{app:verbalized_zeroshot}, we use the verbalized probability produced by the model as the predicted risk.
The prompts used for zero-shot evaluation are provided in Appendix~\ref{app:prompts}.

For all ISMTS baselines, we use Adam as the default optimizer and follow each baseline's released code for model-specific training settings, including the optimizer where the released code uses a different one.
To mitigate class imbalance, we follow the training protocol of~\citet{zhang2022raindrop,li2023vitst,luo2024kedgn} and upsample the minority class threefold while constraining each mini-batch to a 1:1 class ratio. 
Because AUPRC is the primary metric for irregular time series classification under class imbalance~\citep{davis2006relationship, saito2015precision}, we use validation AUPRC as the model selection criterion.
Unless otherwise specified, each model is trained for up to $50$ epochs with an early stopping patience of $5$. We evaluate each hyperparameter configuration over $5$ runs and select the configuration with the highest validation AUPRC averaged over these runs. Method-specific hyperparameter configurations and search spaces are described below.

\paragraph{GRU-D~\citep{che2018recurrent}.}
We reproduce the original authors' Keras implementation\footnote{\url{https://github.com/PeterChe1990/GRU-D}} in PyTorch with the original MLP classification head.
Following the released code, we fix the input and recurrent dropout to $0.3$.
We tune the hidden dimension over $\{32, 64, 128, 256, 512, 1024\}$, the learning rate over $\{5\text{e-}4, 1\text{e-}3, 5\text{e-}3, 1\text{e-}2\}$, and the batch size over $\{32, 64, 128, 256, 512\}$.

\paragraph{mTAND~\citep{shukla2021mtand}.}
We follow the official implementation\footnote{\url{https://github.com/reml-lab/mTAN}} and adopt the mTAND-Full classification variant.
Following the released code, we fix the number of reference time points to $128$, the time-embedding dimension to $128$, the number of attention heads to $1$, and the generator hidden size to $50$.
We tune the recurrent encoder hidden size over $\{32, 64, 128\}$, the latent dimension over $\{32, 64, 128\}$, the classification-loss weight over $\{1, 5, 10, 50, 100\}$, the learning rate over $\{5\text{e-}4, 1\text{e-}3, 5\text{e-}3, 1\text{e-}2\}$, and the batch size over $\{32, 64, 128\}$.

\paragraph{SeFT~\citep{horn2020seft}.}
We reproduce the authors' TensorFlow/Keras implementation\footnote{\url{https://github.com/BorgwardtLab/Set_Functions_for_Time_Series}} in PyTorch.
Following the values dominant across datasets in the original paper's hyperparameter table, we fix the observation encoder to a $4$-layer MLP, the attention aggregator to $2$ layers with $4$ heads, and the output readout to a $2$-layer MLP of width $512$.
We tune the per-observation embedding dimension over $\{256, 512\}$, the observation encoder MLP width over $\{64, 128\}$, the shared dropout rate over $\{0.0, 0.2, 0.4\}$, the time-encoding dimension over $\{8, 16\}$, the maximum time scale over $\{100, 1000\}$, the learning rate over $\{1\text{e-}4, 5\text{e-}4, 1\text{e-}3, 5\text{e-}3\}$, and the batch size over $\{32, 64, 128\}$.

\paragraph{Raindrop~\citep{zhang2022raindrop}.}
We follow the official implementation.\footnote{\url{https://github.com/mims-harvard/Raindrop}} Following the original paper, we fix the batch size to $128$.
We tune the per-sensor observation embedding dimension over $\{2, 4, 8\}$, the number of Raindrop layers over $\{1, 2, 4\}$, the number of attention heads over $\{1, 2, 4\}$, dropout over $\{0.0, 0.2, 0.4\}$, and the learning rate over $\{1\text{e-}4, 5\text{e-}4, 1\text{e-}3, 5\text{e-}3\}$.

\paragraph{STraTS~\citep{tipirneni2022strats}.}
We follow the official PyTorch implementation.\footnote{\url{https://github.com/sindhura97/STraTS}}
Following the released code and paper, we use AdamW with the default weight decay of $0.01$ and fix the learning rate to $5\text{e-}4$, the batch size to $32$, and the per-patient observation cap to $880$.
We tune the hidden dimension over $\{32, 64, 128\}$, the number of Transformer layers over $\{1, 2, 4\}$, the number of attention heads over $\{8, 16, 32\}$, and the shared dropout rate (used for both feature and attention dropout) over $\{0.0, 0.2, 0.4\}$.

\paragraph{ViTST~\citep{li2023vitst}.}
We follow the official implementation\footnote{\url{https://github.com/Leezekun/ViTST}} and adopt the Vision-Text variant, which combines a Swin Transformer image encoder with a RoBERTa text encoder; both encoders are fine-tuned end-to-end.
Following the original paper, we render the time series inputs as line-graph images at a resolution of $384\times384$. We use dataset-specific grid layouts according to the number of input variables: $6\times6$ for P12 and P19, $4\times4$ for MIMIC-III and eICU, and $12\times12$ for MIMIC-IV.
We set the maximum number of training epochs to $4$ for P12, MIMIC-III, MIMIC-IV, and eICU, and to $2$ for P19, with an effective batch size of $48$.
We tune the learning rate over $\{1\text{e-}5, 2\text{e-}5, 5\text{e-}5, 1\text{e-}4, 2\text{e-}4, 5\text{e-}4\}$ and select both the learning rate and the checkpoint based on validation AUPRC.

\paragraph{KEDGN~\citep{luo2024kedgn}.}
We follow the official implementation.\footnote{\url{https://github.com/easonLuo2001/KEDGN}}
Following the released code, we fix the batch size to $512$, $512$, and $256$ and the learning rate to $1\text{e-}3$, $5\text{e-}3$, and $5\text{e-}3$ for P12, P19, and MIMIC-III, respectively, and use the released ChatGPT-extracted textual descriptions.
For MIMIC-IV and eICU, we fix the batch size to $256$, tune the learning rate over $\{1\text{e-}3, 5\text{e-}3\}$, and generate the textual descriptions with GPT-5.6 Sol using the prompt provided in the original paper.
We use GPT-5.6 Sol in place of ChatGPT because the ChatGPT version used in the original paper is no longer available.
The model is used only to generate these variable descriptions, and its input consists of the prompt and the variable names.
At no point are raw measurement values or any other patient-level data from MIMIC-IV or eICU sent to the model.
On all datasets, the textual descriptions are encoded by BERT.
We tune the dimension of the query vectors over $\{5, 7, 9\}$, the dimension of the variable node embeddings over $\{7, 9, 11\}$, the proportion of the density score over $\{1.0, 2.0, 3.0\}$, and the dimension shared by the variable hidden states and the structured encoding representations over $\{8, 12, 16\}$.

\paragraph{Hi-Patch~\citep{luo2025hipatch}.}
We follow the official implementation.\footnote{\url{https://github.com/easonLuo2001/Hi-Patch}}
Following the released code, we train each model for up to $20$ epochs and fix the time-decay coefficient to $1$, the learning rate to $1\text{e-}3$, and the batch size to $8$, $64$, and $16$ for P12, P19, and MIMIC-III, respectively.
For MIMIC-IV and eICU, we fix the batch size to $16$.
We tune the hidden dimension over $\{16, 32, 64, 128\}$, the number of attention heads over $\{1, 2, 4, 8\}$, the number of graph-attention layers over $\{1, 2\}$, and the patch size over $\{T/2, T/4, T/8, T/16, T/32\}$, where the history window $T$ is $60$ hours for P19 and $48$ hours for the other datasets.

\paragraph{GSNF~\citep{gao2026one}.}
We follow the official implementation.\footnote{\url{https://github.com/mzgaooo/GSNF}}
Since the original paper reports hyperparameters for every dataset it evaluates, we adopt them without tuning: the optimizer is Adam with a weight decay of $1\text{e-}4$, the batch size is $50$, the learning rate is $1\text{e-}3$ decayed by a factor of $0.5$ every $20$ epochs, the number of GSNF layers is $2$, the number of hidden layers is $3$, the hidden dimension is $128$, the cross-entropy weight is $1000$, and the weights of the interaction-aware trajectory generation (ITG) and reverse-time trajectory generation (RTG) losses are both $0.1$; the released code sets the latent dimension to the number of variables.
The separation margin is dataset-specific: $1\text{e-}5$ for P12, P19, and eICU, and $1\text{e-}7$ for MIMIC-IV.
MIMIC-III is not evaluated in the original paper, so on this dataset we tune the separation margin over its candidate set $\{1\text{e-}3, 1\text{e-}4, 1\text{e-}5, 1\text{e-}6, 1\text{e-}7, 1\text{e-}8\}$ and leave all other hyperparameters unchanged.
As in the original paper, predictions are taken from the model's own classification head.

\paragraph{GARLIC~\citep{li2026garlic}.}
We follow the official implementation.\footnote{\url{https://github.com/SCAI-Lab/GARLIC}} Following the released code, we train each model for up to $100$ epochs with an early stopping patience of $10$. We switch between the reconstruction and prediction stages whenever the validation loss has not improved for more than $3$ epochs, reloading the checkpoint with the lowest validation loss at each switch. Since AUPRC is our primary evaluation metric, we select the final model based on the best validation AUPRC. We fix the reconstruction learning rate to $1\text{e-}3$, the graph-regularization weight to $0.01$, and both weight-decay coefficients to $1\text{e-}4$. We tune the lag window over $\{3, 5, 7, 9\}$, the batch size over $\{64, 128, 256\}$, the hidden dimension over $\{128, 512\}$, the feature dimension over $\{16, 32\}$, the classification learning rate over $\{5\text{e-}5, 1\text{e-}4\}$, and the classification-loss weight over $\{0.1, 10\}$ for all datasets.

\medskip\noindent
For the LLM-based prediction methods, Record2Vec and TimeCAP, we fix the LLM to Record2Vec's default summarizer, medgemma-27b-text-it~\citep{sellergren2025medgemma}; both methods also convert each record into text with the missingness-aware serialization of the Record2Vec code.
Training follows the ISMTS protocol above, except that TimeCAP keeps the loss defined in its released code and uses neither the minority-class upsampling nor the 1:1 mini-batch ratio.

\paragraph{Record2Vec~\citep{record2vec}.}

We follow the official implementation.\footnote{\url{https://github.com/Jerryji007/Record2Vec-ICLR2026}}
The LLM summarizes each serialized record once, and the summaries are embedded with Qwen3-Embedding-8B~\citep{qwen3embedding}.
Following the released code, we set the batch size to $512$, the maximum number of epochs to $200$, and the early stopping patience to $8$.
We tune the classification head over $\{\text{MLP}, \text{LSTM}, \text{PatchTSMixer}\}$ and the learning rate over $\{1\text{e-}5, 5\text{e-}5, 1\text{e-}4, 5\text{e-}4\}$.

\paragraph{TimeCAP~\citep{lee2025timecap}.}

We follow the official implementation.\footnote{\url{https://github.com/geon0325/TimeCAP}}
The LLM contextualizes each serialized record once, and the resulting text is encoded by BERT~\citep{devlin2019bert}\footnote{\url{https://huggingface.co/google-bert/bert-base-cased}}, which is fine-tuned end-to-end. 
Following the released code, we fix the model width to $64$ and the feed-forward width to $128$ and train for up to $100$ epochs with an early stopping patience of $10$.
We tune the number of attention layers over $\{1, 2\}$, the number of attention heads over $\{4, 8, 16\}$, the dropout rate over $\{0.0, 0.1, 0.2\}$, and the learning rate over $\{1\text{e-}4, 5\text{e-}4, 1\text{e-}3\}$.
The remaining settings are adapted to the scale and irregular sampling of our datasets: we fix the patch length to $8$, the stride to $4$, and the batch size to $64$; replace instance normalization with per-variable standardization using training-fold statistics; and fill missing values by last observation carried forward.
Retrieval uses $5$ neighbors without class stratification, and the weight that combines the predictions of the multi-modal encoder and the LLM is selected on the validation set of each dataset.

\subsection{Implementation Details}
\label{app:impl}

\paragraph{Data collection.}

We collect outcome-specific rationales using GPT-5.1 for P12 and P19, and Kimi K2 Thinking~\citep{kimi} for MIMIC-III, MIMIC-IV, and eICU.
We first collect rationales \emph{separately} for each candidate outcome. For each candidate, the LLM is prompted to assume that outcome and to identify only the patient features that support it. 
To keep each rationale strictly outcome-specific rather than contrastive, we prohibit the LLM from referencing any alternative outcome or fabricating unobserved evidence; when no supporting evidence appears, the rationale is left blank. These constraints are intended to keep the rationales concise and grounded in observed features, without hallucinated or contrastive content.

To mitigate class imbalance, we follow prior work~\citep{zhang2022raindrop, li2023vitst, luo2024kedgn} and oversample the minority class during data collection. For each minority class sample, we collect three rationales per candidate outcome on P12, MIMIC-III, MIMIC-IV, and eICU, and six on P19, whose positive rate is lower. Majority class samples receive one rationale per candidate outcome. The full set of prompts is provided in Appendix~\ref{app:prompts}.

\input{tables/exp_setup_details_sft_rl}

\paragraph{SFT.}
Following prior work~\citep{zhang2022raindrop, li2023vitst, luo2024kedgn}, we employ a custom sampler that maintains a 1:1 minority-to-majority ratio within each batch.
All SFT experiments are conducted on a single H200 GPU. The full set of hyperparameters is listed in Table~\ref{tab:sft_experimental_setup_details}.

\paragraph{RL.}

For RL initialization, we select the best SFT checkpoint up to two epochs rather than the best checkpoint over all three SFT epochs, as this empirically yields better RL performance and aligns with recent observations that high SFT scores may not predict better post-RL performance~\citep{kang2026quagmires}.
Following Dr. GRPO~\citep{drgrpo} and DAPO~\citep{dapo}, we compute advantages without standard-deviation normalization and use a token-level policy gradient loss.
For KL regularization, we adopt the $k_2$-as-loss formulation~\citep{liu2025rethinking}.

To address class imbalance, the advantage is scaled by a factor of 2 for minority class samples and by $\frac{2}{N}$ for majority class samples, where $N:1$ denotes the ratio of the majority to minority class size in the full training set. In particular, in the case of the P19 data, the minority class samples are oversampled by a factor of three due to severe class imbalance.
We also employ a custom stratified sampler to ensure that each rollout batch, rather than the PPO mini-batch, matches the class ratio of the full training set.

All RL experiments are implemented on top of \texttt{verl}~\citep{sheng2024hybridflow} and performed on 2$\times$ B200 GPUs. The full set of training hyperparameters is in Table~\ref{tab:rl_experimental_setup_details}. We use $m=1.0$ as the margin in the reward and $\lambda=0.25$ as the weight of the CE loss in Equation~\ref{eq:entire_loss}.
Note that the batch size is set to 256, which is larger than typical configurations, in order to ensure that our batch-level reward operates stably.

\paragraph{Inference.}

We use a fixed sampling temperature of $0.7$ during inference. Because the model is trained on both rationale orderings in Equation~\ref{eq:chain} to improve generalization, we likewise perform inference under each.
For each pass, we apply a softmax to the logits of $y^+$ and $y^-$ at the $\hat{y}$ position and take the resulting probability of $y^+$ as the risk score.
The patient's final risk score is defined as the mean of these two probabilities.
A comparison between this bidirectional averaging and using each ordering individually is provided in Appendix~\ref{app:direction_ablation}.

\input{tables/main_result_full}

\section{Full Results on Discrimination}
\label{app:full_results}

Table~\ref{tab:main_results_full} reports the results of all methods, including the ISMTS baselines omitted from Table~\ref{tab:main_results_compact}. It also shows how the natural-language-explainable methods perform relative to the ISMTS baselines, which cannot explain their predictions in natural language.

\section{Robustness under Limited Information}
\label{app:robustness}

\begin{figure*}[t]
\centering
\includegraphics[width=\textwidth]{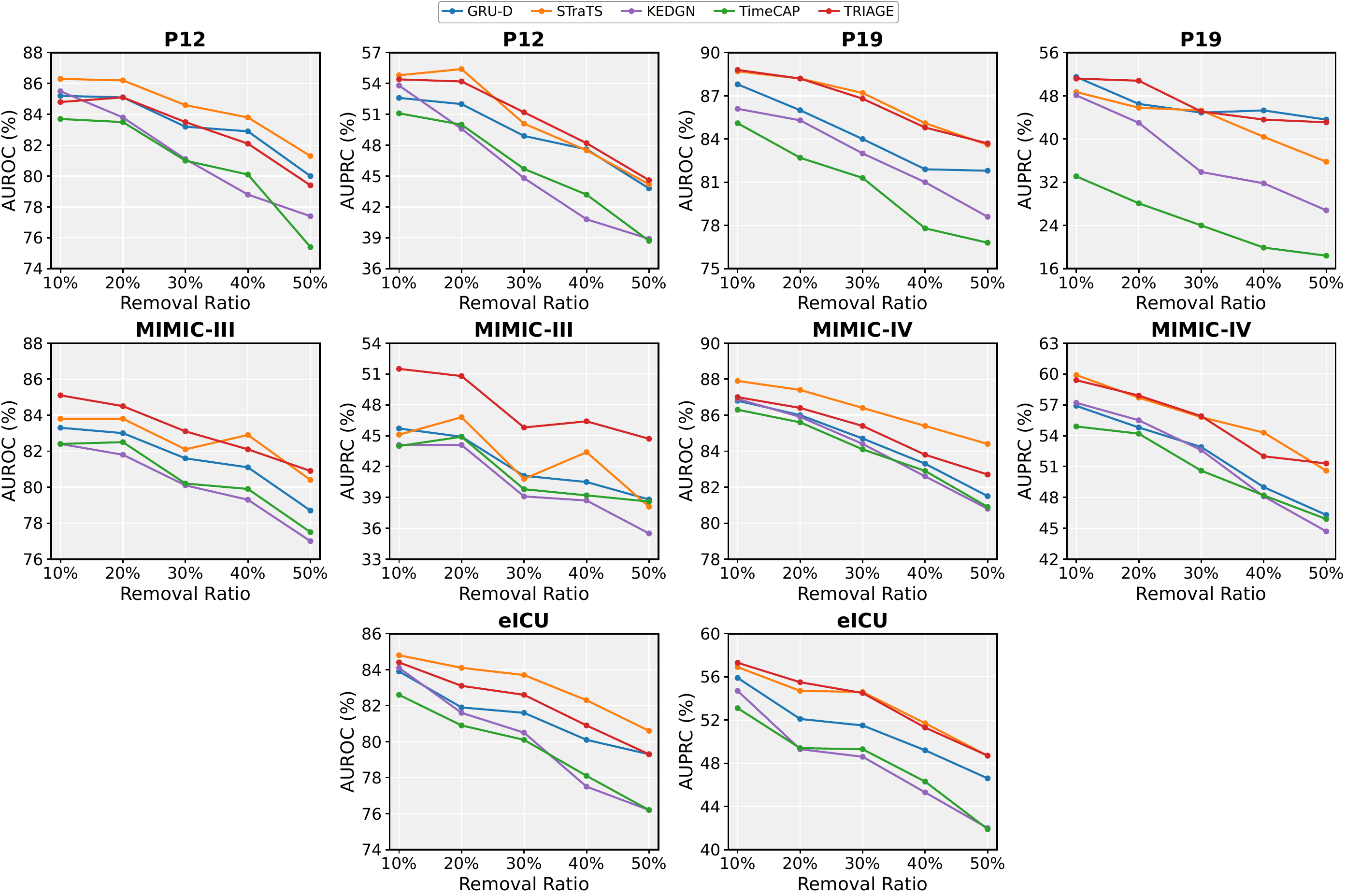}
\vspace{-1.5em}
\caption{
Performance comparison under the leave-random-variables-out setting on five ISMTS benchmarks.
}
\label{fig:lvo}
\end{figure*}

\subsection{Leave-Random-Variables-Out}  
\label{app:lvo}

In practice, clinical risk predictors rarely observe every variable at inference time due to sensor failures, protocol differences, or missing records.
Prior work has therefore evaluated ISMTS models under \emph{variable-level} missingness, where a subset of variables is entirely removed at inference~\citep{zhang2022raindrop, li2023vitst, luo2024kedgn}.
We randomly remove 10\%, 20\%, 30\%, 40\%, and 50\% of the variables and evaluate the discrimination performance to assess the robustness of \ours. We compare \ours against the three top-performing ISMTS models, GRU-D, KEDGN, and STraTS, and the LLM-based method TimeCAP. 

Across the five benchmarks, \ours attains the best AUPRC on 15 out of 25 settings, while outperforming TimeCAP on all 50 settings (Figure~\ref{fig:lvo}). Crucially, \ours is the least sensitive to variable removal: averaged over the ten dataset--metric pairs, its performance drops by 6.6 points (10.4\% relative) as the removal ratio increases from 10\% to 50\%, compared with 6.9 points for GRU-D (11.3\%), 6.9 points for STraTS (11.6\%), 10.5 points for KEDGN (17.8\%), and 8.6 points for TimeCAP (15.8\%).
These results indicate that \ours does not depend on a specific subset of variables and retains its predictive signal when the observed variable set is substantially reduced.

\subsection{Low-Resource Training}
\label{app:few-shot}

Clinical applications frequently suffer from limited labeled data, particularly for rare conditions and newly emerging diseases. These regimes favor models that can draw on pretrained medical and reasoning priors. Unlike ISMTS baselines trained from scratch, \ours builds on a pretrained LLM and inherits such priors directly, without requiring large amounts of task-specific supervision.

\paragraph{Setup.}
To probe this advantage under data scarcity, we fine-tune \ours with rsLoRA~\citep{rslora}, which preserves pretrained representations more reliably than full fine-tuning in low-data regimes. We use rank $r=64$, scaling factor $\alpha=64$, and dropout $0.05$, applying adapters to all linear layers. We omit the RL stage to isolate the data efficiency of supervised reasoning alone. Only the P12 training set is subsampled, to 1\% and 5\% of its original size; the validation and test sets are kept intact for consistent evaluation.

\paragraph{Baseline tuning.}
For a fair comparison, we start from the hyperparameters of the full-data setting and additionally sweep smaller batch sizes and a broader range of learning rates. This ensures that observed differences reflect data efficiency rather than under-tuned baselines.

\begin{wraptable}{r}{0.49\columnwidth}
   \centering
   \vspace{-1.0em}
   \caption{Results on P12 under low-resource training. Only the training set is subsampled. The best and second-best results are indicated in \textbf{bold} and \underline{underline}, respectively.}
   \label{tab:lrt}
   \small
   \setlength{\tabcolsep}{2.5pt}
   \begin{tabular}{lcccc}
      \toprule
      & \multicolumn{2}{c}{P12-1\%} & \multicolumn{2}{c}{P12-5\%} \\
      \cmidrule(lr){2-3} \cmidrule(lr){4-5}
      Method & AUROC & AUPRC & AUROC & AUPRC \\
      \midrule
      GRU-D   & 74.5{\scriptsize$\pm$4.2} & 35.3{\scriptsize$\pm$5.3} & \textbf{82.3}{\scriptsize$\pm$1.1} & \underline{48.0}{\scriptsize$\pm$2.0} \\
      STraTS  & 71.8{\scriptsize$\pm$5.2} & 29.8{\scriptsize$\pm$7.3} & 80.8{\scriptsize$\pm$2.0} & 42.5{\scriptsize$\pm$3.5} \\
      KEDGN   & 69.4{\scriptsize$\pm$5.7} & 28.1{\scriptsize$\pm$8.0} & 80.8{\scriptsize$\pm$1.5} & 45.5{\scriptsize$\pm$2.3} \\
      TimeCAP & \textbf{76.4}{\scriptsize$\pm$1.6} & \underline{37.9}{\scriptsize$\pm$3.2} & 80.3{\scriptsize$\pm$1.7} & 45.5{\scriptsize$\pm$2.5} \\
      \midrule
      \textbf{\ours}$_{\text{SFT}}$ & \underline{76.2}{\scriptsize$\pm$4.8} & \textbf{40.0}{\scriptsize$\pm$5.6} & \textbf{82.3}{\scriptsize$\pm$1.1} & \textbf{49.0}{\scriptsize$\pm$3.6} \\
      \bottomrule
   \end{tabular}
   \vspace{-1.0em}
\end{wraptable}

\paragraph{Results.}
We compare \ours against GRU-D, STraTS, and KEDGN, the three ISMTS models with the best average performance in our main experiments, and the LLM-based method TimeCAP (Table~\ref{tab:lrt}).
LLM-based methods are less sensitive to the reduced label budget: from P12-5\% to P12-1\%, \ours and TimeCAP lose 9.0 and 7.6 AUPRC, while GRU-D, STraTS, and KEDGN lose 12.7, 12.7, and 17.4, respectively. This is consistent with the hypothesis that pretrained knowledge contributes most when supervision is scarce.
TimeCAP, however, plateaus at a lower operating point, and \ours surpasses it in AUPRC at both budgets, by 5.5\% and 7.7\% relative at P12-1\% and P12-5\%, respectively.
Against GRU-D, the strongest of these ISMTS baselines, the margin is larger when labels are scarcer: \ours improves AUPRC by 13.3\% relative at P12-1\% and by 2.1\% relative at P12-5\%.

\section{Additional Analysis of \ours Reasoning}
\label{app:additional_analysis}

\subsection{Human Expert Evaluation}
\label{app:human_study}

\paragraph{Protocol.} We ask two clinicians to conduct an expert evaluation of the reasoning generated by \ours.
From the P12 test split, we draw 30 ICU stays (20 survivors, 10 non-survivors) and present four rationales for each case, giving 120 reasoning chains in total: (i) \ours; (ii) a \emph{one-sided rationale} described in Section~\ref{sec:reasoning_structure};
(iii) zero-shot GPT-5.1; and (iv) the textualized Integrated Gradients (IG;~\citealp{IG}) attributions over STraTS~\citep{tipirneni2022strats} described in Section~\ref{sec:case_study}.
The IG baseline is truncated to the length of the corresponding \ours rationale so that verbosity cannot by itself drive the ratings.
Each rationale is rated on consistency, correctness, and helpfulness using the 1--5 rubric (Figure~\ref{fig:human_eval_criteria}).
We treat the case as the unit of analysis, averaging the two raters, and test differences with two-sided Wilcoxon signed-rank tests.
Table~\ref{tab:human_eval} reports the results. 

\input{tables/human_study_results}

\paragraph{Dialectical reasoning is preferred, specifically for decision support.}
Against the one-sided reasoning, \ours is indistinguishable on consistency ($+0.12$, $p=0.19$) and correctness ($+0.03$, $p=0.66$) but clearly higher on helpfulness ($+0.62$, 95\% CI $[+0.35,+0.88]$, $d_z=0.80$, $p<0.001$).
This isolates the dialectical format's contribution to perceived utility.
Displaying the competing outcome states which findings argued against the prediction and why they were outweighed, which is the information a clinician needs to audit the model's judgment, and which the one-sided rationale removes. Together with the discrimination gains of
Table~\ref{tab:ablation_reasoning}, this indicates that reasoning over competing outcomes improves the interpretability and real-world helpfulness it was introduced to provide.

\paragraph{Attribution-based and frontier-model explanations.}

Textualized IG scores lowest on all three criteria, and its deficit is larger on helpfulness ($2.48$) than on consistency ($3.28$) or correctness ($3.43$).
The attributions report which measurements moved the prediction, but they do not state the physiology that connects them, and the clinician is left to reconstruct the inferential step.
This is consistent with prior reports that saliency-style explanations transfer poorly to clinical use~\citep{hou2025explainable}.
Zero-shot GPT-5.1 receives the highest ratings overall ($4.68$ against $4.48$), by a small margin that is significant on two of the three criteria (consistency $-0.17$, $p=0.18$; correctness $-0.25$, $p<0.01$; helpfulness $-0.18$, $p<0.05$).
We attribute the remaining gap to capacity, since \ours uses a 4B backbone.
Crucially, these ratings do not carry over to risk estimation: the same GPT-5.1 configuration reaches $49.3$ AUPRC on P12 against $59.0$ for \ours (Table~\ref{tab:main_results_compact}).
A rationale that accompanies a weaker prediction is of limited clinical use, so the comparison that matters is whether reasoning quality can be obtained together with calibrated discrimination at a deployable model scale.

\paragraph{Inter-rater agreement.}
Over the 120 rationales, ICC(A,1) between the two clinicians is $0.61$ for helpfulness, $0.35$ for consistency, and $0.11$ for correctness.
The low value for correctness likely reflects, in part, a strong ceiling effect, since one rater gave the maximum score to 87.5\% of rationales (Gwet's AC2 $=0.59$).
The raters also differ in strictness (mean $3.72$ vs.\ $4.51$), but z-scoring each rater leaves the ordering of methods unchanged on every criterion.
Both raters assign higher helpfulness scores to \ours than to the one-sided rationale ($+0.93$, $p<0.001$; $+0.30$, $p=0.14$), whereas the advantage of GPT-5.1 comes from one rater alone.

\paragraph{Limitations.}
Because expert annotation by practicing clinicians is costly, our human evaluation is limited to 30 patient cases and two raters, which constrains the generality of the conclusions. Scaling the study to more raters and a larger, more diverse set of cases is left for future work.

\subsection{Quantitative Analysis}
\label{app:llm_as_a_judge}
Expert evaluation does not scale. The study in Appendix~\ref{app:human_study} required two clinicians to read 120 rationales, which limits it to a few dozen cases. To obtain an approximate measure of reasoning quality at the scale of the test set, we adopt an LLM-as-a-judge evaluation, and then check its verdict against the clinicians on the cases for which both are available.

\paragraph{Protocol.}
We evaluate 1,000 randomly sampled P12 cases.
As a baseline, we apply IG to STraTS, a strong ISMTS model (Table~\ref{tab:main_results_full}), and render the attributions in two ways: the textualized form shown to the clinicians in Appendix~\ref{app:human_study}, and an additional GPT-5.1 interpretation step that translates pointwise attributions into temporal trends, which IG alone cannot express and which strengthens the baseline.
We first give LLM judges the rubric used by the clinicians in
Appendix~\ref{app:human_study}, but the resulting scores discriminate poorly between methods.
We therefore adopt the IDEA assessment tool~\citep{baker2015idea, schaye2021development}: each LLM judge evaluates reasoning traces along four dimensions, \textbf{I}nterpretive summary, \textbf{D}ifferential diagnosis, \textbf{E}xplanation of lead diagnosis, and \textbf{A}lternative diagnosis explained, with maximum subscores of 4/2/2/2, respectively, summing to a 0--10 total score.
We employ three judge models: GPT-5.1~\citep{openai2025gpt51systemcard}, Claude Sonnet 4.5~\citep{sonnet4p5}, and Gemini 3 Flash~\citep{gemini3flash}.
We query each judge three times per sample and report the mean of the nine resulting scores.

\input{tables/app_llm_judge_idea}

\paragraph{Results.}
\ours achieves a higher total IDEA score than the IG baseline even when the attributions are additionally interpreted by GPT-5.1 (7.708 vs.\ 6.680; Table~\ref{tab:idea_reasoning_score}) and is preferred on 83.5\% of cases. 
The largest gain appears on the interpretive summary criterion (+0.806), indicating that the model captures the patient's baseline risk and temporal trajectory more effectively.
The alternative diagnosis dimension also improves (+0.234), in line with our design choice to reason over alternative outcomes, while the scores on the remaining criteria are comparable.
Without the additional LLM interpretation step, the textualized IG attributions score $3.964$ and \ours is preferred on all cases, consistent with the margin observed in the expert study.

\paragraph{Agreement with the expert evaluation.}
The human evaluation protocol and LLM-as-a-judge evaluation protocol use different rubrics and different score ranges, so we compare the preferences their scores induce rather than the scores themselves.
For each case, we derive a preference for \ours, a tie, or IG from the difference in judge totals, and separately from the difference in each human criterion, and measure the agreement between them.
The two agree in every case on consistency and on helpfulness ($30/30$, 95\% CI $[88.6\%, 100\%]$), and in 28 of 30 on correctness.
Despite this agreement, we recognize that the automatic protocol cannot substitute for evaluation by clinical experts, and we therefore report it only as a scalable proxy for reasoning quality.

\subsection{Hallucination Analysis}
\label{app:hallucination}

We employ GPT-5.1 as a judge to detect severe hallucinations in \ours reasoning traces, such as references to features or values that were not present in the patient record (see Figure~\ref{fig:prompt_hallu_detect} for the full prompt).

\begin{figure*}[t]
\centering
\vspace{-2.0em}
\includegraphics[width=\textwidth,height=0.85\textheight,keepaspectratio]{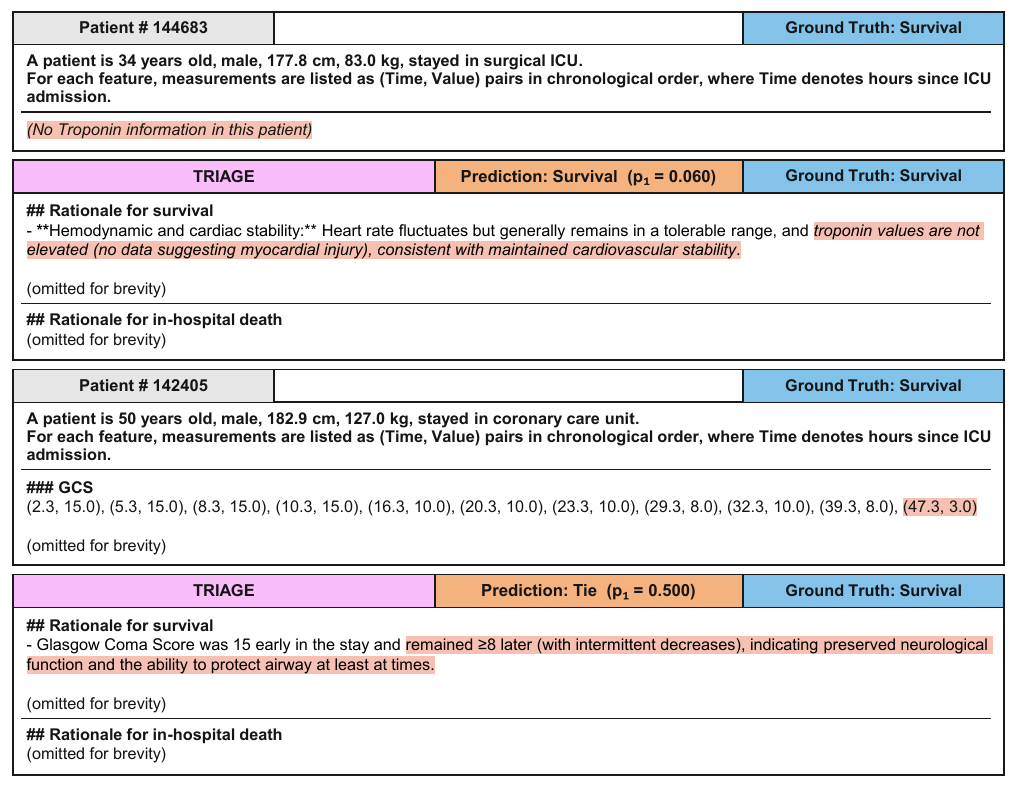}
\vspace{-1.5em}
\caption{
Case study of patients with hallucinated clinical rationales. For brevity, only relevant parts are excerpted. In the case of the first patient, the feature (Troponin T) not present in the patient's information is mentioned. In the case of the second patient, a value present in the patient's information is misidentified.
}
\vspace{-1.25em}
\label{fig:sample_hallu}
\end{figure*}

Out of 1,000 samples from the P12 dataset, only 1.3\% exhibit severe hallucinations.
Representative examples are in Figure~\ref{fig:sample_hallu}.
This indicates that during dialectical reasoning, the model rarely fabricates evidence for the opposing side when the signal from one side is already clear-cut. 
We attribute this favorable outcome to the safeguards built into our data construction process, in particular, the explicit instruction to leave a field blank whenever no supporting evidence is available.

\section{Further Experiments}
\label{app:additional_exp}

\subsection{Ablation Study on RL Reward Design}
\label{app:reward_design}

\begin{figure}[t]
  \centering
  \begin{subfigure}[t]{0.49\linewidth}
    \centering
    \includegraphics[width=\linewidth]{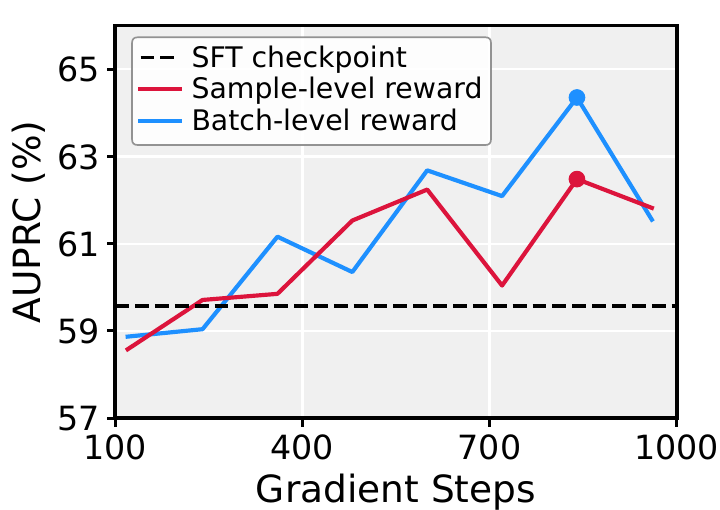}
  \end{subfigure}
  \begin{subfigure}[t]{0.49\linewidth}
    \centering
    \includegraphics[width=\linewidth]{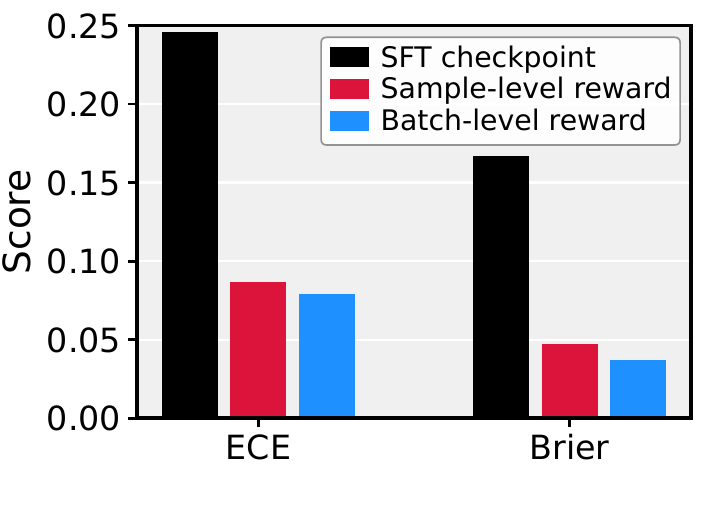}
  \end{subfigure}
  \caption{
    Discrimination (left) and calibration performance (right) under different reward designs on P12.
  }
  \label{fig:batch_vs_sample_reward}
\end{figure}

We compare our batch-level reward against a conventional sample-level reward, in which each rollout receives $R_{i,j}^{\mathrm{sample}} = \log P(y_i \mid x_i,\, r_{i,j})$, followed by the same group normalization.
This reward provides a direct correctness signal for each patient, akin to the standard outcome-based reward in GRPO, but does not explicitly encourage risk estimates that are comparable across patients. 

Relative to the SFT checkpoint, both rewards improve discrimination (AUPRC 59.58 $\to$ 62.48 and 64.35) and calibration, reducing ECE from 0.246 to below 0.09 and the Brier score from 0.167 to below 0.05 (Figure~\ref{fig:batch_vs_sample_reward}), which confirms that reinforcement learning itself is a major source of calibration gain and further improvement of discrimination.
Compared with the sample-level reward, the batch-level reward improves both discrimination with higher AUPRC (64.35 vs.\ 62.48) and calibration with lower ECE (0.079 vs.\ 0.087) and Brier score (0.037 vs.\ 0.047).
These results confirm that the inter-sample signal introduced by the batch-level reward provides a complementary learning objective that improves both the discrimination and calibration of predicted risk scores.

\subsection{Ablation Study on the Backbone}
\label{app:backbone}

To verify robustness to backbone choice, we apply \ours to Qwen3-1.7B and 8B~\citep{qwen3} in addition to our 4B default, and to Llama-3.2-3B~\citep{llama} from a different model family.
As shown in Table~\ref{tab:ablation_backbone}, scaling Qwen3 from 1.7B to 4B yields a clear AUPRC gain (53.8 to 56.4), while further scaling to 8B provides comparable performance (56.0), motivating our choice of 4B as the default backbone.
Across model families, Qwen3-4B achieves higher AUPRC than the similarly sized Llama-3.2-3B (56.4 vs.\ 54.5), while \ours retains a consistent advantage over the corresponding baseline on every backbone, suggesting that our reasoning supervision generalizes beyond a particular model scale or architecture.

\subsection{Inference Direction Analysis}
\label{app:direction_ablation}

Since \ours is trained on both rationale orderings, we examine whether averaging the risk scores from the two directions at inference offers any benefit over using a single ordering.
Table~\ref{tab:ablation_direction} reports the results on P12 with Qwen3-4B-Base under the SFT setting.
The three strategies perform comparably, indicating that a single-direction pass alone already yields a reliable risk estimate and can serve as a lower-cost alternative.
Nonetheless, bidirectional averaging consistently provides a small additional gain (\eg{AUPRC of $56.4$ vs.\ $56.1$/$56.2$}), and we adopt it as the default inference strategy.

\input{tables/app_further_analysis}

\section{Prompts}
\label{app:prompts}

We provide a comprehensive list of all the prompts utilized in our study, organized by their role in the pipeline.

\paragraph{Preliminary study.}
To motivate our method, we first probe zero-shot LLM behavior on ICU mortality prediction.
The default direct-answer prompt is shown in Figure~\ref{fig:prompt_default_preliminary},
its verbalized-probability variant in Figure~\ref{fig:prompt_verbalized_probability},
and the final evaluation prompt for zero-shot LLMs in Figure~\ref{fig:prompt_zeroshot_final}.
The LLM-as-a-judge prompt used to detect whether a chain-of-thought ends with an
explicit verdict-closure sentence is shown in Figure~\ref{fig:prompt_verdict_closure_judge}.

\paragraph{Outcome-conditioned rationale generation.}
To construct supervision for \ours, we elicit outcome-conditioned rationales from a strong LLM, instantiating one prompt per candidate outcome per patient.
The prompt template for the four in-hospital mortality datasets (P12, MIMIC-III, MIMIC-IV, and eICU) is shown in
Figure~\ref{fig:prompt_rationale_elicitation_p12_mimic3},
and the corresponding template for P19 (early sepsis prediction) is shown in
Figure~\ref{fig:prompt_rationale_elicitation_p19}.

\paragraph{Main SFT prompts for \ours.}
\ours is supervised to produce two outcome-conditioned rationale blocks followed
by a single final decision. To control for ordering effects, we use two symmetric
variants per dataset.
For the in-hospital mortality datasets (P12, MIMIC-III, MIMIC-IV, and eICU), the negative-class-first and positive-class-first variants
are shown in Figure~\ref{fig:prompt_method_sft_p12_mimic3_neg_first} and Figure~\ref{fig:prompt_method_sft_pos_first}, respectively.
For P19, the corresponding variants are shown in Figure~\ref{fig:prompt_method_sft_p19_neg_first} and Figure~\ref{fig:prompt_method_sft_p19_pos_first}.

\paragraph{Ablation: one-sided reasoning.}
For the reasoning-structure ablation (Table~\ref{tab:ablation_reasoning}), we use a
single-rationale SFT prompt that retains only the rationale aligned with the
ground-truth label. The P12 template is shown in Figure~\ref{fig:prompt_one_sided_sft_p12}.

\paragraph{Faithfulness analysis.}
For the input-removal experiment (Figure~\ref{fig:faithfulness}), GPT-5.1 locates the evidence behind the rationale sentences selected at each deletion ratio (25\%, 50\%, 75\%, and 100\%) and returns, for each sentence, the features it relies on together with the value range, time window, or whole series to erase; a deterministic script then removes the matching measurements from the input.
This prompt is shown in Figure~\ref{fig:prompt_evidence_erasure}.
For the regeneration test at a deletion ratio of 50\%, a GPT-5.1 judge compares every sentence of the original rationale with the rationale regenerated from the perturbed input and scores it from 1 (retained) to 5 (absent), with intermediate scores for claims restated with different values, findings kept without their values, and partial retention of the cited evidence.
The judge prompt is shown in Figure~\ref{fig:prompt_residual_presence_judge_5pt}.

\paragraph{Human expert evaluation criteria.}
The rubric presented to the clinical experts is adapted from the rationale evaluation protocol of~\citet{jiang2025kare}; the definitions of the criteria and the associated rating scales are shown in Figure~\ref{fig:human_eval_criteria}.
To reduce redundancy among evaluation dimensions, we omit human-likeness and specificity, which capture aspects of rationale quality that partially overlap with helpfulness, and focus on consistency, correctness, and helpfulness as complementary measures of grounding, medical validity, and clinical utility.

\paragraph{Reasoning analysis: XAI interpretation.}
To compare model-internal explanations with LLM-style rationales, we textualize
attribution-based evidence from STraTS and ask an LLM to interpret it in the same
two-sided rationale format. The interpretation prompt is shown in Figure~\ref{fig:prompt_STraTS_XAI_LLM_Interpret}.

\paragraph{Quantitative analysis: IDEA-based reasoning quality assessment.}
To assess the clinical quality of generated reasoning traces, we adapt the
IDEA rubric into an LLM-as-a-judge protocol that scores Interpretive summary (I), Differential diagnosis (D), Explanation of the lead diagnosis (E), and Alternative diagnosis explained (A).
The full evaluation prompt, including task setup, factuality rules, per-domain scoring anchors, overall calibration guidance, and the required JSON output
schema, is shown across Figures~\ref{fig:prompt_llm_as_a_judge_1_4}--\ref{fig:prompt_llm_as_a_judge_4_4}.

\paragraph{Hallucination analysis.}
To quantify hallucination in our reasoning traces, we adopt an LLM-as-a-judge approach that evaluates whether each trace references features or values absent from the patient record. We use GPT-5.1 as a judge, and the exact prompt is in Figure~\ref{fig:prompt_hallu_detect}.

\newpage

\begin{figure*}[t]
\centering
\includegraphics[width=0.49\linewidth]{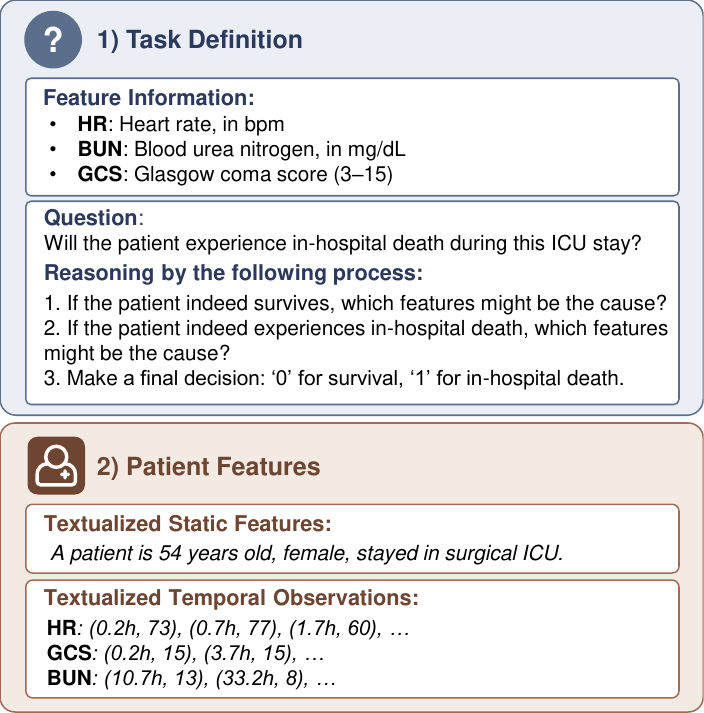}
\caption{
Example prompt format used by \ours. Only some of the features are shown. The input is organized into task definition and textualized patient features.
}
\vspace{-1.5em}
\label{fig:prompt_format}
\end{figure*}

\begin{figure*}[t]
\centering
\vspace{-2.0em}
\includegraphics[width=\textwidth,height=0.85\textheight,keepaspectratio]{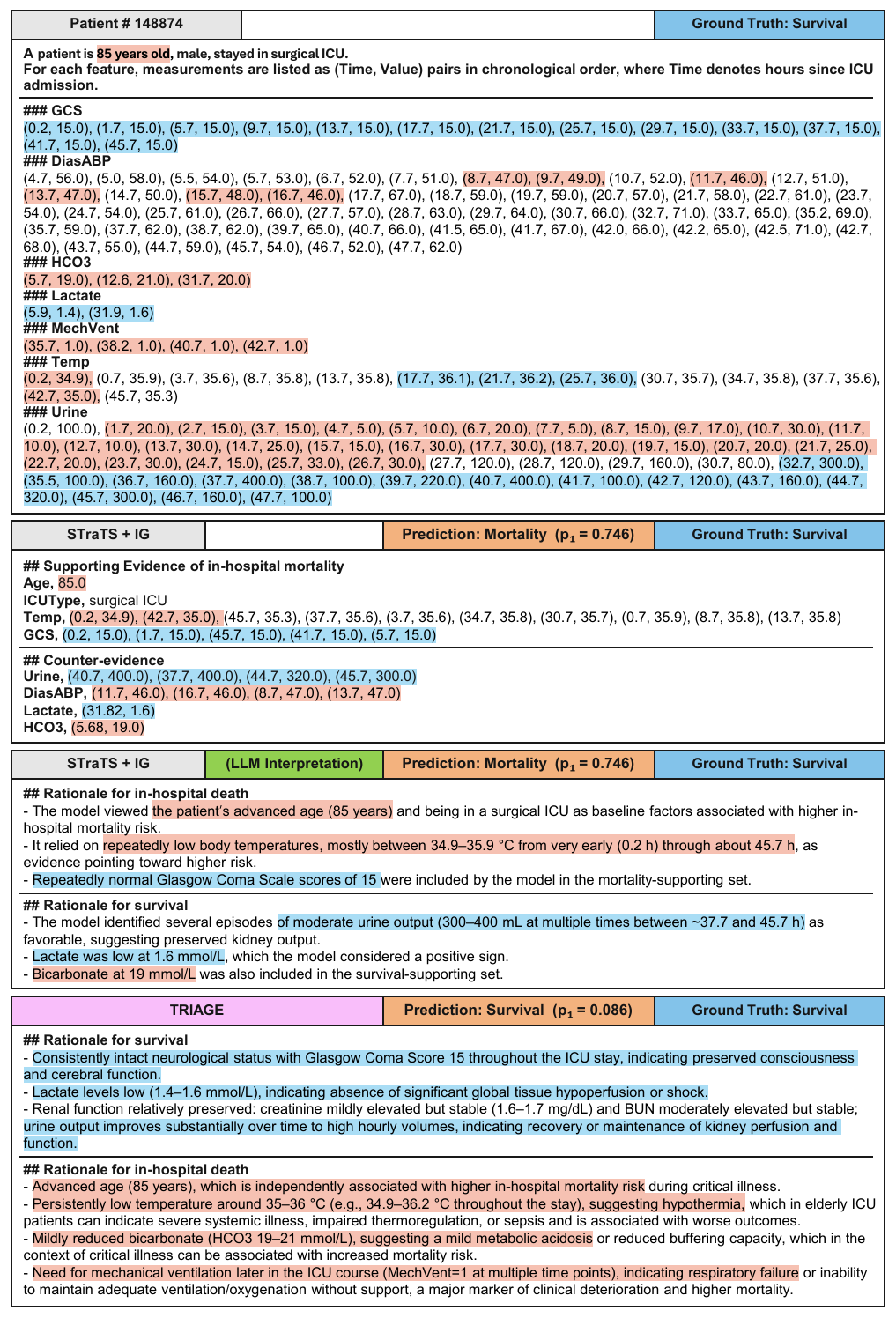}
\caption{
Case study of a surviving patient. For brevity, only a subset of patient characteristics and some of the clinical reasoning parts are excerpted. Text highlighted in {\setlength{\fboxsep}{1pt}\smash{\colorbox{cyan!25}{blue}}} and
{\setlength{\fboxsep}{1pt}\smash{\colorbox{red!20}{red}}} indicates medically favorable and unfavorable signs, respectively.
}
\vspace{-1.25em}
\label{fig:sample1}
\end{figure*}

\begin{figure*}[t]
\centering
\vspace{-2.0em}
\includegraphics[width=\textwidth,height=0.85\textheight,keepaspectratio]{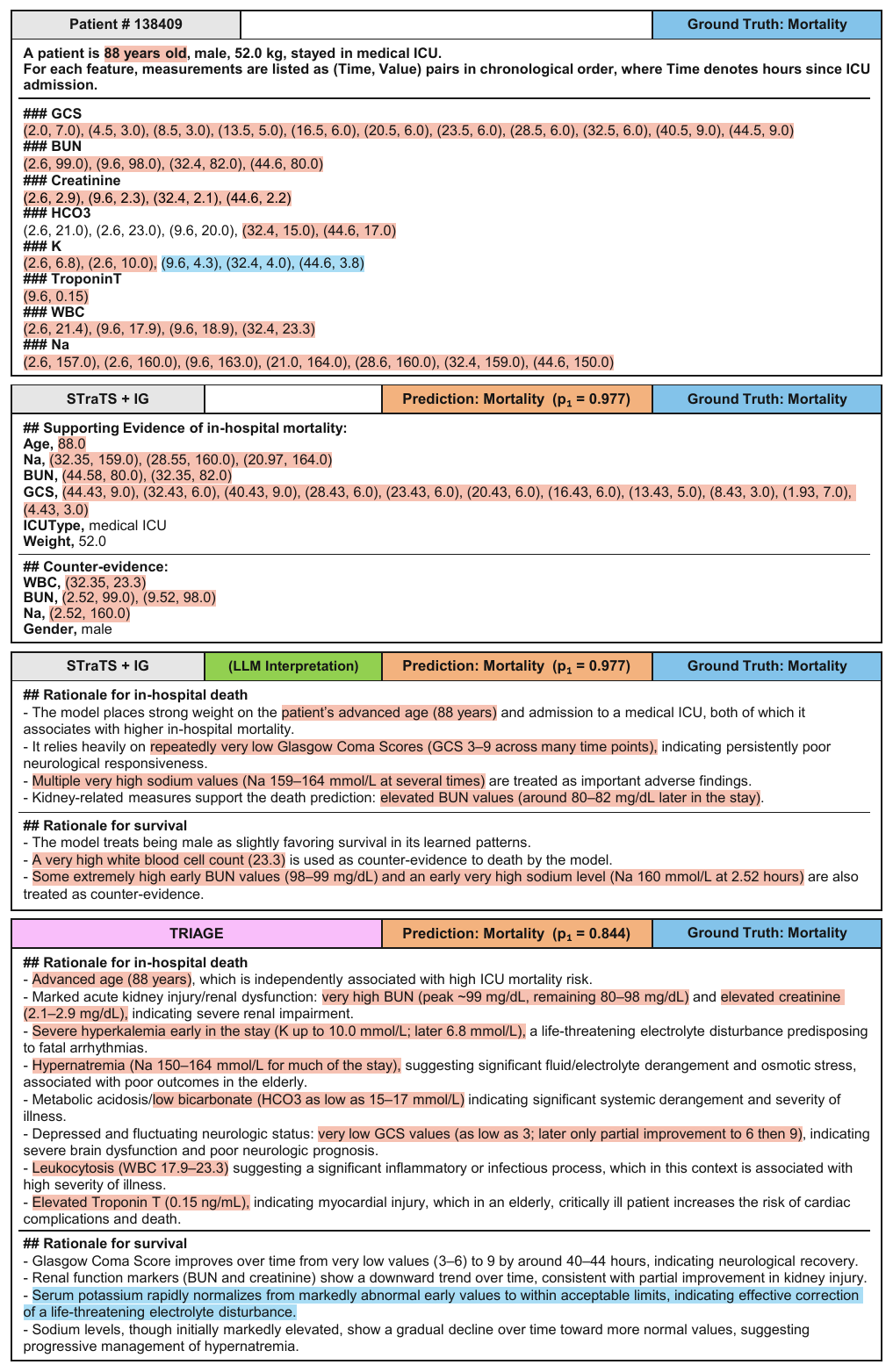}
\caption{
Case study of a non-survivor. For brevity, only a subset of patient characteristics and some of the clinical reasoning parts are excerpted. Text highlighted in {\setlength{\fboxsep}{1pt}\smash{\colorbox{cyan!25}{blue}}} and
{\setlength{\fboxsep}{1pt}\smash{\colorbox{red!20}{red}}} indicates medically favorable and unfavorable signs, respectively.
}
\vspace{-1.25em}
\label{fig:sample2}
\end{figure*}

\input{tables/prompts}

%% file: tables/app_cost_analysis.tex
\begin{table*}[t]
\centering
\small
\setlength{\tabcolsep}{1pt}
\caption{
Inference cost on clinical risk prediction benchmarks. Each entry is inference time per patient in milliseconds, measured on a single B200 GPU. GPT-5.1 is not run locally; its entries are the mean latency of synchronous API calls and are not directly comparable. The last column is the mean over datasets.
}
\vspace{-0.5em}
\label{tab:cost_analysis}
\begin{tabular*}{\textwidth}{@{\extracolsep{\fill}}lcccccc}
\toprule
\textbf{Method} & \textbf{P12} & \textbf{P19} & \textbf{MIMIC-III} & \textbf{MIMIC-IV} & \textbf{eICU} & \textbf{Mean} \\
\midrule

\rowcolor{gray!12}
\multicolumn{7}{c}{\textit{Specialized ISMTS Models} (non-explainable in natural language)} \\
\midrule

GRU-D    & 0.157 & 0.018 & 0.057 & 0.300  & 0.313 & 0.169 \\
mTAND    & 0.033 & 0.013 & 0.024 & 0.024  & 0.018 & 0.022 \\
SeFT     & 0.096 & 0.079 & 0.091 & 0.101  & 0.108 & 0.095 \\
Raindrop & 0.347 & 0.355 & 0.368 & 8.683  & 0.387 & 2.028 \\
STraTS   & 0.471 & 0.130 & 0.483 & 0.304  & 0.378 & 0.353 \\
ViTST    & 3.188 & 3.178 & 3.182 & 3.182  & 3.187 & 3.184 \\
KEDGN    & 0.454 & 0.058 & 0.113 & 10.430 & 0.423 & 2.296 \\
Hi-Patch & 2.239 & 0.740 & 0.590 & 8.533  & 2.033 & 2.827 \\
GSNF     & 0.239 & 0.071 & 0.126 & 1.377  & 0.168 & 0.396 \\
GARLIC   & 1.250 & 0.321 & 0.293 & 6.744  & 0.465 & 1.814 \\

\midrule
\rowcolor{cyan!12}
\multicolumn{7}{c}{\textit{Zero-shot LLMs}} \\
\midrule

GPT-5.1      & 12,609 & 10,878 & -- & -- & -- & 11,744 \\
gpt-oss-120b & 143.0 & 91.1 & 132.4 & 118.4 & 159.1 & 128.8 \\
Qwen3-4B     & 463.9 & 652.5 & 468.1 & 436.4 & 787.2 & 561.6 \\

\midrule
\rowcolor{cyan!12}
\multicolumn{7}{c}{\textit{LLM-based Prediction Methods}} \\
\midrule

Record2Vec & 249.5 & 196.4 & 264.9 & 221.5 & 242.1 & 234.9 \\
TimeCAP    & 400.0 & 332.1 & 374.7 & 348.3 & 345.3 & 360.1 \\

\midrule

\textbf{\ours} & 223.3 & 113.6 & 134.8 & 170.5 & 194.7 & 167.4 \\

\bottomrule
\end{tabular*}
\end{table*}

%% file: tables/exp_setup_details_sft_rl.tex
\begin{table}[t]
\centering

\begin{minipage}[t]{0.48\textwidth}
\vspace{0pt}
\centering
\footnotesize
\setlength{\tabcolsep}{2.5pt}

\captionof{table}{
SFT experimental setup details.
}
\label{tab:sft_experimental_setup_details}

\begin{tabular}{@{}ll@{}}
\toprule
Parameter & Value \\
\midrule

batch\_size
& \begin{tabular}[t]{@{}l@{}}
64 (P12)\\
128 (Others)
\end{tabular} \\

max\_length
& \begin{tabular}[t]{@{}l@{}}
12{,}288 (P12/MIMIC-III)\\
11{,}264 (P19) \\
16{,}384 (MIMIC-IV/eICU)
\end{tabular} \\

num\_train\_epochs
& 3 \\

optimizer
& AdamW \\

learning rate
& $2\times10^{-5}$ \\

lr\_scheduler\_type
& cosine \\

warmup\_ratio
& 0.05 \\

weight\_decay
& 0.01 \\

max\_grad\_norm
& 1.0 \\

completion\_only\_loss
& true \\

\bottomrule
\end{tabular}
\end{minipage}
\hfill
%
\begin{minipage}[t]{0.48\textwidth}
\vspace{0pt}
\centering
\footnotesize
\setlength{\tabcolsep}{2.5pt}

\captionof{table}{
RL experimental setup details.
}
\label{tab:rl_experimental_setup_details}

\begin{tabular}{@{}ll@{}}
\toprule
Parameter & Value \\
\midrule

max\_prompt\_length
& \begin{tabular}[t]{@{}l@{}}
10{,}240 (P12/MIMIC-III)\\
8{,}192 (P19) \\
16{,}384 (MIMIC-IV/eICU)
\end{tabular} \\

max\_response\_length
& 2{,}048 \\

train\_batch\_size
& 256 \\

ppo\_mini\_batch\_size
& 32 \\

total\_training\_steps
& 150 \\

optimizer
& AdamW \\

learning rate
& $1\times10^{-6}$ \\

lr\_warmup\_steps
& 10 \\

lr\_scheduler\_type
& constant \\

rollout.n
& 8 \\

rollout.temperature
& 1.0 \\

clip\_ratio\_low
& 0.20 \\

clip\_ratio\_high
& 0.28 \\

kl\_loss\_coef
& 0.001 \\

kl\_loss\_type
& mse ($k_2$) \\

\bottomrule
\end{tabular}
\end{minipage}

\end{table}

%% file: tables/main_result_full.tex
\begin{table*}[t]
\centering
\small
\setlength{\tabcolsep}{1pt}
\caption{
Discrimination performance on clinical risk prediction benchmarks. The best and second-best results among all methods, including the ISMTS baselines, are indicated in \textbf{bold} and \underline{underline}, respectively.
}
\vspace{-0.5em}
\label{tab:main_results_full}
\begin{tabular*}{\textwidth}{@{\extracolsep{\fill}}lcccccccccc}
\toprule
\multirow{2}{*}{\textbf{Method}}
& \multicolumn{2}{c}{\textbf{P12}}
& \multicolumn{2}{c}{\textbf{P19}}
& \multicolumn{2}{c}{\textbf{MIMIC-III}}
& \multicolumn{2}{c}{\textbf{MIMIC-IV}}
& \multicolumn{2}{c}{\textbf{eICU}} \\
\cmidrule(lr){2-3} \cmidrule(lr){4-5} \cmidrule(lr){6-7} \cmidrule(lr){8-9} \cmidrule(lr){10-11}
& AUROC & AUPRC & AUROC & AUPRC & AUROC & AUPRC & AUROC & AUPRC & AUROC & AUPRC \\
\midrule

\rowcolor{gray!12}
\multicolumn{11}{c}{\textit{Specialized ISMTS Models} (non-explainable in natural language)} \\
\midrule

GRU-D
& 87.1{\scriptsize$\pm$0.9} & 56.3{\scriptsize$\pm$2.3}
& \textbf{89.7{\scriptsize$\pm$0.9}} & \textbf{56.6{\scriptsize$\pm$2.9}}
& 84.7{\scriptsize$\pm$0.2} & 49.0{\scriptsize$\pm$0.6}
& 87.5{\scriptsize$\pm$0.5} & 59.2{\scriptsize$\pm$1.4}
& 84.4{\scriptsize$\pm$0.7} & 56.4{\scriptsize$\pm$2.2} \\

mTAND
& 86.8{\scriptsize$\pm$0.7} & 56.6{\scriptsize$\pm$2.4}
& 84.4{\scriptsize$\pm$1.6} & 39.4{\scriptsize$\pm$4.5}
& 84.6{\scriptsize$\pm$0.3} & 49.1{\scriptsize$\pm$0.9}
& 87.0{\scriptsize$\pm$0.2} & 58.0{\scriptsize$\pm$0.8}
& 84.7{\scriptsize$\pm$0.9} & 56.9{\scriptsize$\pm$1.4} \\

SeFT
& 86.6{\scriptsize$\pm$1.1} & 55.7{\scriptsize$\pm$4.4}
& 88.1{\scriptsize$\pm$1.2} & 51.1{\scriptsize$\pm$2.6}
& 84.7{\scriptsize$\pm$0.4} & 48.0{\scriptsize$\pm$1.2}
& 87.2{\scriptsize$\pm$0.4} & 56.8{\scriptsize$\pm$2.2}
& 85.0{\scriptsize$\pm$0.5} & 56.0{\scriptsize$\pm$0.6} \\

Raindrop
& 82.6{\scriptsize$\pm$1.5} & 45.4{\scriptsize$\pm$1.2}
& 85.5{\scriptsize$\pm$1.1} & 52.2{\scriptsize$\pm$3.0}
& 81.0{\scriptsize$\pm$0.6} & 37.2{\scriptsize$\pm$1.4}
& 83.3{\scriptsize$\pm$0.4} & 47.4{\scriptsize$\pm$1.1}
& 80.9{\scriptsize$\pm$0.8} & 50.8{\scriptsize$\pm$1.8} \\

STraTS
& \underline{87.3{\scriptsize$\pm$1.5}} & 57.4{\scriptsize$\pm$3.9}
& 88.7{\scriptsize$\pm$1.4} & 50.4{\scriptsize$\pm$2.3}
& 85.3{\scriptsize$\pm$0.5} & 48.3{\scriptsize$\pm$0.6}
& \textbf{88.6{\scriptsize$\pm$0.4}} & \underline{61.3{\scriptsize$\pm$1.1}}
& \textbf{85.3{\scriptsize$\pm$0.7}} & 57.4{\scriptsize$\pm$2.6} \\

ViTST
& 85.1{\scriptsize$\pm$0.7} & 50.4{\scriptsize$\pm$2.7}
& 86.8{\scriptsize$\pm$1.0} & 45.4{\scriptsize$\pm$3.1}
& 81.3{\scriptsize$\pm$0.7} & 38.9{\scriptsize$\pm$0.8}
& 86.4{\scriptsize$\pm$0.9} & 57.2{\scriptsize$\pm$2.6}
& 84.3{\scriptsize$\pm$1.0} & 54.4{\scriptsize$\pm$2.8} \\

KEDGN
& \textbf{87.8{\scriptsize$\pm$1.1}} & \underline{58.3{\scriptsize$\pm$3.2}}
& 88.2{\scriptsize$\pm$0.5} & 53.5{\scriptsize$\pm$3.0}
& 84.7{\scriptsize$\pm$0.4} & 48.5{\scriptsize$\pm$0.9}
& \underline{87.9{\scriptsize$\pm$0.5}} & 60.1{\scriptsize$\pm$1.2}
& \textbf{85.3{\scriptsize$\pm$0.6}} & 56.4{\scriptsize$\pm$1.4} \\

Hi-Patch
& \underline{87.3{\scriptsize$\pm$0.8}} & 57.0{\scriptsize$\pm$3.7}
& 88.7{\scriptsize$\pm$1.6} & 52.1{\scriptsize$\pm$4.3}
& 84.6{\scriptsize$\pm$0.6} & 46.2{\scriptsize$\pm$0.3}
& 87.7{\scriptsize$\pm$0.3} & 57.7{\scriptsize$\pm$0.6}
& 85.0{\scriptsize$\pm$0.8} & 56.5{\scriptsize$\pm$2.5} \\

GSNF
& 84.7{\scriptsize$\pm$1.3} & 50.7{\scriptsize$\pm$2.4}
& 86.2{\scriptsize$\pm$2.0} & 46.2{\scriptsize$\pm$3.2}
& 83.6{\scriptsize$\pm$0.4} & 40.5{\scriptsize$\pm$0.7}
& 86.0{\scriptsize$\pm$0.5} & 54.6{\scriptsize$\pm$1.6}
& 83.4{\scriptsize$\pm$0.7} & 54.4{\scriptsize$\pm$1.1} \\

GARLIC
& 86.6{\scriptsize$\pm$0.8} & 55.3{\scriptsize$\pm$2.2}
& 88.8{\scriptsize$\pm$0.6} & 52.4{\scriptsize$\pm$1.8}
& 85.5{\scriptsize$\pm$0.3} & 50.5{\scriptsize$\pm$1.5}
& 87.2{\scriptsize$\pm$0.5} & 57.9{\scriptsize$\pm$1.8}
& \underline{85.2{\scriptsize$\pm$1.0}} & \underline{58.2{\scriptsize$\pm$1.5}} \\

\midrule
\rowcolor{cyan!12}
\multicolumn{11}{c}{\textit{Zero-shot LLMs}} \\
\midrule

GPT-5.1
& 83.9{\scriptsize$\pm$1.0} & 49.3{\scriptsize$\pm$0.5}
& 72.2{\scriptsize$\pm$2.9} & 9.3{\scriptsize$\pm$0.9}
& -- & --
& -- & --
& -- & -- \\

gpt-oss-120b
& 81.2{\scriptsize$\pm$0.7} & 43.4{\scriptsize$\pm$3.0}
& 64.2{\scriptsize$\pm$2.1} & 6.7{\scriptsize$\pm$0.6}
& 76.6{\scriptsize$\pm$0.6} & 31.0{\scriptsize$\pm$1.0}
& 70.8{\scriptsize$\pm$0.5} & 27.9{\scriptsize$\pm$1.0}
& 79.5{\scriptsize$\pm$1.2} & 47.4{\scriptsize$\pm$1.0} \\

Qwen3-4B
& 69.7{\scriptsize$\pm$1.9} & 26.7{\scriptsize$\pm$0.8}
& 56.6{\scriptsize$\pm$1.7} & 5.0{\scriptsize$\pm$0.6}
& 69.3{\scriptsize$\pm$0.5} & 20.2{\scriptsize$\pm$0.2}
& 65.6{\scriptsize$\pm$0.7} & 20.9{\scriptsize$\pm$0.8}
& 71.2{\scriptsize$\pm$1.4} & 31.4{\scriptsize$\pm$0.9} \\

\midrule
\rowcolor{cyan!12}
\multicolumn{11}{c}{\textit{LLM-based Prediction Methods}} \\
\midrule

Record2Vec
& 76.9{\scriptsize$\pm$1.3} & 38.2{\scriptsize$\pm$2.9}
& 79.5{\scriptsize$\pm$1.6} & 33.4{\scriptsize$\pm$3.8}
& 73.6{\scriptsize$\pm$0.1} & 31.9{\scriptsize$\pm$0.1}
& 79.5{\scriptsize$\pm$1.2} & 43.6{\scriptsize$\pm$1.9}
& 75.9{\scriptsize$\pm$0.5} & 42.5{\scriptsize$\pm$1.1} \\

TimeCAP
& 85.2{\scriptsize$\pm$1.0} & 52.8{\scriptsize$\pm$3.9}
& 85.4{\scriptsize$\pm$1.3} & 33.7{\scriptsize$\pm$2.3}
& 84.2{\scriptsize$\pm$0.1} & 46.9{\scriptsize$\pm$0.2}
& 87.3{\scriptsize$\pm$0.3} & 57.5{\scriptsize$\pm$0.5}
& 83.8{\scriptsize$\pm$1.0} & 54.6{\scriptsize$\pm$3.2} \\

\midrule

\textbf{\ours}$_{\text{SFT}}$
& 86.9{\scriptsize$\pm$1.0} & 56.4{\scriptsize$\pm$1.9}
& 88.9{\scriptsize$\pm$1.0} & 52.2{\scriptsize$\pm$3.2}
& \underline{86.4{\scriptsize$\pm$0.3}} & \underline{51.4{\scriptsize$\pm$0.9}}
& 87.5{\scriptsize$\pm$0.2} & 59.7{\scriptsize$\pm$0.8}
& 84.2{\scriptsize$\pm$1.3} & 56.7{\scriptsize$\pm$0.9} \\

\textbf{\ours}
& \underline{87.3{\scriptsize$\pm$1.2}} & \textbf{59.0{\scriptsize$\pm$3.4}}
& \underline{89.3{\scriptsize$\pm$1.0}} & \underline{53.8{\scriptsize$\pm$1.4}}
& \textbf{86.7{\scriptsize$\pm$0.2}} & \textbf{54.1{\scriptsize$\pm$0.5}}
& 87.8{\scriptsize$\pm$0.5} & \textbf{61.6{\scriptsize$\pm$2.1}}
& \underline{85.2{\scriptsize$\pm$0.5}} & \textbf{58.7{\scriptsize$\pm$1.4}} \\

\bottomrule
\end{tabular*}
\vspace{-1.25em}
\end{table*}

%% file: tables/human_study_results.tex
\begin{table}[t]
\centering
\small
\caption{
Expert evaluation of generated reasoning.
Two clinicians rate all rationales on a 1--5 scale.
$^{\dagger}$/$^{\ddagger}$: differs from \ours at $p<0.05$ / $p<0.01$
(two-sided Wilcoxon signed-rank over cases).
}
\vspace{-0.5em}
\label{tab:human_eval}
\setlength{\tabcolsep}{6pt}
\begin{tabular}{lcccc}
\toprule
\textbf{Method} & \textbf{Consistency} & \textbf{Correctness} & \textbf{Helpfulness} & \textbf{Overall} \\
\midrule
STraTS\,+\,IG (textualized) & 3.28$^{\ddagger}$ & 3.43$^{\ddagger}$ & 2.48$^{\ddagger}$ & 3.07$^{\ddagger}$ \\
GPT-5.1 (zero-shot)         & \textbf{4.72} & \textbf{4.60}$^{\ddagger}$ & \textbf{4.73}$^{\dagger}$ & \textbf{4.68}$^{\ddagger}$ \\
\midrule
\ours (one-sided reasoning) & 4.43 & 4.32 & 3.93$^{\ddagger}$ & 4.23$^{\ddagger}$ \\
\ours                          & \underline{4.55} & \underline{4.35} & \underline{4.55} & \underline{4.48} \\
\bottomrule
\end{tabular}
\vspace{-1.0em}
\end{table}

%% file: tables/app_llm_judge_idea.tex
\begin{table}[t]
\centering
\small

\caption{
LLM-as-a-judge evaluation of clinical reasoning quality on 1{,}000 P12 samples.
Baseline denotes STraTS + IG (textualized).
\textit{LLM int.}: GPT-5.1 interpretation of IG attributions. 
Maximum scores for I/D/E/A are 4/2/2/2, respectively.
}
\vspace{-0.5em}
\label{tab:idea_reasoning_score}

\begin{tabular}{@{}lccc@{}}
\toprule
Criterion & Baseline & Baseline+LLM int. & \ours \\
\midrule
Interpretive summary (I)
& 1.384
& 2.619
& \textbf{3.425}\\

Differential diagnosis (D)
& 0.751
& \textbf{1.254}
& 1.213\\

Lead diagnosis exp. (E)
& 0.657
& 1.155
& \textbf{1.182}\\

Alternative diagnosis exp. (A)
& 1.171
& 1.652
& \textbf{1.887}\\

\midrule
Total
& 3.964
& 6.680
& \textbf{7.708}\\
\bottomrule
\end{tabular}
\end{table}

%% file: tables/app_further_analysis.tex
\begin{table}[t]
\centering

\begin{minipage}[t]{0.48\textwidth}
\vspace{0pt}
\centering
\scriptsize
\setlength{\tabcolsep}{4pt}

\captionof{table}{
Ablation on the backbone, on P12.
}
\label{tab:ablation_backbone}

\vspace{-0.5em}

\resizebox{0.9\linewidth}{!}{%
\begin{tabular}{lcc}
\toprule
Backbone & AUROC & AUPRC \\
\midrule
\textbf{Qwen3-4B} (default)
& 86.9{\scriptsize$\pm$1.0}
& \textbf{56.4{\scriptsize$\pm$1.9}} \\
\midrule
\rowcolor{cyan!12}
\multicolumn{3}{l}{\textbf{Scale} (Qwen3 family)} \\
\quad Qwen3-1.7B
& 86.4{\scriptsize$\pm$1.0}
& 53.8{\scriptsize$\pm$1.6} \\
\quad Qwen3-8B
& 86.8{\scriptsize$\pm$0.8}
& 56.0{\scriptsize$\pm$2.1} \\
\midrule
\rowcolor{cyan!12}
\multicolumn{3}{l}{\textbf{Architecture}} \\
\quad Llama-3.2-3B
& \textbf{87.0{\scriptsize$\pm$1.0}}
& 54.5{\scriptsize$\pm$1.3} \\
\bottomrule
\end{tabular}%
}
\end{minipage}
\hfill
%
\begin{minipage}[t]{0.48\textwidth}
\vspace{0pt}
\centering
\scriptsize
\setlength{\tabcolsep}{4pt}

\captionof{table}{
Ablation on the inference direction, on P12 with Qwen3-4B-Base under the SFT setting.
\textit{Positive-first} and \textit{Negative-first} denote inference using only the $(y^+, y^-)$ or $(y^-, y^+)$ ordering, respectively.
\textit{Bidirectional} averages the risk scores from both orderings.
}
\label{tab:ablation_direction}

\vspace{-0.5em}

\resizebox{\linewidth}{!}{%
\begin{tabular}{lcc}
\toprule
Inference strategy & AUROC & AUPRC \\
\midrule
Positive-first only
& 86.8{\scriptsize$\pm$1.0}
& 56.1{\scriptsize$\pm$1.0} \\
Negative-first only
& 86.8{\scriptsize$\pm$1.0}
& 56.2{\scriptsize$\pm$2.2} \\
\midrule
\textbf{Bidirectional (ours)}
& \textbf{86.9}{\scriptsize$\pm$1.0}
& \textbf{56.4}{\scriptsize$\pm$1.9} \\
\bottomrule
\end{tabular}%
}
\end{minipage}

\end{table}

%% file: tables/prompts.tex

\onecolumn
\begin{figure}[p]
\footnotesize
\begin{tcblisting}{
    text only,
    halign=left,
    title=\textbf{Default prompt used in the preliminary study},
    colbacktitle=gray!30!white,
    coltitle=black,
}
I will provide you with medical information from Intensive Care Unit (ICU) visit of a patient, each characterized by number of features.\\
The list of features are as follows:\\
\{FEATURE\_INFO\}\\
~\\
Read the patient's ICU features and answer immediately.\\
~\\
\#\# Task\\
Will the patient experience in-hospital death during this ICU stay?\\
~\\
Reply with only one number, immediately.\\
- 0 for survival\\
- 1 for in-hospital death\\
Do not provide any explanation or additional text.\\
~\\
\#\# Feature of the patient\\
\{PATIENT\_FEATURES\}
\end{tcblisting}
\vspace{-2mm}
\caption{Default prompt used in the preliminary study for in-hospital mortality prediction.}
\label{fig:prompt_default_preliminary}
\end{figure}

\begin{figure}[p]
\footnotesize
\begin{tcblisting}{
    text only,
    halign=left,
    title=\textbf{Prompt for verbalized probability extraction},
    colbacktitle=gray!30!white,
    coltitle=black,
}
I will provide you with medical information from Intensive Care Unit (ICU) visit of a patient, each characterized by number of features.\\
The list of features are as follows:\\
\{FEATURE\_INFO\}\\
~\\
Read the patient's ICU features and answer immediately.\\
~\\
\#\# Task\\
Estimate the probability that this patient will experience in-hospital death during this ICU stay.\\
~\\
Reply with only a single number between 0 and 1, immediately.\\
- 0.0 means certain survival\\
- 1.0 means certain death\\
Do not provide any explanation or additional text.\\
~\\
\#\# Feature of the patient\\
\{PATIENT\_FEATURES\}
\end{tcblisting}
\vspace{-2mm}
\caption{Prompt used for verbalized probability extraction. Only the \texttt{\#\# Task} section differs from Figure~\ref{fig:prompt_default_preliminary}.}
\label{fig:prompt_verbalized_probability}
\end{figure}

\begin{figure}[p]
\footnotesize
\begin{tcblisting}{
    text only,
    halign=left,
    title=\textbf{Prompt used for zero-shot LLM evaluation},
    colbacktitle=gray!30!white,
    coltitle=black,
}
I will provide you with medical information from Intensive Care Unit (ICU) visit of a patient, each characterized by number of features.\\
The list of features are as follows:\\
\{FEATURE\_INFO\}\\
~\\
\#\# Task\\
Estimate the probability that this patient will experience in-hospital death during this ICU stay.\\
~\\
Analyze the patient's clinical state, then answer.\\
~\\
Reply with only a single number between 0 and 1 at the end.\\
- 0.0 means certain survival\\
- 1.0 means certain death\\
~\\
\#\# Feature of the patient\\
\{PATIENT\_FEATURES\}
\end{tcblisting}
\vspace{-2mm}
\caption{Final prompt used for zero-shot LLM evaluation of in-hospital mortality.}
\label{fig:prompt_zeroshot_final}
\end{figure}

\begin{figure}[p]
\footnotesize
\begin{tcblisting}{
    text only,
    halign=left,
    title=\textbf{Prompt used for LLM-as-a-judge verdict-closure detection},
    colbacktitle=gray!30!white,
    coltitle=black,
}
You will read a chain-of-thought reasoning produced by another LLM on a binary in-hospital mortality task (0 = survival, 1 = in-hospital death).\\
~\\
Strict criterion:\\
The VERY LAST SENTENCE of the reasoning must, by itself, explicitly state the final binary answer --- either:\\
~~(a) the digit ``0'' or ``1'' in the context of the answer, OR\\
~~(b) the words ``survival''/``survive''/``live'' or ``death''/``die'' presented as the final decision.\\
~\\
If the last sentence does anything else --- discusses evidence, raises a caveat, gives a meta-instruction (e.g.\ ``Thus output single number''), asks a question, or trails off --- output 0.\\
~\\
Reply with exactly one digit:\\
- 1 = the LAST sentence is itself an explicit confirmatory restatement of the final answer\\
- 0 = it is not\\
~\\
Reasoning to evaluate:\\
\texttt{"""}\\
\{reasoning\}\\
\texttt{"""}
\end{tcblisting}
\vspace{-2mm}
\caption{Prompt used for LLM-as-a-judge verdict-closure detection.}
\label{fig:prompt_verdict_closure_judge}
\end{figure}


\onecolumn
\begin{figure}[p]
\footnotesize
\begin{tcblisting}{
    text only,
    halign=left,
    fontupper=\scriptsize,
    title=\textbf{Prompt for outcome-conditioned rationale elicitation (P12, MIMIC-III, MIMIC-IV, eICU)},
    colbacktitle=gray!30!white,
    coltitle=black,
}
I will provide you with medical information from Intensive Care Unit (ICU) visit of a patient, each characterized by number of features.\\
The list of features are as follows:\\
\{FEATURE\_INFO\}\\
~\\
Based on the given features of a patient, provide possible clinical rationale for why the patient could \{outcome\_phrase\} during this ICU stay.\\
Assume the outcome is \{outcome\_phrase\} and list only supporting features.\\
- Use only the provided features; do not invent data.\\
- Do not describe normal or healthy findings as abnormal.\\
- Do not describe abnormal findings as normal or healthy.\\
- Do not mention any label, class, or numeric outcome.\\
- Do not discuss the opposite outcome.\\
- If there is no supporting evidence, leave the rationale blank.\\
~\\
Your answer format must be:\\
\texttt{\textasciigrave\textasciigrave\textasciigrave}\\
\#\# Rationale for \{outcome\_phrase\}\\
{[blank or concise bullet points]}\\
\texttt{\textasciigrave\textasciigrave\textasciigrave}\\
~\\
\#\# Feature of the patient\\
\{PATIENT\_FEATURES\}
\end{tcblisting}
\vspace{-2mm}
\caption{Prompt used to elicit outcome-conditioned rationales from a strong LLM on the in-hospital mortality datasets (P12, MIMIC-III, MIMIC-IV, and eICU). The same prompt structure is used for all four datasets; only the feature-list block is replaced with the dataset-specific list. The placeholder \texttt{\{outcome\_phrase\}} is instantiated separately as \texttt{in-hospital death} or \texttt{survival}, yielding one outcome-specific prompt per outcome per patient.}
\label{fig:prompt_rationale_elicitation_p12_mimic3}
\end{figure}

\begin{figure}[p]
\footnotesize
\begin{tcblisting}{
    text only,
    halign=left,
    fontupper=\scriptsize,
    title=\textbf{Prompt for outcome-conditioned rationale elicitation (P19)},
    colbacktitle=gray!30!white,
    coltitle=black,
}
I will provide you with medical information from Intensive Care Unit (ICU) visit of a patient, each characterized by number of features.\\
The list of features are as follows:\\
\{FEATURE\_INFO\}\\
~\\
Based on the given features of a patient, provide possible clinical rationale for why the patient could experience \{assumption\}.\\
Assume the outcome is `\{assumption\}' and list only supporting features.\\
- Use only the provided features; do not invent data.\\
- Do not describe normal or healthy findings as abnormal.\\
- Do not describe abnormal findings as normal or healthy.\\
- Do not mention any label, class, or numeric outcome.\\
- Do not discuss the opposite outcome.\\
- If there is no supporting evidence, leave the rationale blank.\\
~\\
Your answer format must be:\\
\texttt{\textasciigrave\textasciigrave\textasciigrave}\\
\#\# \{header\}\\
{[blank or concise bullet points]}\\
\texttt{\textasciigrave\textasciigrave\textasciigrave}\\
~\\
\#\# Feature of the patient\\
\{PATIENT\_FEATURES\}
\end{tcblisting}
\vspace{-2mm}
\caption{Prompt used to elicit outcome-conditioned rationales from a strong LLM on P19 (early sepsis prediction). For each candidate outcome, the placeholders \texttt{\{assumption\}} and \texttt{\{header\}} are paired and substituted as either (\texttt{sepsis onset within the next 6 hours}, \texttt{Rationale for sepsis}) or (\texttt{no sepsis onset within the next 6 hours}, \texttt{Rationale for no sepsis}).}
\label{fig:prompt_rationale_elicitation_p19}
\end{figure}


\onecolumn
\begin{figure}[p]
\footnotesize
\begin{tcblisting}{
    text only,
    halign=left,
    fontupper=\scriptsize,
    title=\textbf{Main SFT prompt for \ours (P12, MIMIC-III, MIMIC-IV, eICU) \\--- negative-class-first variant},
    colbacktitle=gray!30!white,
    coltitle=black,
}
I will provide you with medical information from Intensive Care Unit (ICU) visit of a patient, each characterized by number of features.\\
The list of features are as follows:\\
\{FEATURE\_INFO\}\\
~\\
Based on the given feature of a patient, answer the question below.\\
~\\
\#\# Question\\
Will the patient experience in-hospital death during this ICU stay?\\
~\\
Reasoning by the following process:\\
1. If the patient indeed survives, which of the patient's given features might be the cause?\\
2. If the patient indeed experiences in-hospital death, which of the patient's given features might be the cause?\\
3. Make a final decision: `0' for survival, `1' for in-hospital death.\\
~\\
Your answer format must be as follows:\\
\texttt{\textasciigrave\textasciigrave\textasciigrave}\\
\#\# Rationale for survival\\
{[possible justification if patient survives]}\\
~\\
\#\# Rationale for in-hospital death\\
{[possible justification if patient experiences in-hospital death]}\\
~\\
\#\# Final Decision\\
{[0 (survival) or 1 (in-hospital death); respond by single number only]}\\
\texttt{\textasciigrave\textasciigrave\textasciigrave}\\
~\\
\#\# Feature of the patient\\
\{PATIENT\_FEATURES\}
\end{tcblisting}
\vspace{-2mm}
\caption{Main SFT prompt used for \ours on the in-hospital mortality datasets (P12, MIMIC-III, MIMIC-IV, and eICU; \emph{negative-class-first} variant). The model is supervised to produce two outcome-conditioned rationale blocks followed by a single \texttt{\#\# Final Decision}, with the rationale for the negative class (\texttt{survival}, label \texttt{0}) preceding the rationale for the positive class (\texttt{in-hospital death}, label \texttt{1}).}
\label{fig:prompt_method_sft_p12_mimic3_neg_first}
\end{figure}

\begin{figure}[p]
\footnotesize
\begin{tcblisting}{
    text only,
    halign=left,
    fontupper=\scriptsize,
    title=\textbf{Main SFT prompt for \ours (P12, MIMIC-III, MIMIC-IV, eICU) \\--- positive-class-first variant},
    colbacktitle=gray!30!white,
    coltitle=black,
}
I will provide you with medical information from Intensive Care Unit (ICU) visit of a patient, each characterized by number of features.\\
The list of features are as follows:\\
\{FEATURE\_INFO\}\\
~\\
Based on the given feature of a patient, answer the question below.\\
~\\
\#\# Question\\
Will the patient experience in-hospital death during this ICU stay?\\
~\\
Reasoning by the following process:\\
1. If the patient indeed experiences in-hospital death, which of the patient's given features might be the cause?\\
2. If the patient indeed survives, which of the patient's given features might be the cause?\\
3. Make a final decision: `0' for survival, `1' for in-hospital death.\\
~\\
Your answer format must be as follows:\\
\texttt{\textasciigrave\textasciigrave\textasciigrave}\\
\#\# Rationale for in-hospital death\\
{[possible justification if patient experiences in-hospital death]}\\
~\\
\#\# Rationale for survival\\
{[possible justification if patient survives]}\\
~\\
\#\# Final Decision\\
{[0 (survival) or 1 (in-hospital death); respond by single number only]}\\
\texttt{\textasciigrave\textasciigrave\textasciigrave}\\
~\\
\#\# Feature of the patient\\
\{PATIENT\_FEATURES\}
\end{tcblisting}
\vspace{-2mm}
\caption{Main SFT prompt used for \ours on the in-hospital mortality datasets (P12, MIMIC-III, MIMIC-IV, and eICU; \emph{positive-class-first} variant): the rationale for the positive class (\texttt{in-hospital death}, label \texttt{1}) precedes the rationale for the negative class (\texttt{survival}, label \texttt{0}). Compared with the negative-class-first variant (Figure~\ref{fig:prompt_method_sft_p12_mimic3_neg_first}), only the ordering of the reasoning steps and rationale blocks differs; all other text is identical.}
\label{fig:prompt_method_sft_pos_first}
\end{figure}

\begin{figure}[p]
\footnotesize
\begin{tcblisting}{
    text only,
    halign=left,
    fontupper=\scriptsize,
    title=\textbf{Main SFT prompt for \ours (P19) --- negative-class-first variant},
    colbacktitle=gray!30!white,
    coltitle=black,
}
I will provide you with medical information from Intensive Care Unit (ICU) visit of a patient, each characterized by number of features.\\
The list of features are as follows:\\
\{FEATURE\_INFO\}\\
~\\
Based on the given feature of a patient, answer the question below.\\
~\\
\#\# Question\\
Will the patient experience sepsis onset within the next 6 hours?\\
~\\
Reasoning by the following process:\\
1. If the patient indeed does not experience sepsis onset within the next 6 hours, which of the patient's given features might be the cause?\\
2. If the patient indeed experiences sepsis onset within the next 6 hours, which of the patient's given features might be the cause?\\
3. Make a final decision: `0' for no sepsis onset within the next 6 hours, `1' for sepsis onset within the next 6 hours.\\
~\\
Your answer format must be as follows:\\
\texttt{\textasciigrave\textasciigrave\textasciigrave}\\
\#\# Rationale for no sepsis\\
{[possible justification if patient does not experience sepsis onset within the next 6 hours]}\\
~\\
\#\# Rationale for sepsis\\
{[possible justification if patient experiences sepsis onset within the next 6 hours]}\\
~\\
\#\# Final Decision\\
{[0 (for no sepsis onset within the next 6 hours) or 1 (for sepsis onset within the next 6 hours); respond by single number only]}\\
\texttt{\textasciigrave\textasciigrave\textasciigrave}\\
~\\
\#\# Feature of the patient\\
\{PATIENT\_FEATURES\}
\end{tcblisting}
\vspace{-2mm}
\caption{Main SFT prompt used for \ours on P19 (\emph{negative-class-first} variant). The model is supervised to produce two outcome-conditioned rationale blocks followed by a single \texttt{\#\# Final Decision}, with the rationale for the negative class (\texttt{no sepsis}, label \texttt{0}) preceding the rationale for the positive class (\texttt{sepsis}, label \texttt{1}).}
\label{fig:prompt_method_sft_p19_neg_first}
\end{figure}

\begin{figure}[p]
\footnotesize
\begin{tcblisting}{
    text only,
    halign=left,
    fontupper=\scriptsize,
    title=\textbf{Main SFT prompt for \ours (P19) --- positive-class-first variant},
    colbacktitle=gray!30!white,
    coltitle=black,
}
I will provide you with medical information from Intensive Care Unit (ICU) visit of a patient, each characterized by number of features.\\
The list of features are as follows:\\
\{FEATURE\_INFO\}\\
~\\
Based on the given feature of a patient, answer the question below.\\
~\\
\#\# Question\\
Will the patient experience sepsis onset within the next 6 hours?\\
~\\
Reasoning by the following process:\\
1. If the patient indeed experiences sepsis onset within the next 6 hours, which of the patient's given features might be the cause?\\
2. If the patient indeed does not experience sepsis onset within the next 6 hours, which of the patient's given features might be the cause?\\
3. Make a final decision: `0' for no sepsis onset within the next 6 hours, `1' for sepsis onset within the next 6 hours.\\
~\\
Your answer format must be as follows:\\
\texttt{\textasciigrave\textasciigrave\textasciigrave}\\
\#\# Rationale for sepsis\\
{[possible justification if patient experiences sepsis onset within the next 6 hours]}\\
~\\
\#\# Rationale for no sepsis\\
{[possible justification if patient does not experience sepsis onset within the next 6 hours]}\\
~\\
\#\# Final Decision\\
{[0 (for no sepsis onset within the next 6 hours) or 1 (for sepsis onset within the next 6 hours); respond by single number only]}\\
\texttt{\textasciigrave\textasciigrave\textasciigrave}\\
~\\
\#\# Feature of the patient\\
\{PATIENT\_FEATURES\}
\end{tcblisting}
\vspace{-2mm}
\caption{Main SFT prompt used for \ours on P19 (\emph{positive-class-first} variant): the rationale for the positive class (\texttt{sepsis}, label \texttt{1}) precedes the rationale for the negative class (\texttt{no sepsis}, label \texttt{0}). Compared with the negative-class-first variant (Figure~\ref{fig:prompt_method_sft_p19_neg_first}), only the ordering of the reasoning steps and rationale blocks differs; all other text is identical.}
\label{fig:prompt_method_sft_p19_pos_first}
\end{figure}


\begin{figure}[p]
\footnotesize
\begin{tcblisting}{
    text only,
    halign=left,
    fontupper=\scriptsize,
    title=\textbf{Prompt for one-sided rationale SFT (P12)},
    colbacktitle=gray!30!white,
    coltitle=black,
}
I will provide you with medical information from Intensive Care Unit (ICU) visit of a patient, each characterized by number of features.\\
The list of features are as follows:\\
\{FEATURE\_INFO\}\\
~\\
Based on the given feature of a patient, answer the question below.\\
~\\
\#\# Question\\
Will the patient experience in-hospital death during this ICU stay?\\
~\\
Reasoning by the following process:\\
1. Describe the clinical evidence observed in the patient's features.\\
2. Make a final decision: `0' for survival, `1' for in-hospital death.\\
~\\
Your answer format must be as follows:\\
\texttt{\textasciigrave\textasciigrave\textasciigrave}\\
\#\# Rationale\\
{[clinical evidence observed in the patient's features]}\\
~\\
\#\# Final Decision\\
{[0 (survival) or 1 (in-hospital death); respond by single number only]}\\
\texttt{\textasciigrave\textasciigrave\textasciigrave}\\
~\\
\#\# Feature of the patient\\
\{PATIENT\_FEATURES\}
\end{tcblisting}
\vspace{-2mm}
\caption{Prompt template used for the \emph{one-sided rationale SFT} baseline in our reasoning-structure ablation (Table~\ref{tab:ablation_reasoning}), conducted on P12. The supervision retains only the rationale steps that support the ground-truth label from our collected outcome-conditioned traces, so a single \texttt{\#\# Rationale} block is followed by the corresponding \texttt{\#\# Final Decision}.}
\label{fig:prompt_one_sided_sft_p12}
\end{figure}


\onecolumn
\begin{figure}[p]
\footnotesize
\begin{tcblisting}{
    text only,
    halign=left,
    fontupper=\scriptsize,
    title=\textbf{Faithfulness: Prompt for locating the input evidence of selected rationale sentences},
    colbacktitle=gray!30!white,
    coltitle=black,
}
You are a clinical data annotator for a faithfulness experiment on ICU mortality prediction.\\
~\\
You will be given:\\
1. Observed patient information from the first 48 hours of ICU stay.\\
2. Sentences selected from a rationale that a prediction model wrote, based on this patient information, to argue for the outcome \texttt{"\{OUTCOME\}"}.\\
~\\
Your task is to list every piece of data in the patient information that the selected sentences rely on.\\
The listed data will be erased from the patient information; after the erasure, none of the selected sentences should be derivable from what remains.\\
Do not judge whether the sentences or the prediction are medically correct.\\
~\\
\# What to list\\
~\\
For each selected sentence, find every feature it draws on, including features mentioned only as context (e.g., \texttt{"on mechanical ventilation"}), and list the data it relies on:\\
- Whole-series statements: if the sentence characterizes a feature as a whole (its overall level or range, a trend, a first-to-last change, persistence, or wording such as \texttt{"throughout"}, \texttt{"mostly"}, \texttt{"consistently"}, \texttt{"repeatedly"}), the whole feature is to be erased: use \texttt{"all"}.\\
- Specific statements: if the sentence refers only to particular values, time points, or episodes (e.g., \texttt{"initial K 6.4"}, \texttt{"MAP 56-60 mmHg around 37.5-37.7 h"}), give the condition that covers them and any other measurement from which the same statement could still be made: a range of values, a time window, or both. A script erases every pair of the feature that satisfies the condition, so do not copy pairs.\\
- Invasive and non-invasive blood pressure are different features (SysABP, DiasABP, MAP are invasive; NISysABP, NIDiasABP, NIMAP are non-invasive). If the sentence names one of them, by its feature name or by the word \texttt{"invasive"} or \texttt{"non-invasive"}, list only that feature. If the sentence does not say which one it means (e.g., \texttt{"blood pressure"}, \texttt{"diastolic pressure"}), apply the rules above to every observed variant. If the feature the sentence names has no measurements for this patient, list the observed variant that the statement evidently describes.\\
- Clinical concepts and derived quantities: if the sentence names a concept or a computed quantity rather than a feature (e.g., \texttt{"oliguria"}, \texttt{"lactic acidosis"}, \texttt{"PaO2/FiO2 ratio"}), list the data of the features that directly express or compute it.\\
- Static descriptors: list age, gender, height, weight, or ICU type (given in the sentence that begins with \texttt{"A patient is"}) only if the selected sentence explicitly refers to that descriptor (e.g., \texttt{"advanced age"}, \texttt{"cardiac surgery unit"}). Generic phrases such as \texttt{"critical illness"} or \texttt{"ICU stay"} do not refer to the ICU type.\\
- Cited values may be approximate or rounded; choose bounds that include them (e.g., \texttt{"in the 50s-60s"} is the value range 50 to 69.9). When unsure whether a measurement contributed to a selected sentence, include it.\\
- Do not list data that is only clinically related to a sentence but not needed to write it.\\
- If a sentence relies on nothing that is present in the patient information (e.g., it cites unrecorded data or the absence of a measurement), list nothing for it.\\
~\\
\# Output Format\\
~\\
Return one JSON object and nothing else:\\
\texttt{\{"deletions": [\{"sentence": \textless sentence id\textgreater, "feature": "\textless feature header, or Age / Gender / Height / Weight / ICUType\textgreater", "removed": "all" or \{"value": [low, high], "time": [start, end]\}\}]\}}\\
~\\
- \texttt{"removed"}: use \texttt{"all"} when every pair of the feature is to be erased, and always for a static descriptor. Otherwise give a condition: \texttt{"value"} is a range of measured values and \texttt{"time"} a window in hours since ICU admission. Give either one or both (then both must hold). Bounds are inclusive, null leaves an end open, and a single value or time point is a range with equal bounds, e.g., \texttt{\{"value": [120, null]\}}, \texttt{\{"time": [0, 6]\}}, \texttt{\{"value": [6.4, 6.4], "time": [1.5, 1.5]\}}.\\
- One object per (sentence, feature) pair. If a sentence needs several separate conditions on one feature (e.g., two episodes), use one object per condition.\\
- Every condition must select at least one measurement that is present in the patient information.\\
- Every selected sentence must appear at least once. If nothing is to be erased for a sentence, use \texttt{\{"sentence": \textless sentence id\textgreater, "feature": null, "removed": "none"\}}.\\
- Write compact JSON on a single line, without indentation or line breaks.\\
~\\
\# Input\\
~\\
\texttt{[Patient Information]}\\
\texttt{"""\{PATIENT\_FEATURES\}"""}\\
~\\
\texttt{[Selected Sentences]}\\
\texttt{"""\{SELECTED\_SENTENCES\}"""}
\end{tcblisting}
\vspace{-2mm}
\caption{Prompt used to locate, in the model input, the evidence behind selected rationale sentences. At deletion ratio $r \in \{0.25, 0.5, 0.75, 1.0\}$, $\lfloor rn \rfloor$ of the $n$ sentences in the rationale block of the predicted outcome are pre-selected at random (seed $42+i$ for the $i$-th patient) and passed as \texttt{\{SELECTED\_SENTENCES\}}; \texttt{\{OUTCOME\}} is \texttt{survival} or \texttt{in-hospital death}.}
\label{fig:prompt_evidence_erasure}
\end{figure}

\begin{figure}[p]
\footnotesize
\begin{tcblisting}{
    text only,
    halign=left,
    fontupper=\scriptsize,
    title=\textbf{Faithfulness: Prompt for judging the residual presence of rationale statements},
    colbacktitle=gray!30!white,
    coltitle=black,
}
You are an evaluator of clinical rationales written by an ICU mortality prediction model.\\
~\\
You will be given:\\
1. A numbered list of statements taken from a rationale that the model wrote for the outcome \texttt{"\{OUTCOME\}"} of one patient.\\
2. A second rationale for the same outcome and the same patient, which the model wrote later from a modified version of the patient's data.\\
~\\
Your task is to score, for every numbered statement, how completely it has disappeared from the second rationale.\\
Judge only what is written. Do not judge medical correctness, and do not guess which statements should or should not have changed.\\
~\\
\# How to compare\\
~\\
For each statement, identify (a) its clinical claim (e.g., \texttt{"acute kidney injury"}, \texttt{"stable blood pressure"}) and (b) its cited evidence: every feature it names, together with the values, ranges, or trends given for that feature.\\
Then find the passage of the second rationale that is most similar to the statement and assign exactly one score:\\
~\\
- 5 (Absent): no passage makes this claim or cites this evidence.\\
- 4 (Revised): a passage makes the same or a closely related claim, but none of the cited evidence is repeated: the features are cited with different numbers, or the claim rests on other features.\\
- 3 (Finding retained without values): a passage states the same finding for at least one of the cited features (e.g., \texttt{"creatinine remains normal"}), but none of the cited values, ranges, or trend end points is repeated.\\
- 2 (Partly retained): a passage repeats part of the cited evidence unchanged: at least one feature with the same value, range, or trend, while the rest is missing or different.\\
- 1 (Retained): a passage makes the same claim and repeats all of the cited evidence: the same features with the same values, ranges, or trends.\\
~\\
\# Rules\\
~\\
- Abbreviations and synonyms denote the same feature (e.g., \texttt{"bicarbonate"} and \texttt{"HCO3"}). Invasive and non-invasive blood pressure are different features.\\
- Two numbers are the same if they agree after rounding to the precision used in the statement. A range is the same only if both of its ends agree. A trend is the same if its direction and its end points agree. Words such as \texttt{"about"} or \texttt{"\textasciitilde"} do not matter.\\
- Wording, order, and interpretation do not matter; only the claim and the cited evidence do.\\
- If several passages match a statement, use the one that gives the lowest score.\\
- For scores 1--4, copy the matching passage (one sentence or bullet) verbatim from the second rationale. For score 5, use an empty string.\\
~\\
\# Output Format\\
~\\
Return one JSON object and nothing else:\\
\texttt{\{"judgments": [\{"statement": \textless statement id\textgreater, "score": 1 or 2 or 3 or 4 or 5, "passage": "\textless verbatim passage, or empty\textgreater"\}]\}}\\
~\\
Every statement must appear exactly once. Write compact JSON on a single line, without indentation or line breaks.\\
~\\
\# Input\\
~\\
\texttt{[Statements]}\\
\texttt{"""\{STATEMENTS\}"""}\\
~\\
\texttt{[Second Rationale]}\\
\texttt{"""\{SECOND\_RATIONALE\}"""}
\end{tcblisting}
\vspace{-2mm}
\caption{Prompt used to score, on a five-point scale (5 = absent, 1 = retained in full), how completely rationale statements disappear after input erasure. \texttt{\{STATEMENTS\}} lists \emph{all} sentences of the original rationale of the predicted outcome, without marking which ones were targeted by the erasure, and \texttt{\{SECOND\_RATIONALE\}} is the rationale for the same outcome regenerated from the input perturbed at a deletion ratio of 50\% (Figure~\ref{fig:prompt_evidence_erasure}); \texttt{\{OUTCOME\}} is \texttt{survival} or \texttt{in-hospital death}. The judge therefore does not know the label, the predicted probability, the targeted sentences, or the perturbation condition; targeted and untargeted sentences are separated only when the scores are aggregated.}
\label{fig:prompt_residual_presence_judge_5pt}
\end{figure}


\begin{figure}[p]
\footnotesize
\begin{tcblisting}{
    text only,
    halign=left,
    fontupper=\scriptsize,
    title=\textbf{Criteria presented to the clinical experts in the human evaluation},
    colbacktitle=gray!30!white,
    coltitle=black,
}
Rate the rationale generated by each method independently along the three criteria below, and record an integer score from 1 to 5 for every criterion. Higher scores indicate higher quality. Ties across methods are permitted: the scores are absolute judgments rather than a ranking. Leave a score blank and annotate the case if it cannot be assessed.\\
~\\
\#\# C1. Consistency (1--5 per method)\\
\textbf{Definition.} Consistency measures how well the generated rationale aligns with the model's prediction. It evaluates whether the reasoning contains contradictions.\\
- 5 (Excellent): The rationale is entirely consistent with the prediction. No contradictions are present.\\
- 4 (Good): Minor inconsistencies are present but do not significantly detract from the overall coherence.\\
- 3 (Fair): Some inconsistencies exist that partially undermine the rationale's reliability.\\
- 2 (Poor): Significant inconsistencies exist, making the rationale difficult to trust.\\
- 1 (Very Poor): The rationale is fundamentally contradictory to the prediction.\\
~\\
\#\# C2. Correctness (1--5 per method)\\
\textbf{Definition.} Correctness assesses the medical validity of the knowledge and reasoning presented in the rationale.\\
- 5 (Excellent): The rationale is entirely medically accurate and reflects evidence-based clinical knowledge.\\
- 4 (Good): Minor inaccuracies are present but do not affect the overall clinical validity.\\
- 3 (Fair): The rationale contains some medically incorrect statements that could mislead clinicians.\\
- 2 (Poor): Multiple inaccuracies significantly reduce the credibility of the rationale.\\
- 1 (Very Poor): The rationale is largely or entirely medically incorrect, making it unusable.\\
~\\
\#\# C3. Helpfulness (1--5 per method)\\
\textbf{Definition.} Helpfulness measures the extent to which the rationale aids the prediction toward the correct diagnosis.\\
- 5 (Excellent): The rationale strongly supports the correct prediction and adds valuable clinical insights.\\
- 4 (Good): The rationale is helpful overall but lacks some critical insights.\\
- 3 (Fair): The rationale provides some useful guidance but is incomplete or not compelling.\\
- 2 (Poor): The rationale adds minimal value to the prediction and lacks actionable insights.\\
- 1 (Very Poor): The rationale is unhelpful and does not contribute meaningfully to the diagnosis.\\
~\\
\#\# Optional patient input\\
The observed patient record may be consulted when needed. Time is reported as hours since ICU admission, and measurements are listed as (Time, Value) in chronological order. If the record is consulted for a case, the same input is applied while rating all four methods. Ground-truth outcomes are not provided to the raters.
\end{tcblisting}
\vspace{-2mm}
\caption{Definitions and rating scales presented to the clinical experts in the human evaluation (Appendix~\ref{app:human_study}), adapted from~\citet{jiang2025kare}. Each rationale is scored independently on consistency, correctness, and helpfulness using a five-point integer scale, and the raters are not given the ground-truth outcome.}
\label{fig:human_eval_criteria}
\end{figure}



\onecolumn
\begin{figure}[p]
\footnotesize

\begin{tcblisting}{
    text only,
    halign=left,
    fontupper=\scriptsize,
    title=\textbf{Prompt for the interpretation of the IG result of STraTS.},
    colbacktitle=gray!30!white,
    coltitle=black,
}
You are an interpreter for explanations of a trained ICU mortality prediction model.\\

You will be given a textualized summary of the model's prediction evidence.
This summary was generated from important patient-specific observations selected from the model.
Your task is to interpret what the provided evidence means about the model's prediction.\\
~\\

\# Input\\
~\\

\#\# List of features\\
\{FEATURE\_INFO\}\\

\#\# Evidence Summary\\
\{XAI\_RESULT\_TEXTUALIZE\}\\
~\\

\# Your Task\\
~\\
Explain what the provided evidence summary means about the model's prediction.\\

Interpretation means:\\
- identify what kinds of observations the model relied on,\\
- distinguish evidence for the predicted class from evidence for the opposite class,\\
- group related observations when appropriate,\\
- explain the model's prediction in plain language.\\

Interpretation does NOT mean:\\
- making a new clinical judgment,\\
- deciding whether the model is medically correct,\\
- inferring missing diagnoses, treatments, or outcomes,\\
- explaining the XAI method,\\
- discussing attribution scores or importance scores.\\
~\\

\# Rules\\
~\\

- Use only the information in the textualized evidence summary.\\
- Do not introduce new clinical interpretations that are not directly supported by the summary.\\
- Do not overstate the evidence.\\
- Do not evaluate the adequacy of the evidence.\\
- Do not add unnecessary concluding statements.\\
- Keep the interpretation concise.\\
~\\
\# Output Format\\
~\\
Only write the interpretation of the evidence predicted by the model. Write a concise interpretation ($\leq$ 10 sentences for each section) along with specific figures in the following format:\\
\#\# Rationale for survival\\
- Interpretation of evidences\\
\#\# Rationale for in-hospital death\\
- Interpretation of evidences
\end{tcblisting}

\vspace{-2mm}
\caption{Prompt for the interpretation of the IG result of STraTS.}
\label{fig:prompt_STraTS_XAI_LLM_Interpret}
\end{figure}


\onecolumn
\begin{figure}[p]
\footnotesize

\begin{tcblisting}{
    text only,
    halign=left,
    fontupper=\scriptsize,
    title=\textbf{Evaluation: Prompt for IDEA assessment tool (1/4)},
    colbacktitle=gray!30!white,
    coltitle=black,
}
You are an expert in clinical reasoning assessment and ICU prognostic reasoning. Your task is to evaluate one clinical reasoning trace and score how medically reasonable it is for predicting in-hospital mortality during the ICU stay.\\
~\\

\# Input Data \\
~\\

[Clinical Question]\\
"""Will the patient experience in-hospital death during this ICU stay?"""\\
~\\

[Patient Features]\\
"""[[PATIENT\_FEATURES]]"""\\
~\\

[Reasoning]\\
"""[[MODEL\_REASONING]]"""\\
~\\

\# Task Instructions\\
~\\
Evaluate whether the reasoning trace is clinically reasonable, faithful to the provided patient features, temporally aware, balanced, and appropriately calibrated for ICU in-hospital mortality prediction.\\

Your job is to judge the quality of the medical reasoning based only on the provided patient features.\\
Your job is NOT to determine the true outcome.\\
Your job is NOT to infer or reward matching any hidden label.\\
Your job is NOT to reward longer, more fluent, more confident, or more detailed writing unless the added detail is accurate, patient-specific, clinically relevant, and proportional to the evidence.\\

Score the reasoning using the adapted Revised-IDEA rubric below.\\
~\\

\# Mandatory Factuality and Data-Use Rules\\
~\\

Use only the provided patient features.

You may use general medical knowledge to interpret the provided features, but you must not add clinical facts that are not present.

Do NOT assume unstated diagnoses, medications, procedures, organ support, code status, goals of care, complications, outcomes, or information after the prediction window.\\

Do NOT treat missing data as normal or abnormal without justification.\\

Do NOT infer the true label.\\
Do NOT reward the reasoning for matching any hidden label.\\
Do NOT reward a model simply because it gives a more extreme mortality probability.\\
Do NOT reward generic medical knowledge unless it is accurately connected to this patient.\\

Penalize reasoning that:\\
- Hallucinates unsupported major diagnoses, organ failures, treatments, procedures, or events.\\
- Contradicts the provided patient data.\\
- Misreads or reverses the temporal trajectory.\\
- Treats time-series measurements as static when the trend is prognostically important.\\
- Overemphasizes minor or isolated abnormalities while ignoring the main prognostic problem.\\
- Gives very high or very low mortality risk without evidence proportional to that confidence.\\
- Cherry-picks only risk-increasing or only risk-decreasing evidence when the data are mixed.\\
~\\

\# Adapted Revised-IDEA Rubric for ICU Mortality Reasoning\\
~\\

Score the reasoning out of 10 points using four domains:\\
~\\
I. Interpretive Prognostic Summary: 0-4 points\\
D. Differential Prognostic Assessment: 0-2 points\\
E. Explanation of Lead Prognostic Judgment: 0-2 points\\
A. Alternative Prognostic Explanation / Counterevidence: 0-2 points\\
~\\

The total score is I + D + E + A, from 0 to 10.\\

The scoring should emphasize clinical reasoning quality, factuality, temporal awareness, and calibration.
Do not mechanically reward length or the number of facts listed.\\
~\\

\end{tcblisting}

\vspace{-2mm}
\caption{Prompt used for the quantitative evaluation of generated reasoning traces (1/4). We adopt the IDEA assessment tool as the evaluation rubric.}
\label{fig:prompt_llm_as_a_judge_1_4}
\end{figure}

\onecolumn
\begin{figure}[p]
\footnotesize

\begin{tcblisting}{
    text only,
    halign=left,
    fontupper=\scriptsize,
    title=\textbf{Evaluation: Prompt for IDEA assessment tool (2/4)},
    colbacktitle=gray!30!white,
    coltitle=black,
}
\#\# I --- Interpretive Prognostic Summary, 0-4 points\\
~\\

This assesses whether the reasoning gives an accurate, patient-specific, temporally aware summary of the patient's main prognostic problem.

Prefer reasoning that accurately identifies, when supported by the patient features:\\
~\\

1. Key baseline or contextual mortality-risk factors:\\
   - Examples: age, major comorbidities, chronic organ disease, frailty-relevant context, admission context, or other baseline risks.\\

2. Main acute ICU problem or organ dysfunction:\\
   - Examples: respiratory failure, shock or hemodynamic instability, renal dysfunction, metabolic acidosis, neurologic depression, hepatic or coagulation dysfunction, infection signal, multi-organ dysfunction, or other acute problems explicitly supported by the data.\\

3. Illness time course or trajectory:\\
   - Correctly distinguishes worsening, improving, fluctuating, or stable course.\\
   - Gives special attention to persistent instability or late deterioration near the end of the observation window.\\

4. Clinically meaningful prognostic abstractions:\\
   - Examples: persistent hypotension, escalating oxygen needs, worsening renal function, severe acidosis, multi-organ dysfunction, improving hemodynamics, stable oxygenation, transient isolated abnormality.\\
   - These abstractions must be grounded in the provided data.\\
~\\

Scoring anchors:\\
- 4: Accurate, patient-specific, and temporally aware summary that captures the main acute problem, key relevant baseline/contextual risks if provided, trajectory, and meaningful prognostic abstractions.\\
- 3: Mostly accurate summary with the main prognostic problem identified, but with a minor omission, limited temporal interpretation, or limited abstraction.\\
- 2: Partially accurate summary; mentions some relevant patient-specific factors but misses an important main problem, trajectory, or severity distinction, or includes minor unsupported statements.\\
- 1: Minimal or generic summary; only one relevant patient-specific element is accurately identified, or the reasoning mostly lists data without integrating the main prognostic problem.\\
- 0: No meaningful prognostic summary, or the summary is mostly contradicted by the patient features or dominated by hallucinated information.\\

Do not penalize the reasoning for not mentioning information that is not provided.\\
Do penalize it for inventing missing baseline risks, diagnoses, treatments, or organ failures.\\
~\\

\#\# D --- Differential Prognostic Assessment, 0-2 points\\
~\\

This assesses whether the reasoning considers competing prognostic interpretations and explicitly prioritizes the most plausible mortality-risk interpretation.

For this ICU mortality task, “differential” means competing mortality-risk hypotheses, not a diagnostic differential diagnosis.\\
~\\

Relevant competing interpretations include:\\
- High vs moderate vs low mortality risk.\\
- Mortality risk driven by multi-organ dysfunction vs lower risk because abnormalities are mild or transient.\\
- High-risk deteriorating trajectory vs stabilization or survival trajectory.\\
- Risk-increasing signals vs risk-decreasing signals.\\
- Uncertainty due to mixed or incomplete evidence.\\
~\\

Scoring anchors:\\
- 2: Clearly states a lead mortality-risk interpretation and explicitly discusses at least one plausible alternative or countervailing interpretation, then prioritizes the lead interpretation using patient-specific data.\\
- 1: Gives a lead mortality-risk interpretation but alternatives are implicit, generic, weakly discussed, or not clearly prioritized; or lists competing ideas without clearly weighing them.\\
- 0: No meaningful prognostic assessment; only a bare conclusion, generic statement, or diagnostic laundry list that does not support mortality-risk reasoning.\\

If the evidence is strongly one-directional, the reasoning can still receive 2 points if it explicitly explains why alternative interpretations are less supported and acknowledges remaining uncertainty appropriately.
\end{tcblisting}

\vspace{-2mm}
\caption{Prompt used for the quantitative evaluation of generated reasoning traces (2/4). We adopt the IDEA assessment tool as the evaluation rubric.}
\label{fig:prompt_llm_as_a_judge_2_4}
\end{figure}

\onecolumn
\begin{figure}[p]
\footnotesize

\begin{tcblisting}{
    text only,
    halign=left,
    fontupper=\scriptsize,
    title=\textbf{Evaluation: Prompt for IDEA assessment tool (3/4)},
    colbacktitle=gray!30!white,
    coltitle=black,
}
\#\# E --- Explanation of Lead Prognostic Judgment, 0-2 points\\
~\\

This assesses whether the reasoning explains why its lead mortality-risk judgment follows from the objective patient data.\\
~\\

Prefer reasoning that:\\
- Clearly links objective patient-specific evidence to the lead mortality-risk judgment.\\
- Uses clinically relevant data such as vital signs, laboratory values, organ-support indicators, severity scores, comorbidities, or temporal trends when provided.\\
- Interprets the severity and persistence of abnormalities appropriately.\\
- Gives a final risk interpretation that is proportional to the evidence.\\
~\\

Calibration requirements:\\
- Very high mortality risk requires strong evidence, such as severe or persistent instability, major organ dysfunction, multiple converging high-risk features, or marked late deterioration.\\
- Very low mortality risk should not ignore major organ dysfunction, severe derangements, or sustained deterioration.\\
- Mild, isolated, or transient abnormalities should not be overstated as extreme risk.\\
- If evidence is mixed, the reasoning should express appropriate uncertainty.\\
~\\

Scoring anchors:\\
- 2: Uses at least two objective patient-specific data points or temporal trends to support the lead judgment, interprets them clinically and accurately, and gives a calibrated final mortality-risk interpretation.\\
- 1: Provides some patient-specific support, but the explanation is incomplete, weakly linked, based on only one main data point, partially miscalibrated, or contains minor unsupported claims.\\
- 0: No explanation of the lead mortality-risk judgment, or the explanation is unsupported, contradicted by the data, dominated by hallucinated facts, or purely generic.\\
~\\

Do not give full credit for simply naming abnormal values without explaining their prognostic relevance.\\
~\\

\#\# A --- Alternative Prognostic Explanation / Counterevidence, 0-2 points\\
~\\

This assesses whether the reasoning identifies and explains patient-specific evidence that could support a different, lower-risk, higher-risk, or more uncertain interpretation.

For this ICU mortality task, “alternative diagnosis explained” is adapted to mean alternative prognostic explanation or counterevidence.\\
~\\

Relevant counterevidence may include:\\
- Reassuring or risk-decreasing factors.\\
- Stable or improving organ function.\\
- Transient rather than sustained abnormalities.\\
- Lack of support for severe conditions that the reasoning might otherwise imply.\\
- Risk-increasing evidence that contradicts an overly reassuring conclusion.\\
- Mixed evidence requiring uncertainty.\\
- Absence of provided evidence for major interventions or complications, without treating missing data as definitely normal.\\
~\\

Scoring anchors:\\
- 2: Identifies and explains at least two patient-specific countervailing data points, alternative prognostic explanations, or sources of uncertainty, and weighs them fairly against the lead judgment.\\
- 1: Mentions one patient-specific countervailing data point or gives a generic/limited discussion of uncertainty or alternative risk interpretation.\\
- 0: No meaningful counterevidence or alternative prognostic explanation; cherry-picks only evidence supporting the conclusion; or treats missing information as definitely normal or definitely abnormal.\\
~\\

Counterevidence can point in either direction:\\
- If the lead judgment is high mortality risk, counterevidence may include stabilization, improvement, preserved organ function, or transient abnormalities.\\
- If the lead judgment is low mortality risk, counterevidence may include persistent instability, organ dysfunction, severe derangements, or late deterioration.\\
\end{tcblisting}

\vspace{-2mm}
\caption{Prompt used for the quantitative evaluation of generated reasoning traces (3/4). We adopt the IDEA assessment tool as the evaluation rubric.}
\label{fig:prompt_llm_as_a_judge_3_4}
\end{figure}

\onecolumn
\begin{figure}[p]
\footnotesize

\begin{tcblisting}{
    text only,
    halign=left,
    fontupper=\scriptsize,
    title=\textbf{Evaluation: Prompt for IDEA assessment tool (4/4)},
    colbacktitle=gray!30!white,
    coltitle=black,
}
\# Overall Score Calibration\\
~\\
Use the following general score interpretation:\\
~\\
- 9-10: Excellent reasoning. Accurate, patient-specific, temporally aware, balanced, well-explained, and well-calibrated. No important hallucinations or major omissions.\\
- 7-8: Good reasoning. Mostly accurate and clinically grounded, with minor omissions, limited counterevidence, or mild calibration issues.\\
- 5-6: Mixed reasoning. Some accurate patient-specific evidence, but important omissions, weak temporal reasoning, one-sided assessment, or moderate calibration problems.\\
- 3-4: Weak reasoning. Generic, poorly grounded, several unsupported claims, major omissions, or significant misreading of the clinical picture.\\
- 0-2: Poor reasoning. Little to no patient-specific reasoning, severe hallucinations, major contradictions, hidden-label reasoning, or conclusion unsupported by the provided data.\\
~\\

\# Internal Deliberation\\
~\\

Before producing the final judgment, internally check:\\
1. What objective patient-specific evidence the reasoning used correctly.\\
2. Any unsupported claims, hallucinations, contradictions, or missing-data assumptions.\\
3. Whether the reasoning correctly handled the temporal trajectory.\\
4. Whether the reasoning considered counterevidence or uncertainty.\\

Do not output a long chain-of-thought.
Output only the JSON object below.\\
~\\

\# Output Format\\
~\\

Return valid JSON only. Do not include markdown, comments, or extra text.

The score and all subscores must be integers.
The final score must equal the sum of the four subscores after applying any cap or penalty.

The confidence field reflects your confidence in the reasoning-quality score, not confidence in the patient's mortality outcome.
The confidence field must be exactly one of: "High", "Medium", or "Low".

Use empty arrays if there are no notable items.\\
~\\
Return this JSON object:\\

\{\\
~~"score": 0 - 10,\\
~~"confidence": "High" | "Medium" | "Low",\\
~~"subscores": \{\\
~~ ~~"I\_interpretive\_prognostic\_summary": 0,\\
~~ ~~"D\_differential\_prognostic\_assessment": 0,\\
~~ ~~"E\_explanation\_of\_lead\_prognostic\_judgment": 0,\\
~~ ~~"A\_alternative\_or\_counterevidence\_explained": 0\\
~~\},\\
~~"rationale": "Briefly explain the score. Mention the most important patient-specific evidence, temporal reasoning, calibration, counterevidence, and any major errors or cap applied.",\\
~~"key\_strengths": [],\\
~~"key\_errors": []\\
\}
\end{tcblisting}

\vspace{-2mm}
\caption{Prompt used for the quantitative evaluation of generated reasoning traces (4/4). We adopt the IDEA assessment tool as the evaluation rubric.}
\label{fig:prompt_llm_as_a_judge_4_4}
\end{figure}

\onecolumn
\begin{figure}[p]
\footnotesize
\begin{tcblisting}{
    text only,
    halign=left,
    fontupper=\scriptsize,
    title=\textbf{Prompt used to detect severe hallucination in reasoning traces},
    colbacktitle=gray!30!white,
    coltitle=black,
}
You are a clinical evidence-grounding judge for ICU mortality/survival rationales.

You will be given:\\
1. Observed patient information from the first 48 hours of ICU stay.\\
2. A rationale explaining why the patient is predicted to die or survive.

Your task is to judge whether the rationale is grounded in the observed patient information, or whether it contains serious hallucination: concrete clinical facts that are completely absent from the patient data.

Return \texttt{True} if the rationale's substantive clinical evidence is present in, paraphrased from, summarized from, or reasonably inferred from the patient information.\\
Return \texttt{False} if the rationale uses at least one clinically meaningful fabricated fact as evidence, such as a diagnosis, treatment, event, demographic detail, lab/vital value, trend, medication, procedure, or condition that is not present in the patient information or is clearly contradicted by it.

Be lenient. The goal is to detect serious hallucination, not minor imprecision.\\
Do not mark \texttt{False} for small wording issues, approximate values, rounding, broad ranges, mild exaggeration, or imperfect temporal alignment.\\
If a value or feature appears anywhere in the patient data and the rationale refers to it in a broadly consistent way, consider it grounded.\\
For example, if the patient data contains PaO2 values of 98, 100, and 125 mmHg, then a rationale saying ``PaO2 around 98--125 mmHg indicating adequate oxygenation'' should be considered grounded, even if the exact timing is not perfectly aligned with another feature such as mechanical ventilation.\\
Do not require every cited feature to co-occur at the exact same timestamp unless the rationale makes a very specific timing claim that is central and clearly contradicted.\\
Accept clinical interpretations such as ``stable,'' ``adequate oxygenation,'' ``renal dysfunction,'' ``shock,'' or ``high mortality risk'' if they are reasonably supported by the observed measurements.\\
Do not evaluate whether the mortality/survival prediction is medically correct; only judge whether the rationale is based on information actually observed in the input.

Mark \texttt{False} only when there is a clear, material unsupported claim, for example:\\
- The rationale cites lactate, creatinine, vasopressors, mechanical ventilation, sepsis, pneumonia, cancer, renal failure, surgery, or another condition/treatment/event that is not present anywhere in the patient data.\\
- The rationale gives a concrete numeric value or range that is grossly inconsistent with the observed values.\\
- The rationale describes a trend that is clearly opposite to the observed trend.\\
- The rationale relies on medical history or outcome information not included in the patient data.

If the unsupported statement is only a minor non-evidentiary phrase and the main clinical evidence is grounded, return \texttt{True}.\\
When uncertain whether something is a minor overstatement or a serious hallucination, prefer \texttt{True}.

\vspace{1mm}
Input:\\
\texttt{[Patient Information]}\\
\texttt{"""\{patient\_information\}"""}

\texttt{[Rationale]}\\
\texttt{"""\{rationale\}"""}

\vspace{1mm}
Output only a Python-style dict with exactly these two keys:\\
\texttt{\{}\\
\texttt{~~"reason": "\textless brief explanation in 10 sentences or fewer\textgreater ",}\\
\texttt{~~"result": True or False}\\
\texttt{\}}
\end{tcblisting}
\vspace{-2mm}
\caption{Prompt used to detect severe hallucination in reasoning traces.
The judge model decides whether the rationale's substantive clinical evidence is grounded in the input. Only clearly fabricated facts such as unobserved diagnoses, treatments, lab values, or trends contradicting the data are marked as hallucinations.}
\label{fig:prompt_hallu_detect}
\end{figure}